%% file: paper.tex
\pdfoutput=1
\documentclass[11pt]{article}

\usepackage{graphicx}
\usepackage{amsthm}
\usepackage{amsmath}
\usepackage{amsfonts}
\usepackage{amssymb}
\usepackage{nicematrix}
\usepackage{url}

\usepackage[bottom]{footmisc}
\usepackage[margin=1in]{geometry}
\usepackage[shortlabels]{enumitem}

\usepackage{listings}
\usepackage{xcolor}
\usepackage{tikz}
\usetikzlibrary{positioning, fit, shapes.geometric, calc}
\usepackage{placeins}
\usepackage{float}
\usepackage{caption}
\usepackage{booktabs}
\usepackage{microtype}
\usepackage{pgfplots}
\pgfplotsset{compat=1.18}
\usepgfplotslibrary{groupplots}
\usepackage[hidelinks]{hyperref}

\definecolor{qcolor}{RGB}{30, 144, 255}  
\definecolor{kcolor}{RGB}{220, 20, 60}   
\definecolor{vcolor}{RGB}{34, 139, 34}   
\definecolor{scolor}{RGB}{255, 140, 0}   
\definecolor{statecolor}{RGB}{128, 0, 128}
\definecolor{codegray}{rgb}{0.5,0.5,0.5}
\definecolor{backcolour}{rgb}{0.95,0.95,0.92}

\lstdefinestyle{blogstyle}{
    backgroundcolor=\color{backcolour},
    commentstyle=\color{codegray},
    keywordstyle=\color{qcolor}\bfseries,
    basicstyle=\ttfamily\footnotesize,
    breaklines=true,
    keepspaces=true,
    showspaces=false,
    showstringspaces=false,
    language=Python,
    columns=fullflexible,
}
\definecolor{flacolor}{HTML}{FF8C00}
\definecolor{qlacolor}{HTML}{6495ED}
\definecolor{fkdacolor}{HTML}{9370DB}
\definecolor{mssmcolor}{HTML}{B22222}
\definecolor{ourscolor}{HTML}{76B900}
\definecolor{oursoncolor}{HTML}{2F6E00}
\pgfplotsset{
    perfaxis/.style={
        ybar, bar width=5pt, ybar=1pt,
        width=0.3\textwidth, height=3.0cm,
        symbolic x coords={2048,8192,32768}, xtick=data,
        enlarge x limits=0.22,
        ymin=0,
        ymajorgrids, grid style={gray!25},
        axis line style={gray!60},
        tick label style={font=\scriptsize},
        ylabel style={font=\scriptsize, yshift=-4pt},
        xlabel style={font=\scriptsize, yshift=3pt},
        title style={font=\small, yshift=-2pt, align=center},
        every axis plot/.append style={draw=black!50, line width=0.2pt},
        clip=false,
    },
}
\tikzset{speedup/.style={font=\scriptsize\bfseries, anchor=south, yshift=1pt}}

\newcommand{\trans}{\mathsf{T}}
\newcommand{\diag}{\textnormal{diag}}
\newcommand{\tr}{\textnormal{tr}}
\newcommand{\cudnn}{cuDNN}

\allowdisplaybreaks
\title{\cudnn{} Linear Attention: Optimized Mamba-2, GDN, GDP, KDA, and GDN-2 Kernels for NVIDIA GPUs}
\author{Joshua Park, Jingze Cui, Sherry Jiang, Mengyu Guo,\\ Anerudhan Gopal, Vedaanta Agarwalla, Timmy Liu, Yu-Jung Chen, Yang Xu\\[0.4em] \normalsize NVIDIA}
\date{}

\begin{document}
\maketitle

\begin{abstract}
Linear attention replaces the softmax attention of Transformers with a fixed-size recurrent state, so that the runtime of prefill grows linearly rather than quadratically with the sequence length and the runtime of every decode step is constant in it. Many recent models have adopted a hybrid architecture where some layers swap dense attention for linear attention variants. \cudnn{} now ships optimized fused kernels for five linear attention variants, Mamba-2, Gated DeltaNet (GDN), Gated DeltaProduct (GDP), Kimi Delta Attention (KDA), and Gated DeltaNet-2 (GDN-2), on Blackwell (SM100, SM103), covering the training forward and backward passes. All of the kernels are open source and ship in \cudnn{} at \url{https://github.com/NVIDIA/cudnn-frontend}. In this paper, we walk through the optimizations that produce speedups over the state-of-the-art baselines. We start the kernel design from a single persistent kernel per pass. Then, in low $B \cdot H$ settings, where a grid of one CTA per \texttt{(batch, head)} pair does not cover the SMs, we use intra-card context parallelism (intra-card CP) or a split of the value dimension (split-$d_v$), both introduced by prior work in FlashQLA. We also fuse the other components that surround the linear attention operation in the block, the L2 normalization of the queries and keys and the gate and write-strength transforms, in both the forward and the backward pass. On GB200, the fused kernels for the four delta-rule variants run up to $7.1\times$ faster than the reference Triton implementations in the forward pass and up to $8.1\times$ in the backward pass, up to $3.2\times$ faster than FlashQLA, and up to $6.1\times$ faster than FlashKDA. The Mamba-2 kernels run up to $15.9\times$ faster than FLA and up to $4.2\times$ faster than mamba\_ssm in the forward pass, and up to $6.9\times$ and $5.5\times$ in the backward pass.
\end{abstract}

\section{Introduction}
\label{sec:intro}

Traditional softmax attention computes a score between every query and every key, so the cost of prefill grows with the square of the sequence length, and decode must keep a key-value (KV) cache that grows with every token. With today's models serving context windows of hundreds of thousands to millions of tokens and spending much of their compute on long agentic trajectories, traditional attention starts to become a bottleneck.

Linear attention replaces the cache with a recurrent state of fixed size, so that prefill grows linearly with the sequence length, and every decode step runs in constant time with respect to the sequence length. A fixed-size state does have its drawbacks though. As such, many new models adopt a hybrid approach, in which a few layers of softmax attention sit among many layers of linear attention. Recently, many open-source models have adopted this approach. NVIDIA's Nemotron 3 family of hybrid Mamba-Transformer mixture-of-experts models~\cite{nvidia2025nemotron3} showed that Mamba-2 could be used at scale. Qwen3-Next~\cite{qwen2025qwen3next} and the Qwen3.8 family~\cite{qiu2026qwen38next} adopted Gated DeltaNet. Finally, Kimi Linear~\cite{kimiteam2025kimilinearexpressiveefficient} and Kimi K3~\cite{kimiteam2026kimik3} are built on top of KDA.

In this paper, we present the optimized \cudnn{} linear attention kernels for five variants, Mamba-2, GDN, GDP, KDA, and GDN-2, on Blackwell (SM100, SM103), covering the training forward and backward passes. We fuse every stage of each pass into one persistent kernel that keeps the recurrent state in tensor memory in fp32 for the whole sequence (Section~\ref{sec:fused}). We keep every \texttt{tcgen05} MMA at $M = 128$ by computing transposed products and stacking operands and carefully choose the matrix inversion algorithms for better utilization and numerics (Section~\ref{sec:chunk}). A fused kernel has a grid of only $B \cdot H$ CTAs, so at low $B \cdot H$ we apply intra-card context parallelism, split-$d_v$, and split-K, all three building on FlashQLA. The specific strategy is selected per shape by the number of segments we can split each sequence into (Section~\ref{sec:regimes}). Finally, we fuse the L2 normalization, the gate transforms, and the write-strength sigmoid that surround the recurrence into both passes (Section~\ref{sec:block}). On GB200, the delta-rule kernels run up to $7.1\times$ faster than FLA in the forward pass and up to $8.1\times$ in the backward pass, up to $3.2\times$ faster than FlashQLA and $6.1\times$ faster than FlashKDA, and the Mamba-2 kernels up to $15.9\times$ faster than FLA and $4.2\times$ faster than mamba\_ssm in the forward pass and up to $6.9\times$ and $5.5\times$ in the backward pass (Section~\ref{sec:perf}). All of the kernels are open source in \cudnn{}.

\section{Background}
\label{sec:background}

Standard attention~\cite{vaswani2023attentionneed} computes the following for queries, keys, and values $\mathbf{Q}, \mathbf{K} \in \mathbb{R}^{s \times d_{qk}}$ and $\mathbf{V} \in \mathbb{R}^{s \times d_{v}}$.
\begin{align*}
\mathbf{A} &= \mathbf{Q} \mathbf{K}^\trans \in \mathbb{R}^{s \times s} && \textcolor{gray}{\mathcal{O}\left(s^2 d_{qk}\right)} \\
\mathbf{P} &= \text{softmax}\left(\mathbf{A}\right) \in \mathbb{R}^{s \times s} && \textcolor{gray}{\mathcal{O}\left(s^2\right)} \\
\mathbf{O} &= \mathbf{P} \mathbf{V} \in \mathbb{R}^{s \times d_{v}} && \textcolor{gray}{\mathcal{O}\left(s^2 d_{v}\right)}
\end{align*}
Both the compute and the size of the intermediate $\mathbf{A}$ scale quadratically with the sequence length $s$. If we remove the softmax, associativity gives us a nice reordering.
\[
\mathbf{O} \;=\; \left(\mathbf{Q}\mathbf{K}^{\trans}\right)\mathbf{V} \;=\; \mathbf{Q}\left(\mathbf{K}^{\trans}\mathbf{V}\right)
\]
The left grouping is quadratic in $s$. However, the right grouping computes $\mathbf{K}^{\trans}\mathbf{V} \in \mathbb{R}^{d_{qk} \times d_{v}}$ first, and its cost is $\mathcal{O}\left(s\, d_{qk} d_{v}\right)$, linear in the sequence length. Importantly, both compute exactly the same $\mathbf{O}$. This observation, refined through feature maps and normalization choices~\cite{katharopoulos2020transformersrnnsfastautoregressive}, is the starting point of linear attention.

\subsection{State as a Compressed KV Cache}

With a causal mask, token $t$ may only attend to tokens $j \le t$, so the sum picks up a time-dependent upper limit. Keeping a running cumulative sum turns the computation into the following recurrence.
\begin{align*}
\textcolor{red}{\mathbf{S}_{t}} &= \textcolor{red}{\mathbf{S}_{t-1}} + \mathbf{k}_t^\trans \mathbf{v}_t \\
\mathbf{o}_t &= \mathbf{q}_{t}\textcolor{red}{\mathbf{S}_{t}}
\end{align*}
This is similar to a recurrent neural network (RNN). The hidden state $\textcolor{red}{\mathbf{S}_{t}} \in \mathbb{R}^{d_{qk} \times d_{v}}$ has a fixed size, independent of the sequence length, so there is no KV cache to grow and decode costs $\mathcal{O}(1)$ per token. Intuitively, $\mathbf{S}_t$ is the KV cache, compressed into a $d_{qk} \times d_{v}$ matrix. Writing $\mathbf{k}_t^\trans \mathbf{v}_t$ files the value $\mathbf{v}_t$ under the key direction $\mathbf{k}_t$, and multiplying by a query reads it back out. The downside of a fixed-size memory is that it can only hold so much, and that limitation motivates the two mechanisms behind the gated delta rule family, decay and the delta rule.

\subsubsection{Decay (Forgetting)}

The first mechanism lets the model shrink old content with a data-dependent gate $\alpha_t \in (0, 1]$ before each write, using the recurrence $\mathbf{S}_t = \alpha_{t} \mathbf{S}_{t-1} + \mathbf{k}_{t}^\trans \mathbf{v}_t$.

\begin{figure}[H]
\centering
\begin{tikzpicture}[
    token/.style={rectangle, draw=gray!30, fill=gray!5, minimum width=1.1cm, minimum height=0.6cm, font=\footnotesize\ttfamily, rounded corners=2pt},
    reset/.style={fill=red!10, draw=red!30},
    alpha_val/.style={font=\scriptsize\bfseries, text=black!70},
    alpha_reset/.style={font=\scriptsize\bfseries, text=red!80!black},
    m_keep/.style={rectangle, fill=blue!40, minimum height=0.25cm, rounded corners=1pt},
    m_reset/.style={rectangle, fill=red!50, minimum height=0.25cm, rounded corners=1pt}
]
    \node[token] (t1) {The};
    \node[token, right=0.06cm of t1] (t2) {cat};
    \node[token, right=0.06cm of t2] (t3) {in};
    \node[token, right=0.06cm of t3] (t4) {the};
    \node[token, right=0.06cm of t4] (t5) {hat};
    \node[token, right=0.06cm of t5, reset] (t6) {.};

    \node[below=0.15cm of t1, alpha_val] (a1) {0.99};
    \node[below=0.15cm of t2, alpha_val] (a2) {0.98};
    \node[below=0.15cm of t3, alpha_val] (a3) {0.97};
    \node[below=0.15cm of t4, alpha_val] (a4) {0.96};
    \node[below=0.15cm of t5, alpha_val] (a5) {0.99};
    \node[below=0.15cm of t6, alpha_reset] (a6) {0.50};

    \node[m_keep, minimum width=0.990cm, anchor=north west] (m1) at ([xshift=0.055cm, yshift=-0.15cm]a1.south -| t1.west) {};
    \node[m_keep, minimum width=0.970cm, anchor=north west] (m2) at ([xshift=0.055cm, yshift=-0.15cm]a2.south -| t2.west) {};
    \node[m_keep, minimum width=0.941cm, anchor=north west] (m3) at ([xshift=0.055cm, yshift=-0.15cm]a3.south -| t3.west) {};
    \node[m_keep, minimum width=0.903cm, anchor=north west] (m4) at ([xshift=0.055cm, yshift=-0.15cm]a4.south -| t4.west) {};
    \node[m_keep, minimum width=0.894cm, anchor=north west] (m5) at ([xshift=0.055cm, yshift=-0.15cm]a5.south -| t5.west) {};
    \node[m_reset, minimum width=0.447cm, anchor=north west] (m6) at ([xshift=0.055cm, yshift=-0.15cm]a6.south -| t6.west) {};

    \node[font=\scriptsize, anchor=east] at ([xshift=-0.25cm]a1.west -| t1.west) {Gate $\alpha_t$};
    \node[font=\scriptsize, anchor=east] at ([xshift=-0.25cm]m1.west -| t1.west) {Cumulative Product $\prod_{r \le t} \alpha_r$};
\end{tikzpicture}
\caption{Data-dependent decay.}
\label{fig:decay}
\end{figure}
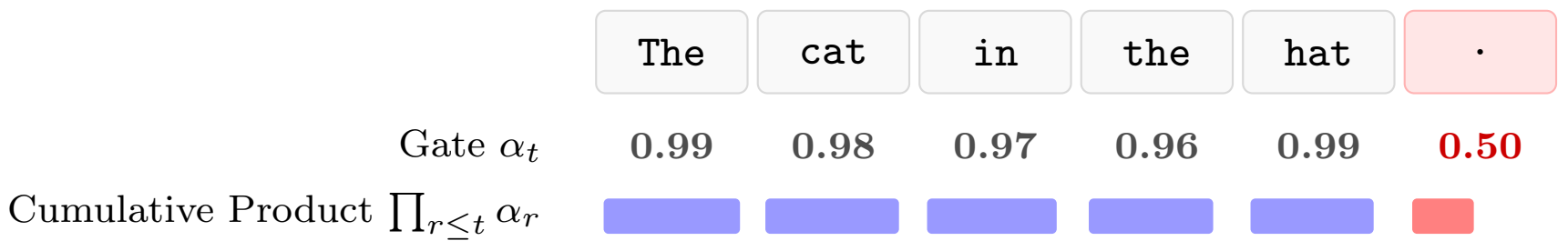

\subsubsection{Delta Rule (Editing)}

Plain accumulation cannot overwrite. If token 3 arrives with the same key as token 1 but a different value, retrieval returns $\mathbf{v}_1 + \mathbf{v}_3$, and neither value is recoverable. DeltaNet~\cite{schlag2021lineartransformerssecretlyfast, yang2025parallelizinglineartransformersdelta} fixes this by writing only the error between the new value and what the memory already returns for that key.
\[
\mathbf{S}_{t} = \mathbf{S}_{t-1} + \beta_{t} \mathbf{k}_t^\trans \underbrace{\left(\mathbf{v}_t - \mathbf{k}_t\mathbf{S}_{t-1}\right)}_{\textcolor{red}{\text{error term}}}
\]
The learning rate $\beta_t = \sigma(\mathbf{x}_t \mathbf{W}_\beta)$ is learned and data dependent, where $\mathbf{x}_t$ is the input token.

\subsection{The Five Variants}

The kernels in this paper implement five members of the linear attention family. They differ in whether they use the delta rule and in how finely the gates are resolved.

\subsubsection{Mamba-2}

Mamba-2~\cite{dao2024transformersssmsgeneralizedmodels} uses decay alone and has no delta rule. In our notation, $\mathbf{q}_t$ and $\mathbf{k}_t$ are the input-dependent $\mathbf{C}_t$ and $\mathbf{B}_t$ projections, $\mathbf{v}_t$ is the input $\mathbf{x}_t$, and the step size $\Delta_t = \mathrm{softplus}(\delta_t + b_{\Delta}) > 0$ sets both the decay, $\alpha_t = \exp(A\,\Delta_t)$ with a learned $A < 0$ per head, and the write strength.
\[
\mathbf{S}_{t} = \alpha_{t}\mathbf{S}_{t-1} + \Delta_{t}\,\mathbf{k}_t^\trans \mathbf{v}_t \qquad
\mathbf{o}_{t} = \mathbf{q}_{t} \mathbf{S}_{t}
\]
Mamba-2 also shares $\mathbf{B}_t$ and $\mathbf{C}_t$ across a group of heads and may add a skip term $\mathbf{D} \odot \mathbf{x}_t$ to the output, and our kernels fuse both. We ship optimized Mamba-2 forward and backward kernels built on a similar design to the other four variants. Most of what we will discuss, the single fused kernel, the strategies for low $B \cdot H$, checkpointing, and the fused gate transform, applies to them as well. Nevertheless, the rest of this paper focuses on the four variants that combine decay with the delta rule, since the delta rule brings the chunk-local inverse and some additional kernel design questions with it.

\subsubsection{Gated DeltaNet (GDN)}

GDN~\cite{yang2025gateddeltanetworksimproving} combines decay and the delta rule with scalar per-token gates $\alpha_t, \beta_t \in \mathbb{R}$.
\[
\mathbf{S}_{t} = \alpha_{t}\mathbf{S}_{t-1} + \beta_{t} \mathbf{k}_t^\trans \left(\mathbf{v}_t - \alpha_{t}\mathbf{k}_t\mathbf{S}_{t-1}\right) \qquad
\mathbf{o}_{t} = \mathbf{q}_{t} \mathbf{S}_{t}
\]

\subsubsection{Gated DeltaProduct (GDP)}
\label{sec:gdp}

GDP~\cite{siems2025deltaproduct} keeps GDN's scalar gates and applies $n_h$ delta-rule updates per token instead of one, each with its own key, value, and write strength.
\begin{align*}
\mathbf{S}_{t}^{(0)} &= \alpha_{t}\mathbf{S}_{t-1} \\
\mathbf{S}_{t}^{(j)} &= \mathbf{S}_{t}^{(j-1)} + \beta_{t,j}\,\mathbf{k}_{t,j}^\trans\left(\mathbf{v}_{t,j} - \mathbf{k}_{t,j}\mathbf{S}_{t}^{(j-1)}\right) && j = 1, \ldots, n_h \\
\mathbf{o}_{t} &= \mathbf{q}_{t} \mathbf{S}_{t}^{(n_h)}
\end{align*}
The per-token transition becomes a product of $n_h$ generalized Householder matrices $\mathbf{I} - \beta_{t,j}\mathbf{k}_{t,j}^\trans\mathbf{k}_{t,j}$, which can express rotations and permutations of the state. Note that $n_h = 1$ reduces exactly to GDN. From an implementation perspective, GDP is GDN on a longer token timeline. Each token expands into $n_h$ sub-token rows of $\mathbf{K}$, $\mathbf{V}$, and $\beta$, the decay attaches to the first sub-token of each token, and the output $\mathbf{o}_t$ is from the last sub-token.

\subsubsection{Kimi Delta Attention (KDA)}

KDA~\cite{kimiteam2025kimilinearexpressiveefficient} promotes the decay to a per-channel vector $\boldsymbol{\alpha}_t \in (0,1]^{d_{qk}}$, so each of the key channels of the state forgets at its own rate, while $\beta_t$ stays scalar.
\[
\mathbf{S}_{t} = \diag\left(\boldsymbol{\alpha}_{t}\right)\mathbf{S}_{t-1} + \beta_{t} \mathbf{k}_t^\trans \left(\mathbf{v}_t - \left(\mathbf{k}_t \odot \boldsymbol{\alpha}_{t}\right)\mathbf{S}_{t-1}\right) \qquad
\mathbf{o}_{t} = \mathbf{q}_{t} \mathbf{S}_{t}
\]

\subsubsection{Gated DeltaNet-2 (GDN-2)}

Finally, GDN-2~\cite{hatamizadeh2026gdn2} observes that erasing and writing act on different axes of the state, whereas prior models tie both to the one scalar $\beta_t$. It decouples them into a per-key-channel erase gate $\boldsymbol{\beta}_t \in \mathbb{R}^{d_{qk}}$ and a per-value-channel write gate $\mathbf{w}_t \in \mathbb{R}^{d_v}$, keeping KDA's per-channel decay.
\[
\mathbf{S}_{t} = \diag\left(\boldsymbol{\alpha}_{t}\right)\mathbf{S}_{t-1} + \mathbf{k}_t^\trans \left(\left(\mathbf{w}_t \odot \mathbf{v}_t\right) - \left(\boldsymbol{\beta}_t \odot \mathbf{k}_t \odot \boldsymbol{\alpha}_{t}\right)\mathbf{S}_{t-1}\right) \qquad
\mathbf{o}_{t} = \mathbf{q}_{t} \mathbf{S}_{t}
\]
Collapsing $\boldsymbol{\beta}_t = \mathbf{w}_t$ to a shared scalar recovers KDA exactly, and further collapsing $\boldsymbol{\alpha}_t$ to a scalar recovers GDN.

\begin{table}[H]
\centering
\small
\begin{tabular}{lcccc}
\toprule
 & decay $\alpha$ & erase gate $\beta$ & write gate $w$ & updates per token \\
\midrule
Mamba-2 & scalar & none & scalar ($\Delta_t$) & 1 \\
GDN   & scalar & scalar & tied to $\beta$ & 1 \\
GDP   & scalar & scalar per update & tied to $\beta$ & $n_h$ \\
KDA   & per key channel & scalar & tied to $\beta$ & 1 \\
GDN-2 & per key channel & per key channel & per value channel & 1 \\
\bottomrule
\end{tabular}
\caption{The five variants and how finely their gates are resolved.}
\label{tab:variants}
\end{table}

\subsection{Chunking}
\label{sec:chunking}

Note that these recurrences as presented are strictly serial. Every $\mathbf{S}_t$ needs $\mathbf{S}_{t-1}$ before it can start. For the delta rule, the carry from one token to the next, $\mathbf{F}_t = \alpha_t(\mathbf{I} - \beta_t \mathbf{k}_t^\trans \mathbf{k}_t)$, is a full $d_{qk} \times d_{qk}$ matrix rather than a scalar. Executing it token by token would issue one matrix-vector product per token, which would not saturate our tensor cores.

The standard fix, developed by the DeltaNet parallelization work~\cite{yang2025parallelizinglineartransformersdelta} and generalized in Gated DeltaNet and FLA~\cite{yang2024fla}, is chunking. Split the sequence into chunks of $C$ tokens. Across chunks, pass the state serially as shown in the recurrence. Within a chunk, unroll the recurrence into dense quadratic matrix products. We start with the recurrence for GDN and GDP. First, define the following chunk-local cumulative decays.
\[
\left(\boldsymbol{\Gamma}_{[t]}\right)_{i} = \prod_{r \le i} \alpha_{r} \qquad
\left(\boldsymbol{\Lambda}_{[t]}\right)_{i} = \prod_{r > i} \alpha_{r} \qquad
\left(\mathbf{L}_{[t]}\right)_{ij} = \left(\mathbf{1}^{\ge}\right)_{ij} \prod_{j < r \le i} \alpha_{r} \qquad
\gamma_{[t]} = \prod_{r} \alpha_{r}
\]
Here, $\mathbf{1}^{\ge}$ is the causal mask with ones on and below the diagonal and zeros above it, and $\mathbf{1}^{>}$ is its strict counterpart with ones only below the diagonal. Thus, $\mathbf{L}_{[t]}$ is zero above the diagonal, one on it, and the product of the decays between $j$ and $i$ below it. The chunk update then becomes the following.
\begin{align*}
\mathbf{Y}_{[t]} &= \mathbf{V}_{[t]} - \left(\boldsymbol{\Gamma}_{[t]} \odot \mathbf{K}_{[t]}\right)\mathbf{S}_{[t-1]} && \text{(read the incoming state)} \\
\mathbf{U}_{[t]} &= \mathbf{T}_{[t]}\, \mathbf{Y}_{[t]} && \text{(resolve within-chunk erases)} \\
\mathbf{S}_{[t]} &= \gamma_{[t]}\, \mathbf{S}_{[t-1]} + \mathbf{K}_{[t]}^{\trans}\left(\boldsymbol{\Lambda}_{[t]} \odot \mathbf{U}_{[t]}\right) && \text{(state update)} \\
\mathbf{O}_{[t]} &= \underbrace{\left(\boldsymbol{\Gamma}_{[t]} \odot \mathbf{Q}_{[t]}\right) \mathbf{S}_{[t-1]}}_{\textcolor{qcolor}{\text{inter-chunk (linear)}}} + \underbrace{\Big[\big(\mathbf{Q}_{[t]}\mathbf{K}_{[t]}^{\trans}\big) \odot \mathbf{L}_{[t]}\Big]\mathbf{U}_{[t]}}_{\textcolor{kcolor}{\text{intra-chunk (quadratic)}}} && \text{(output)}
\end{align*}
Here, $\mathbf{T}_{[t]}$ and $\mathbf{M}_{[t]}$ are defined as follows.
\[
\mathbf{T}_{[t]} = \mathbf{M}_{[t]}^{-1}\diag\left(\boldsymbol{\beta}_{[t]}\right) \qquad
\mathbf{M}_{[t]} = \mathbf{I} + \mathbf{1}^{>} \odot \left[\diag\left(\boldsymbol{\beta}_{[t]}\right)\left(\mathbf{L}_{[t]} \odot \mathbf{K}_{[t]}\mathbf{K}_{[t]}^{\trans}\right)\right]
\]
$\mathbf{M}_{[t]}$ is unit lower triangular, and this additional structure allows us to avoid a full cubic inversion (Sections~\ref{sec:pairinv} and~\ref{sec:inverse}). For simplicity, we name the intra-chunk score tile $\mathbf{A}_{[t]} = \big(\mathbf{Q}_{[t]}\mathbf{K}_{[t]}^{\trans}\big) \odot \mathbf{L}_{[t]}$ and the decayed corrected values $\mathbf{U}'_{[t]} = \boldsymbol{\Lambda}_{[t]} \odot \mathbf{U}_{[t]}$, so we can write $\mathbf{S}_{[t]} = \gamma_{[t]}\mathbf{S}_{[t-1]} + \mathbf{K}_{[t]}^{\trans}\mathbf{U}'_{[t]}$.

KDA and GDN-2 have the same chunked form with one structural difference. Their decay is per key channel, so the cumulative decays gain a channel index $c$ and are defined as follows.
\[
\left(\boldsymbol{\Gamma}_{[t]}\right)_{ic} = \prod_{r \le i} \alpha_{r,c} \qquad
\left(\boldsymbol{\Lambda}_{[t]}\right)_{ic} = \prod_{r > i} \alpha_{r,c} \qquad
\left(\mathbf{L}_{[t]}\right)_{ijc} = \left(\mathbf{1}^{\ge}\right)_{ij} \prod_{j < r \le i} \alpha_{r,c} \qquad
\left(\boldsymbol{\gamma}_{[t]}\right)_{c} = \prod_{r} \alpha_{r,c}
\]
Thus, $\boldsymbol{\Gamma}_{[t]}$ and $\boldsymbol{\Lambda}_{[t]}$ become $C \times d_{qk}$ matrices, and the decay mask $\mathbf{L}_{[t]}$ becomes a $C \times C \times d_{qk}$ tensor. Since they now vary along the contraction axis of every product that involves $\mathbf{Q}$ or $\mathbf{K}$, they cannot be applied to accumulators in the epilogue and are instead folded into four operands below.
\begin{align*}
\widehat{\mathbf{Q}}_{[t]} &= \mathbf{Q}_{[t]} \odot \boldsymbol{\Gamma}_{[t]} & \overline{\mathbf{K}}_{[t]} &= \mathbf{K}_{[t]} \odot \boldsymbol{\Gamma}_{[t]}^{-1} \\
\widehat{\mathbf{K}}_{[t]} &= \mathbf{K}_{[t]} \odot \boldsymbol{\Gamma}_{[t]} \;\;\text{(KDA)} \quad \text{or} \quad \boldsymbol{\beta}_{[t]} \odot \mathbf{K}_{[t]} \odot \boldsymbol{\Gamma}_{[t]} \;\;\text{(GDN-2)} & \widetilde{\mathbf{K}}_{[t]} &= \mathbf{K}_{[t]} \odot \boldsymbol{\Lambda}_{[t]}
\end{align*}
Here, $\boldsymbol{\Gamma}_{[t]}^{-1}$ is the elementwise reciprocal, and $\boldsymbol{\beta}_{[t]} \in \mathbb{R}^{C \times d_{qk}}$ holds GDN-2's per-channel erase gates. Since $(\mathbf{L}_{[t]})_{ijc} = (\mathbf{1}^{\ge})_{ij}\,(\boldsymbol{\Gamma}_{[t]})_{ic} / (\boldsymbol{\Gamma}_{[t]})_{jc}$, the operands scaled by $\boldsymbol{\Gamma}_{[t]}$ carry its numerator and $\overline{\mathbf{K}}_{[t]}$ carries its denominator, so only the causal mask remains outside their products. With these operands, the chunk update reads as follows.
\begin{align*}
\mathbf{Y}_{[t]} &= \mathbf{V}_{[t]} - \widehat{\mathbf{K}}_{[t]}\mathbf{S}_{[t-1]} && \text{(read the incoming state)} \\
\mathbf{U}_{[t]} &= \mathbf{T}_{[t]}\, \mathbf{Y}_{[t]} && \text{(resolve within-chunk erases)} \\
\mathbf{S}_{[t]} &= \diag\left(\boldsymbol{\gamma}_{[t]}\right)\mathbf{S}_{[t-1]} + \widetilde{\mathbf{K}}_{[t]}^{\trans}\mathbf{U}_{[t]} && \text{(state update)} \\
\mathbf{O}_{[t]} &= \widehat{\mathbf{Q}}_{[t]}\mathbf{S}_{[t-1]} + \Big[\big(\widehat{\mathbf{Q}}_{[t]}\overline{\mathbf{K}}_{[t]}^{\trans}\big) \odot \mathbf{1}^{\ge}\Big]\mathbf{U}_{[t]} && \text{(output)}
\end{align*}
For KDA, we define the analogue $\mathbf{T}_{[t]} = \Big(\mathbf{I} + \diag(\boldsymbol{\beta}_{[t]})\big(\mathbf{1}^{>} \odot \widehat{\mathbf{K}}_{[t]}\overline{\mathbf{K}}_{[t]}^{\trans}\big)\Big)^{-1}\diag(\boldsymbol{\beta}_{[t]})$, while GDN-2 carries $\boldsymbol{\beta}$ inside $\widehat{\mathbf{K}}_{[t]}$ and its $\mathbf{T}_{[t]}$ is the plain inverse $\mathbf{T}_{[t]} = \big(\mathbf{I} + \mathbf{1}^{>} \odot \widehat{\mathbf{K}}_{[t]}\overline{\mathbf{K}}_{[t]}^{\trans}\big)^{-1}$.

With chunking, since $C$ is a constant, total work stays linear in $s$. As we will later discuss, growing $C$ better suits our tensor cores. The backward pass needs the gradient of every quantity above with respect to every operand, and the math becomes very involved. We spell it out for all four variants in Appendix~\ref{app:bwdmath}.

\section{Related Work}
\label{sec:prior}

This work is based on the pioneering work of three prior open-source projects, which we also used as the baselines. We build on the ideas of all three and push the same algorithms further on Blackwell. Throughout the paper, we acknowledge their contributions.

\subsection{FLA}

FLA's~\cite{yang2024fla} kernels, written in Triton~\cite{tillet2019triton}, established the chunked formulations that we use. The chunked forward of every delta-rule variant in this paper, GDN, GDP, KDA, and GDN-2, decomposes into roughly the same five steps.
\begin{enumerate}[(1)]
    \item cumulative sum over the gate logits to compute the chunk-local decays $\boldsymbol{\Gamma}_{[t]}$, $\boldsymbol{\Lambda}_{[t]}$, and $\gamma_{[t]}$, per token for GDN and GDP and per key channel for KDA and GDN-2
    \item $\mathbf{K}_{[t]}\mathbf{K}_{[t]}^{\trans}$ for GDN and GDP or $\widehat{\mathbf{K}}_{[t]}\overline{\mathbf{K}}_{[t]}^{\trans}$ for KDA and GDN-2, and the inverse that yields $\mathbf{T}_{[t]}$
    \item application of $\mathbf{T}_{[t]}$ to compute $\mathbf{W} = \mathbf{T}_{[t]}(\boldsymbol{\Gamma}_{[t]} \odot \mathbf{K}_{[t]})$ and $\mathbf{T}_{[t]}\mathbf{V}_{[t]}$
    \item sequential state $\mathbf{S}_{[t]}$ pass
    \item output step to form $\mathbf{O}_{[t]}$
\end{enumerate}
The backward decomposes into seven steps, two of which re-run forward steps to recompute intermediates.
\begin{enumerate}[(1)]
    \item recomputation of $\mathbf{W} = \mathbf{T}_{[t]}(\boldsymbol{\Gamma}_{[t]} \odot \mathbf{K}_{[t]})$ and $\mathbf{T}_{[t]}\mathbf{V}_{[t]}$ (forward step (3))
    \item recomputation of $\mathbf{S}_{[t]}$ (forward step (4))
    \item intra-chunk $d\mathbf{V}$ gradient
    \item sequential $d\mathbf{S}$ pass, accumulating the $d\mathbf{V}$ contributions
    \item $d\mathbf{Q}$, $d\mathbf{K}$, $d\mathbf{W}$, and partial $d\mathbf{g}$ gradients from the output path and $\mathbf{A}_{[t]}$
    \item inverse backward, contributing to $d\mathbf{K}$, $d\mathbf{V}$, $d\beta$, and $d\mathbf{g}$ back through the correction $\mathbf{U} = \mathbf{T}\mathbf{Y}$
    \item reverse cumulative sum to finish $d\mathbf{g}$
\end{enumerate}
Every implementation we discuss covers the same steps, and they differ in how many of them they fuse. FLA generally launches each step as one or more kernels of its own, and every handoff between them is a round trip through global memory.

\subsection{FlashQLA}
\label{sec:fqla}

FlashQLA~\cite{qwen2026flashqla} fuses steps (3) to (5) of that forward into one warp-specialized TileLang~\cite{wang2025tilelang} kernel, and keeps steps (1) and (2) as separate launches ahead of it. Inside the fused kernel, each chunk issues $\mathbf{Q}_{[t]}\mathbf{K}_{[t]}^{\trans}$, the state read $\mathbf{K}_{[t]}\mathbf{S}_{[t-1]}$, the residual $\mathbf{Y}_{[t]} = \mathbf{V}_{[t]} - \mathbf{K}_{[t]}\mathbf{S}_{[t-1]}$, the correction $\mathbf{U}_{[t]} = \mathbf{T}_{[t]}\mathbf{Y}_{[t]}$ from the precomputed $\mathbf{M}_{[t]}^{-1}$, the inter-chunk output $\mathbf{Q}_{[t]}\mathbf{S}_{[t-1]}$, the intra-chunk output $\mathbf{A}_{[t]}\mathbf{U}_{[t]}$, and the state update. Its backward fuses step (1) and steps (3) to (6) into a second warp-specialized kernel. A separate kernel ahead of it recomputes the per-chunk states $\mathbf{S}_{[t]}$, for backward step (2), and the reverse cumulative sum of step (7) remains separate at the very end.

As we will later discuss, the fusion causes some under-utilization of SMs at smaller shapes. Therefore, FlashQLA also uses gate-driven intra-card context parallelism. When a batch of one long sequence would leave the GPU underfilled, the sequence is cut into equal segments. Each segment's incoming state is rebuilt by a warmup pre-pass from a zero state over the gate-determined number of chunks before the segment for which the gates have decayed below $e^{-10}$. When no such window exists, the pre-pass also forms the segment's exact transfer matrix, and a small correction kernel chains the segment states through it. Then, the fused kernel runs one CTA per segment, resulting in better occupancy. In Section~\ref{sec:regimes}, we will further expand on this idea by splitting it into an exact chain and a gate-driven skip.

\subsection{FlashKDA}

FlashKDA~\cite{moonshot2026flashkda} is a KDA prefill implementation built on CUTLASS~\cite{nvidia2017cutlass} that similarly attempts to fuse some of the above steps. A preparation kernel normalizes $\mathbf{q}$ and $\mathbf{k}$, applies the gate activation and its cumulative sum, forms the decay-folded operands $\widehat{\mathbf{K}}_{[t]}$, $\widehat{\mathbf{Q}}_{[t]}$, $\widetilde{\mathbf{K}}_{[t]}$, and $\boldsymbol{\gamma}_{[t]}$ of Section~\ref{sec:chunking}, computes the $16 \times 16$ inverse that yields $\mathbf{T}_{[t]}$ and $\mathbf{A}_{[t]} = \big(\widehat{\mathbf{Q}}_{[t]}\overline{\mathbf{K}}_{[t]}^{\trans}\big) \odot \mathbf{1}^{\ge}$, and writes these six intermediates per 16-token chunk to global memory. Then, a recurrence kernel reads them and issues, per chunk, the state read $\widehat{\mathbf{K}}_{[t]}\mathbf{S}_{[t-1]}$, the inter-chunk output $\widehat{\mathbf{Q}}_{[t]}\mathbf{S}_{[t-1]}$, the correction $\mathbf{U}_{[t]} = \mathbf{T}_{[t]}\mathbf{Y}_{[t]}$, the intra-chunk output $\mathbf{A}_{[t]}\mathbf{U}_{[t]}$, and the state update $\diag(\boldsymbol{\gamma}_{[t]})\mathbf{S}_{[t-1]} + \widetilde{\mathbf{K}}_{[t]}^{\trans}\mathbf{U}_{[t]}$.

\section{Algorithmic Details}
\label{sec:chunk}

Before discussing the kernels themselves, this section discusses the algorithmic choices we considered. In particular, we consider the chunk size, how $\mathbf{M}_{[t]}$ is inverted, and the checkpoint cadence required for the backward pass.

\subsection{Chunk Size}

$C = 1$ reduces to the serial token recurrence, and $C = s$ materializes the full attention matrix and is quadratic. As we increase $C$, we increase the number of FLOPs, but we also can better utilize the tensor cores. Thus, the goal is the smallest chunk that still feeds the tensor cores and is numerically safe. To fully utilize Blackwell's tensor core lanes, we would first like $M = 128$. While we may still be bound by shared memory reads even if $M = 128$, this is a good starting point. As we will discuss in Sections~\ref{sec:lanes} and~\ref{sec:stack}, this need not mean that $C = 128$.

\subsubsection{GDN and GDP}

For GDN and GDP, we follow FlashQLA and use $C = 64$. Choosing a higher $C$ like $C = 128$ creates high register pressure and also increases the cost of the inverse disproportionately, as the inverse grows quadratically with chunk size. For Mamba-2, since there is no delta rule, the inverse disappears, and the chunk can grow to $C = 128$.

\subsubsection{KDA and GDN-2}

With the decay being per-channel, numerics now becomes the main limiter for chunk size and forces $C = 16$, in line with FlashKDA. The decay $(\mathbf{L}_{[t]})_{ijc} = (\boldsymbol{\Gamma}_{[t]})_{ic} / (\boldsymbol{\Gamma}_{[t]})_{jc}$ between tokens $i$ and $j$ now varies along the contraction axis $c$, so it cannot be hoisted to the accumulator. Instead, it must be folded into the operands themselves, as $\widehat{\mathbf{Q}}_{[t]} = \mathbf{Q}_{[t]} \odot \boldsymbol{\Gamma}_{[t]}$ and $\overline{\mathbf{K}}_{[t]} = \mathbf{K}_{[t]} \odot \boldsymbol{\Gamma}_{[t]}^{-1}$ from Section~\ref{sec:chunking}. The cumulative decay compounds over the chunk, so the longer the chunk, the smaller the entries of $\boldsymbol{\Gamma}_{[t]}$ and the larger those of $\boldsymbol{\Gamma}_{[t]}^{-1}$ can get. Ideally, neither may underflow or overflow. As such, as FlashKDA~\cite{moonshot2026flashkda} notes, $C = 16$ is the largest numerically safe chunk size that we can use with this approach.

\subsection{Swap-AB}
\label{sec:lanes}

As we discussed previously, Blackwell's tensor core instruction (\texttt{tcgen05.mma}) needs $M = 128$ to saturate its tensor core lanes. An MMA with $M = 64$ or $M = 16$ leaves some of the tensor core lanes idle. We choose $C = 64$ for GDN and GDP and $C = 16$ for KDA and GDN-2. In the natural orientation, this does not saturate all the tensor core lanes. To see the problem more concretely, Table~\ref{tab:naive} lists the natural, token-major shapes of the forward's seven MMAs per chunk.

\begin{table}[H]
\centering
\small
\begin{tabular}{lccc}
\toprule
MMA & $M$ & $N$ & $K$ \\
\midrule
$\mathbf{K} \;@\; \mathbf{K}^{\trans}$ & $C$ & $C$ & $d_{qk}$ \\
$\mathbf{Q} \;@\; \mathbf{K}^{\trans}$ & $C$ & $C$ & $d_{qk}$ \\
$\mathbf{K} \;@\; \mathbf{S}$ & $C$ & $d_v$ & $d_{qk}$ \\
$\mathbf{Q} \;@\; \mathbf{S}$ & $C$ & $d_v$ & $d_{qk}$ \\
$\mathbf{T} \;@\; \mathbf{Y}$ & $C$ & $d_v$ & $C$ \\
$\mathbf{A} \;@\; \mathbf{U}$ & $C$ & $d_v$ & $C$ \\
$\mathbf{K}^{\trans} \;@\; \mathbf{U}'$ & $d_{qk}$ & $d_v$ & $C$ \\
\bottomrule
\end{tabular}
\caption{Natural token-major shapes of the forward's seven MMAs per chunk.}
\label{tab:naive}
\end{table}

Only the state update reaches $M = 128$, assuming $d_{qk} = d_{v} = 128$. Every other MMA has the chunk on its $M$ side, 64 rows for GDN and GDP and 16 for KDA and GDN-2. Thus, we use a couple of optimizations to get the desired $M = 128$. The first is swap-AB. Most of the MMAs have the head dimension ($d_{qk}$ or $d_{v}$) on the $N$ dimension, so we compute the transpose instead, and keep every consumer transposed as well. As shown in Table~\ref{tab:swapab}, using transposes, five of the MMAs now have $M = 128$ assuming that $d_v = 128$.

\begin{table}[H]
\centering
\small
\begin{tabular}{lcccl}
\toprule
MMA & $M$ & $N$ & $K$ & \\
\midrule
$\mathbf{K} \;@\; \mathbf{K}^{\trans}$ & $C$ & $C$ & $d_{qk}$ & unchanged \\
$\mathbf{Q} \;@\; \mathbf{K}^{\trans}$ & $C$ & $C$ & $d_{qk}$ & unchanged \\
$\mathbf{S}^{\trans} \;@\; \mathbf{K}^{\trans}$ & $d_v$ & $C$ & $d_{qk}$ & swap-AB \\
$\mathbf{S}^{\trans} \;@\; \mathbf{Q}^{\trans}$ & $d_v$ & $C$ & $d_{qk}$ & swap-AB \\
$\mathbf{Y}^{\trans} \;@\; \mathbf{T}^{\trans}$ & $d_v$ & $C$ & $C$ & swap-AB \\
$\mathbf{U}^{\trans} \;@\; \mathbf{A}^{\trans}$ & $d_v$ & $C$ & $C$ & swap-AB \\
$\mathbf{U}'^{\trans} \;@\; \mathbf{K}$ & $d_v$ & $d_{qk}$ & $C$ & swap-AB \\
\bottomrule
\end{tabular}
\caption{The same seven MMAs after swap-AB.}
\label{tab:swapab}
\end{table}

For KDA and GDN-2, this is especially important. Their chunk is 16, so a token-major MMA would leave $7/8$ of the datapaths idle. We should note that it would be hard to increase beyond $M = 128$ using 2-CTA MMA due to the recurrence. For linear attention, nothing is shared between CTAs, since we loop over all of $\mathbf{Q}$, $\mathbf{K}$, and $\mathbf{V}$ simultaneously.

\subsection{GDN and GDP: Stack $\mathbf{K}$ and $\mathbf{Q}$}
\label{sec:stack}

Table~\ref{tab:swapab} still has its first two MMAs at $M = C$, and we use stacking to fix this for GDN and GDP. Note that the two MMAs, $\mathbf{K}\mathbf{K}^{\trans}$ (for $\mathbf{M}_{[t]}$) and $\mathbf{Q}\mathbf{K}^{\trans}$ (for the intra-chunk attention), share the same right operand $\mathbf{K}^{\trans}$. Each MMA alone has $M = 64$. Thus, we stack the two left operands along $M$ and issue a single $M = 128$ MMA.

\begin{figure}[H]
\centering
\begin{tikzpicture}[scale=0.6, every node/.style={font=\scriptsize}]
    \draw[fill=kcolor!15, draw=kcolor, thick] (2.6,2.2) rectangle (3.8,4.6);
    \node at (3.2,3.4) {$\mathbf{K}^{\trans}$};
    \draw[<->] (2.6,4.85) -- (3.8,4.85);
    \node at (3.2,5.15) {$N = 64$};
    \draw[fill=kcolor!15, draw=kcolor, thick] (0,1) rectangle (2.4,2);
    \node at (1.2,1.5) {$\mathbf{K}$};
    \draw[fill=qcolor!15, draw=qcolor, thick] (0,0) rectangle (2.4,1);
    \node at (1.2,0.5) {$\mathbf{Q}$};
    \draw[<->] (-0.25,0) -- (-0.25,2);
    \node[rotate=90, anchor=south] at (-0.55,1) {$M = 128$};
    \draw[<->] (0,-0.25) -- (2.4,-0.25);
    \node at (1.2,-0.55) {$K = 128$};
    \draw[fill=kcolor!15, draw=kcolor, thick] (2.6,1) rectangle (3.8,2);
    \node at (3.2,1.5) {$\mathbf{K}\mathbf{K}^{\trans}$};
    \draw[fill=qcolor!15, draw=qcolor, thick] (2.6,0) rectangle (3.8,1);
    \node at (3.2,0.5) {$\mathbf{Q}\mathbf{K}^{\trans}$};
\end{tikzpicture}
\caption{The stacked score MMA for GDN. $\mathbf{K}$ and $\mathbf{Q}$ share the right operand $\mathbf{K}^{\trans}$, so one $M = 128$ MMA yields both $\mathbf{K}\mathbf{K}^{\trans}$ and $\mathbf{Q}\mathbf{K}^{\trans}$.}
\label{fig:stack}
\end{figure}
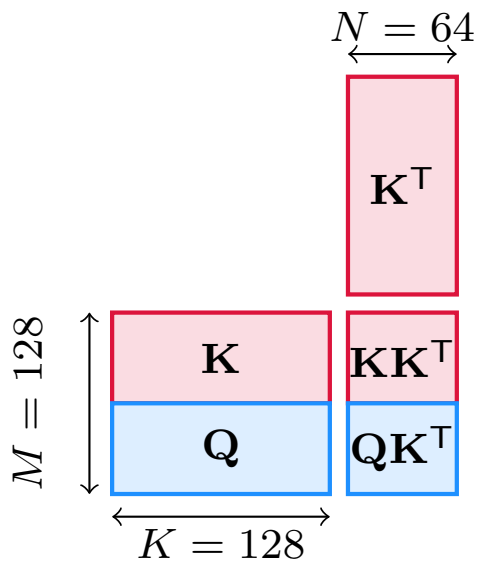

The kernel loads $\mathbf{K}$ and $\mathbf{Q}$ into one interleaved shared memory tile, and it issues this fused MMA once per chunk. With that, the GDN and GDP forward's seven logical products per chunk issue as six MMAs, all at full width, as Table~\ref{tab:stack} shows.

\begin{table}[H]
\centering
\small
\begin{tabular}{lcccl}
\toprule
MMA & $M$ & $N$ & $K$ & \\
\midrule
$[\mathbf{K}; \mathbf{Q}] \;@\; \mathbf{K}^{\trans}$ & 128 & 64 & 128 & fused pair \\
$\mathbf{S}^{\trans} \;@\; \mathbf{K}^{\trans}$ & 128 & 64 & 128 & swap-AB \\
$\mathbf{S}^{\trans} \;@\; \mathbf{Q}^{\trans}$ & 128 & 64 & 128 & swap-AB \\
$\mathbf{Y}^{\trans} \;@\; \mathbf{T}^{\trans}$ & 128 & 64 & 64 & swap-AB \\
$\mathbf{U}^{\trans} \;@\; \mathbf{A}^{\trans}$ & 128 & 64 & 64 & swap-AB \\
$\mathbf{U}'^{\trans} \;@\; \mathbf{K}$ & 128 & 128 & 64 & swap-AB \\
\bottomrule
\end{tabular}
\caption{The GDN and GDP forward's six MMAs per chunk after swap-AB and stacking, at $C = 64$.}
\label{tab:stack}
\end{table}

For KDA and GDN-2, the stacking trick is not available, as $C = 16$. For those MMAs, we will use a separate register-MMA warp described in Section~\ref{sec:regwarp}.

\subsection{GDN and GDP: Inverting $\mathbf{M}_{[t]}$ Blockwise}
\label{sec:pairinv}

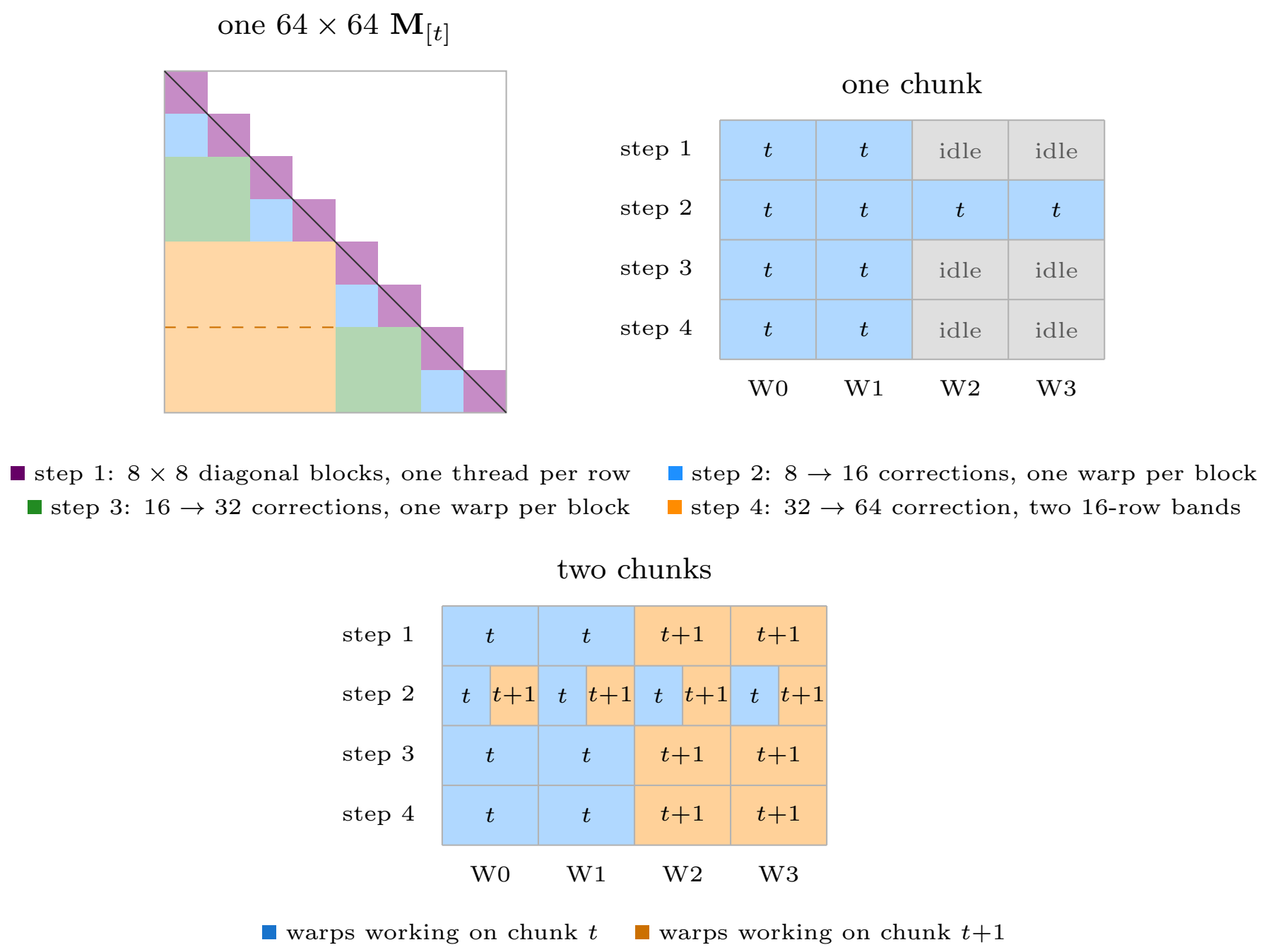
\begin{figure}[H]
\centering
\begin{tikzpicture}[every node/.style={font=\scriptsize}]
  \def\u{0.05}
  \def\cw{0.9}
  \def\ch{0.56}
  \begin{scope}[shift={(0,0)}]
    \fill[gray!8] (0,0) -- (3.2,0) -- (0,3.2) -- cycle;
    \foreach \b in {0,...,7} {
      \fill[statecolor!45] ({\b*8*\u},{3.2-(\b+1)*8*\u}) rectangle ({(\b+1)*8*\u},{3.2-\b*8*\u});
    }
    \foreach \b in {0,...,3} {
      \fill[qcolor!35] ({\b*16*\u},{3.2-(\b+1)*16*\u}) rectangle ({\b*16*\u+8*\u},{3.2-\b*16*\u-8*\u});
    }
    \foreach \b in {0,1} {
      \fill[vcolor!35] ({\b*32*\u},{3.2-(\b+1)*32*\u}) rectangle ({\b*32*\u+16*\u},{3.2-\b*32*\u-16*\u});
    }
    \fill[scolor!35] (0,0) rectangle ({32*\u},{32*\u});
    \draw[dashed, scolor!80!black] (0,{16*\u}) -- ({32*\u},{16*\u});
    \draw[gray!60] (0,0) rectangle (3.2,3.2);
    \draw[black!80] (0,3.2) -- (3.2,0);
    \node[anchor=south, font=\scriptsize] at (1.6,3.3) {one $64 \times 64$ $\mathbf{M}_{[t]}$};
  \end{scope}
  \begin{scope}[shift={(5.2,0.5)}]
    \node[anchor=south, font=\scriptsize] at ({2*\cw},{4*\ch+0.12}) {one chunk};
    \foreach \s/\y in {1/3,2/2,3/1,4/0} { \node[anchor=east, font=\tiny] at (-0.12,{\y*\ch+0.5*\ch}) {step \s}; }
    \foreach \y in {3,1,0} {
      \foreach \w in {0,1} { \fill[qcolor!35] ({\w*\cw},{\y*\ch}) rectangle ({(\w+1)*\cw},{(\y+1)*\ch}); \node[font=\tiny] at ({\w*\cw+0.5*\cw},{\y*\ch+0.5*\ch}) {$t$}; }
      \foreach \w in {2,3} { \fill[gray!25] ({\w*\cw},{\y*\ch}) rectangle ({(\w+1)*\cw},{(\y+1)*\ch}); \node[font=\tiny, text=gray!70!black] at ({\w*\cw+0.5*\cw},{\y*\ch+0.5*\ch}) {idle}; }
    }
    \foreach \w in {0,...,3} { \fill[qcolor!35] ({\w*\cw},{2*\ch}) rectangle ({(\w+1)*\cw},{3*\ch}); \node[font=\tiny] at ({\w*\cw+0.5*\cw},{2.5*\ch}) {$t$}; }
    \foreach \i in {0,...,4} { \draw[gray!60] (0,{\i*\ch}) -- ({4*\cw},{\i*\ch}); \draw[gray!60] ({\i*\cw},0) -- ({\i*\cw},{4*\ch}); }
    \foreach \w in {0,...,3} { \node[anchor=north, font=\tiny] at ({\w*\cw+0.5*\cw},{-0.06}) {W\w}; }
  \end{scope}
  \node[anchor=north, align=center, font=\tiny] at (4.4,-0.35) {%
    \textcolor{statecolor!80!black}{\rule{0.5em}{0.5em}} step 1: $8 \times 8$ diagonal blocks, one thread per row \quad
    \textcolor{qcolor}{\rule{0.5em}{0.5em}} step 2: $8 \to 16$ corrections, one warp per block\\[2pt]
    \textcolor{vcolor}{\rule{0.5em}{0.5em}} step 3: $16 \to 32$ corrections, one warp per block \quad
    \textcolor{scolor}{\rule{0.5em}{0.5em}} step 4: $32 \to 64$ correction, two 16-row bands};
  \begin{scope}[shift={(2.6,-4.05)}]
    \node[anchor=south, font=\scriptsize] at ({2*\cw},{4*\ch+0.12}) {two chunks};
    \foreach \s/\y in {1/3,2/2,3/1,4/0} { \node[anchor=east, font=\tiny] at (-0.12,{\y*\ch+0.5*\ch}) {step \s}; }
    \foreach \y in {3,1,0} {
      \foreach \w in {0,1} { \fill[qcolor!35] ({\w*\cw},{\y*\ch}) rectangle ({(\w+1)*\cw},{(\y+1)*\ch}); \node[font=\tiny] at ({\w*\cw+0.5*\cw},{\y*\ch+0.5*\ch}) {$t$}; }
      \foreach \w in {2,3} { \fill[scolor!40] ({\w*\cw},{\y*\ch}) rectangle ({(\w+1)*\cw},{(\y+1)*\ch}); \node[font=\tiny] at ({\w*\cw+0.5*\cw},{\y*\ch+0.5*\ch}) {$t{+}1$}; }
    }
    \foreach \w in {0,...,3} {
      \fill[qcolor!35] ({\w*\cw},{2*\ch}) rectangle ({\w*\cw+0.5*\cw},{3*\ch}); \node[font=\tiny] at ({\w*\cw+0.25*\cw},{2.5*\ch}) {$t$};
      \fill[scolor!40] ({\w*\cw+0.5*\cw},{2*\ch}) rectangle ({(\w+1)*\cw},{3*\ch}); \node[font=\tiny] at ({\w*\cw+0.75*\cw},{2.5*\ch}) {$t{+}1$};
      \draw[gray!60] ({\w*\cw+0.5*\cw},{2*\ch}) -- ({\w*\cw+0.5*\cw},{3*\ch});
    }
    \foreach \i in {0,...,4} { \draw[gray!60] (0,{\i*\ch}) -- ({4*\cw},{\i*\ch}); \draw[gray!60] ({\i*\cw},0) -- ({\i*\cw},{4*\ch}); }
    \foreach \w in {0,...,3} { \node[anchor=north, font=\tiny] at ({\w*\cw+0.5*\cw},{-0.06}) {W\w}; }
  \end{scope}
  \node[anchor=north, font=\tiny] at (4.4,-4.65) {%
    \textcolor{qcolor!80!black}{\rule{0.5em}{0.5em}} warps working on chunk $t$ \quad
    \textcolor{scolor!80!black}{\rule{0.5em}{0.5em}} warps working on chunk $t{+}1$};
\end{tikzpicture}
\caption{The blockwise inverse of $\mathbf{M}_{[t]}$ and how it occupies a warpgroup. \textbf{Top Left}: The diagram shows one $64 \times 64$ $\mathbf{M}_{[t]}$ with the blocks for each of steps 1 to 4, colored by step. \textbf{Top Right}: The diagram shows which of the four warps have work in each step when one chunk is inverted at a time. Steps 1, 3, and 4 have work for only two warps, so warps 2 and 3 sit idle. \textbf{Bottom}: The diagram shows the same schedule for a chunk pair. Warps 0 and 1 invert $\mathbf{M}_{[t]}$ while warps 2 and 3 invert $\mathbf{M}_{[t+1]}$. All four warps are busy in every step.}
\label{fig:pairinv}
\end{figure}

For GDN and GDP, $\mathbf{M}_{[t]}$ is a $64 \times 64$ unit lower triangular matrix. To avoid a serial Gauss--Jordan inversion for the entire matrix and to take advantage of tensor cores, we use a blockwise inversion scheme.
\[
\begin{pmatrix} \mathbf{A} & \mathbf{0} \\ \mathbf{C} & \mathbf{D} \end{pmatrix}^{-1} = \begin{pmatrix} \mathbf{A}^{-1} & \mathbf{0} \\ -\mathbf{D}^{-1}\mathbf{C}\mathbf{A}^{-1} & \mathbf{D}^{-1} \end{pmatrix}
\]
Applied recursively from the diagonal outward, we transform $\mathbf{M}_{[t]}$ into $\mathbf{T}_{[t]}$ in five steps, as shown in Figure~\ref{fig:pairinv}.
\begin{enumerate}[(1)]
    \item invert the eight $8 \times 8$ diagonal blocks by Gauss--Jordan elimination, one thread per row, with each finished row broadcast to the rows below it through warp shuffles
    \item merge them into four $16 \times 16$ diagonal blocks by correcting each off-diagonal $8 \times 8$ block, $\mathbf{C} \leftarrow -\mathbf{D}^{-1}\mathbf{C}\mathbf{A}^{-1}$, with two register MMAs
    \item merge pairs of $16 \times 16$ blocks into two $32 \times 32$ blocks with the same correction
    \item merge those into the full $64 \times 64$ inverse, using two bands of 16 rows
    \item scale the columns by $\boldsymbol{\beta}_{[t]}$, which turns $\mathbf{M}_{[t]}^{-1}$ into $\mathbf{T}_{[t]}$
\end{enumerate}

Assuming we assign an entire warpgroup of four warps to computing this inverse, inverting a single $64 \times 64$ $\mathbf{M}_{[t]}$ at a time does not keep all four warps busy. Step (1) has 64 rows, so it occupies only 64 threads or two warps. Step (3) has two off-diagonal blocks and step (4) has one block cut into two bands, both only requiring two warps each. Only step (2), with four blocks, keeps all four warps busy. If we inverted one chunk at a time, half of the warpgroup would idle for three of the four steps. Therefore, we choose to invert $\mathbf{M}$ for two chunks at once as illustrated in Figure~\ref{fig:pairinv}. Chunks are processed in pairs, with warps 0 and 1 owning $\mathbf{M}_{[t]}$ and warps 2 and 3 owning $\mathbf{M}_{[t+1]}$.

\subsection{KDA/GDN-2: Neumann Series}
\label{sec:inverse}

For KDA and GDN-2, the matrix we want to invert is now $16 \times 16$, since $C = 16$. This gives us the ability to try a second exact way to invert the matrix using the Neumann series. Let $\mathbf{M}_{[t]} = \mathbf{I} + \mathbf{N}$, where $\mathbf{N}$ is strictly lower triangular, with $\mathbf{N} = \diag(\boldsymbol{\beta}_{[t]})\big(\mathbf{1}^{>} \odot \widehat{\mathbf{K}}_{[t]}\overline{\mathbf{K}}_{[t]}^{\trans}\big)$ for KDA and $\mathbf{N} = \mathbf{1}^{>} \odot \widehat{\mathbf{K}}_{[t]}\overline{\mathbf{K}}_{[t]}^{\trans}$ for GDN-2 (Section~\ref{sec:chunking}). The Neumann series expands the inverse as follows.
\[
\left(\mathbf{I} + \mathbf{N}\right)^{-1} = \sum_{k \ge 0} \left(-\mathbf{N}\right)^{k} = \mathbf{I} - \mathbf{N} + \mathbf{N}^{2} - \mathbf{N}^{3} + \cdots
\]
A $C \times C$ strictly lower triangular matrix is nilpotent with $\mathbf{N}^{C} = \mathbf{0}$, so the series terminates after the term $\mathbf{N}^{C-1}$, and the finite sum is the exact inverse. For $C = 16$, the sum has 16 terms, and it factors as follows.
\[
\left(\mathbf{I} + \mathbf{N}\right)^{-1} = \sum_{k=0}^{15} \left(-\mathbf{N}\right)^{k} = \left(\mathbf{I} - \mathbf{N}\right)\left(\mathbf{I} + \mathbf{N}^{2}\right)\left(\mathbf{I} + \mathbf{N}^{4}\right)\left(\mathbf{I} + \mathbf{N}^{8}\right)
\]
We can evaluate this in three doubling rounds. Starting from $\mathbf{X} = \mathbf{I} - \mathbf{N}$, round $r = 1, 2, 3$ squares the previous power, $\mathbf{N}^{2^r} \leftarrow \left(\mathbf{N}^{2^{r-1}}\right)^{2}$, and folds it in, $\mathbf{X} \leftarrow \mathbf{X} + \mathbf{X}\mathbf{N}^{2^r}$, which is two $16 \times 16$ register MMAs per round. Each $16 \times 16$ product is two \texttt{mma.sync} instructions of shape $16 \times 8 \times 16$, one per 8-column half of the output, so the series costs twelve instructions in six dependent stages. This strategy was first introduced in FlashKDA~\cite{moonshot2026flashkda}.

The Neumann series has a numerical weakness, though. Its intermediate powers are not bounded by the final output. With near-parallel unit keys and $\beta$ near one, every entry of $\mathbf{N}$ below the diagonal is close to $1$. In such a case, the powers of such a matrix grow combinatorially. To see this, let every entry below the diagonal equal $\rho$, so that $\mathbf{N} = \rho\,\mathbf{1}^{>}$. The entry of $\left(\mathbf{1}^{>}\right)^{k}$ at row $i$ and column $j$ counts the chains $j < m_1 < \cdots < m_{k-1} < i$, one for every choice of $k - 1$ intermediate indices among the $i - j - 1$ indices strictly between $j$ and $i$. Hence, the corner entry, with $i - j - 1 = 14$ at $C = 16$, is the largest, so the powers' max element grows as follows.
\[
\max_{i,j}\left|\left(\mathbf{N}^{k}\right)_{ij}\right| = \binom{14}{k-1}\rho^{k}
\]
Here, the binomial peaks at $k = 8$ with $\binom{14}{7} = 3432$. For $\rho$ near one, the third doubling round forms an $\mathbf{N}^{8}$ with entries in the hundreds to thousands. However, the final output is much smaller in magnitude. For this example, the inverse has ones on the diagonal, zeros above it, and the following entries below it.
\[
\left(\left(\mathbf{I} + \mathbf{N}\right)^{-1}\right)_{ij} = -\rho\left(1 - \rho\right)^{i-j-1} \qquad i > j
\]
No entry of the inverse exceeds one in magnitude. Thus, we need exact cancellation of these large quantities. However, every operand of a register MMA is quantized to a 16-bit number, and bf16 carries too few mantissa bits to hold an intermediate in the hundreds to the absolute precision that this cancellation needs, so the rounding error can become non-negligible. $\mathbf{U} = \mathbf{T}\mathbf{Y}$ then amplifies the shared key direction instead of erasing it, and the state grows geometrically until it overflows.

\begin{figure}[H]
\centering
\begin{tikzpicture}[scale=0.8, every node/.style={font=\scriptsize}]
  \def\u{0.2}
  \def\cw{1.6}
  \def\ch{0.5}
  \begin{scope}[shift={(0,0)}]
    \fill[gray!8] (0,0) -- (3.2,0) -- (0,3.2) -- cycle;
    \foreach \b in {0,...,3} {
      \fill[statecolor!45] ({\b*4*\u},{3.2-(\b+1)*4*\u}) rectangle ({(\b+1)*4*\u},{3.2-\b*4*\u});
    }
    \foreach \q in {0,1} {
      \fill[qcolor!35] ({\q*8*\u},{3.2-(\q*8+8)*\u}) rectangle ({\q*8*\u+4*\u},{3.2-(\q*8+4)*\u});
    }
    \fill[vcolor!35] (0,0) rectangle ({8*\u},{8*\u});
    \draw[gray!60] (0,0) rectangle (3.2,3.2);
    \draw[black!80] (0,3.2) -- (3.2,0);
    \node[anchor=south, font=\scriptsize] at (1.6,3.3) {one $16 \times 16$ $\mathbf{I} + \mathbf{N}$};
  \end{scope}
  \begin{scope}[shift={(5.2,0.1)}]
    \node[anchor=south, font=\scriptsize] at ({\cw},{6*\ch+0.2}) {one warp, one chunk};
    \foreach \s/\y in {1/5,2/4,3/3,4/2,5/1,6/0} { \node[anchor=east, font=\tiny] at (-0.12,{\y*\ch+0.5*\ch}) {stage \s}; }
    \foreach \y in {5,4} {
      \fill[statecolor!45] (0,{\y*\ch}) rectangle ({\cw},{(\y+1)*\ch}); \node[font=\tiny] at ({0.5*\cw},{\y*\ch+0.5*\ch}) {rows 0--7};
      \fill[statecolor!45] ({\cw},{\y*\ch}) rectangle ({2*\cw},{(\y+1)*\ch}); \node[font=\tiny] at ({1.5*\cw},{\y*\ch+0.5*\ch}) {rows 8--15};
    }
    \foreach \y in {3,2} {
      \fill[qcolor!35] (0,{\y*\ch}) rectangle ({\cw},{(\y+1)*\ch}); \node[font=\tiny] at ({0.5*\cw},{\y*\ch+0.5*\ch}) {rows 0--7};
      \fill[qcolor!35] ({\cw},{\y*\ch}) rectangle ({2*\cw},{(\y+1)*\ch}); \node[font=\tiny] at ({1.5*\cw},{\y*\ch+0.5*\ch}) {rows 8--15};
    }
    \foreach \y in {1,0} {
      \fill[vcolor!35] (0,{\y*\ch}) rectangle ({\cw},{(\y+1)*\ch}); \node[font=\tiny] at ({0.5*\cw},{\y*\ch+0.5*\ch}) {rows 8--15};
      \fill[gray!25] ({\cw},{\y*\ch}) rectangle ({2*\cw},{(\y+1)*\ch}); \node[font=\tiny, text=gray!70!black] at ({1.5*\cw},{\y*\ch+0.5*\ch}) {idle};
    }
    \foreach \i in {0,...,6} { \draw[gray!60] (0,{\i*\ch}) -- ({2*\cw},{\i*\ch}); }
    \foreach \i in {0,1,2} { \draw[gray!60] ({\i*\cw},0) -- ({\i*\cw},{6*\ch}); }
    \node[anchor=north, font=\tiny] at ({0.5*\cw},{-0.06}) {cols 0--7};
    \node[anchor=north, font=\tiny] at ({1.5*\cw},{-0.06}) {cols 8--15};
    \node[anchor=west, font=\tiny] at ({2*\cw+0.15},{5.5*\ch}) {$\mathbf{N}_4^{2}$};
    \node[anchor=west, font=\tiny] at ({2*\cw+0.15},{4.5*\ch}) {$(\mathbf{I} - \mathbf{N}_4)(\mathbf{I} + \mathbf{N}_4^{2})$};
    \node[anchor=west, font=\tiny] at ({2*\cw+0.15},{3.5*\ch}) {$\mathbf{D}^{-1}\mathbf{C}$, $4 \to 8$};
    \node[anchor=west, font=\tiny] at ({2*\cw+0.15},{2.5*\ch}) {$-(\mathbf{D}^{-1}\mathbf{C})\mathbf{A}^{-1}$, $4 \to 8$};
    \node[anchor=west, font=\tiny] at ({2*\cw+0.15},{1.5*\ch}) {$\mathbf{D}^{-1}\mathbf{C}$, $8 \to 16$};
    \node[anchor=west, font=\tiny] at ({2*\cw+0.15},{0.5*\ch}) {$-(\mathbf{D}^{-1}\mathbf{C})\mathbf{A}^{-1}$, $8 \to 16$};
  \end{scope}
  \node[anchor=north, font=\tiny] at (5.6,-0.55) {%
    \textcolor{statecolor!80!black}{\rule{0.5em}{0.5em}} step 1: $4 \times 4$ diagonal blocks, two-term series \quad
    \textcolor{qcolor}{\rule{0.5em}{0.5em}} step 2: $4 \to 8$ corrections \quad
    \textcolor{vcolor}{\rule{0.5em}{0.5em}} step 3: $8 \to 16$ correction};
\end{tikzpicture}
\caption{The blockwise inverse of one $16 \times 16$ $\mathbf{I} + \mathbf{N}$ for KDA and GDN-2. \textbf{Left}: The diagram shows the blocks each step touches. Step 1 inverts the four $4 \times 4$ diagonal blocks by the two-term series, step 2 corrects two $4 \times 4$ blocks into $8 \times 8$ inverses, and step 3 corrects one $8 \times 8$ block. \textbf{Right}: The diagram shows the six dependent MMA stages of the register-MMA warp, colored by step, with the product each stage forms.}
\label{fig:blockinv16}
\end{figure}
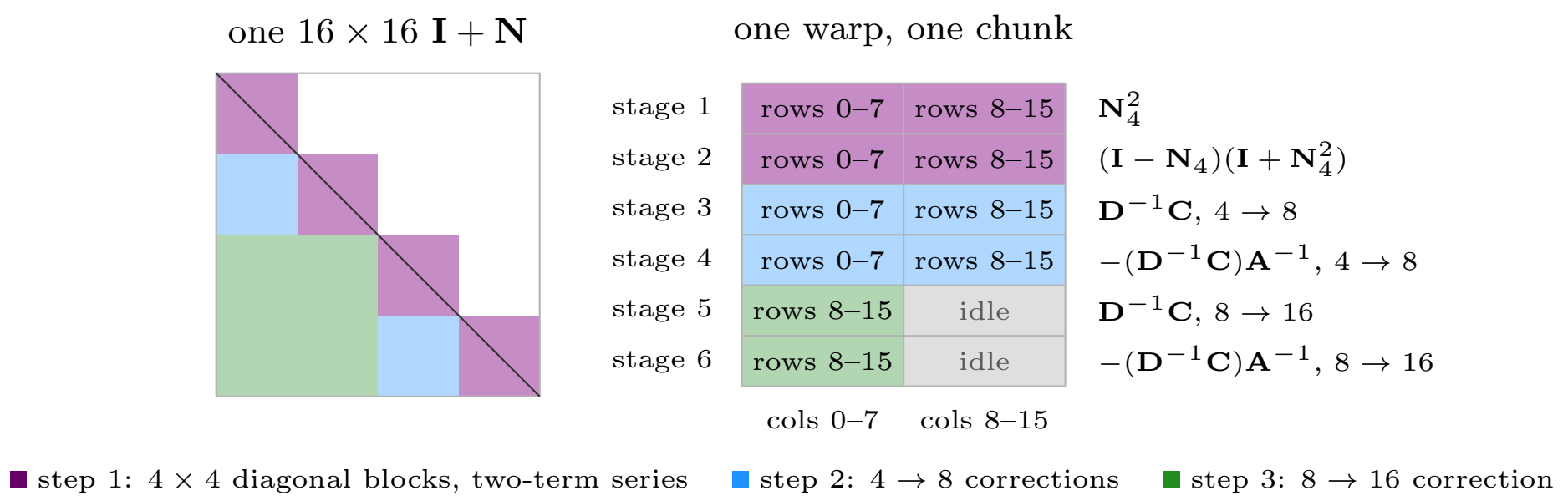

In order to avoid this issue, we combine the best of both worlds. We use the same block corrections from Section~\ref{sec:pairinv} with a shorter Neumann series for the smaller powers. The $4 \times 4$ diagonal blocks of $\mathbf{I} + \mathbf{N}$ are inverted by the two-term series $(\mathbf{I} - \mathbf{N}_4)(\mathbf{I} + \mathbf{N}_4^{2})$, with $\mathbf{N}_4$ the matching block of $\mathbf{N}$. The $4 \to 8$ and $8 \to 16$ steps are then the block corrections $\mathbf{C} \leftarrow -\mathbf{D}^{-1}\mathbf{C}\mathbf{A}^{-1}$ applied to the same register fragments. Figure~\ref{fig:blockinv16} shows the blocks each step uses and the six dependent MMA stages in which the register-MMA warp of Section~\ref{sec:regwarp} carries them out. The first four stages are full $16 \times 16$ products, while the two stages of the $8 \to 16$ correction only touch the lower-left $8 \times 8$ block and need one \texttt{mma.sync} instruction each. Thus, the blockwise inverse costs ten \texttt{mma.sync} instructions per chunk against the twelve of the series, with the same six dependent stages.

\subsection{KDA/GDN-2: Register-MMA Warp for the Small Matrices}
\label{sec:regwarp}

At $C = 16$, the forward's $\mathbf{T}_{[t]}$ (Section~\ref{sec:inverse}) and $\mathbf{A}_{[t]}$, and the backward's $d\mathbf{A}_{[t]}$ and $d\mathbf{M}_{[t]}$ are all $16 \times 16$. We separate the MMAs that create these matrices into warps of their own, where we issue the classic register-level \texttt{mma.sync} instructions with fp32 register accumulators. A dedicated register-MMA warp owns $\widehat{\mathbf{K}}_{[t]}\overline{\mathbf{K}}_{[t]}^{\trans}$ and the inverse $\mathbf{T}_{[t]}$ (plus $d\mathbf{M}_{[t]}$ in the backward), while the epilogue warp computes $\mathbf{A}_{[t]}$ (and $d\mathbf{A}_{[t]}$ in the backward) between its stores of the output. The MMAs that apply them, such as $\mathbf{Y}^{\trans}\mathbf{T}^{\trans}$ and $\mathbf{U}^{\trans}\mathbf{A}^{\trans}$, stay on the \texttt{tcgen05} MMA warp. The advantage of this split is that we can reduce register pressure between the MMA warps and not require the \texttt{tcgen05} MMA warp to also issue register-level MMA instructions.

In the backward, we treat $d\mathbf{A}_{[t]}$ with additional care, due to numerical issues. The entry of $\mathbf{A}_{[t]}$ for query row $i$ and key row $j$ of the chunk is the following.
\[
\left(\mathbf{A}_{[t]}\right)_{ij} = \sum_{c} q_{ic}\, k_{jc}\, \frac{\Gamma_{ic}}{\Gamma_{jc}} \qquad i \ge j
\]
Here, $\Gamma_{ic}$ is the shorthand for $(\boldsymbol{\Gamma}_{[t]})_{ic}$, so the ratio $\Gamma_{ic} / \Gamma_{jc} = (\mathbf{L}_{[t]})_{ijc}$ is the decay of channel $c$ between the two tokens. On the diagonal the ratio is one, so $(\mathbf{A}_{[t]})_{ii} = \mathbf{q}_i \cdot \mathbf{k}_i$ has no decay in it, and its gradient also has no decay in it.
\[
d\mathbf{q}_i \leftarrow d\mathbf{q}_i + \left(d\mathbf{A}_{[t]}\right)_{ii}\, \mathbf{k}_i \qquad
d\mathbf{k}_i \leftarrow d\mathbf{k}_i + \left(d\mathbf{A}_{[t]}\right)_{ii}\, \mathbf{q}_i
\]
Row $i$ of $d\mathbf{A}_{[t]}$ contributes to $d\mathbf{q}_i$ as $\sum_{j \le i} (d\mathbf{A}_{[t]})_{ij}\, \mathbf{k}_j\, \Gamma_{ic} / \Gamma_{jc}$, with one term per key at or before token $i$ each weighted by its decay ratio. Column $j$ of $d\mathbf{A}_{[t]}$ contributes to $d\mathbf{k}_j$ the same way against the queries at or after token $j$. The $\overline{\mathbf{K}}_{[t]}^{\trans}d\mathbf{A}_{[t]}^{\trans}$ and $\widehat{\mathbf{Q}}_{[t]}^{\trans}d\mathbf{A}_{[t]}$ MMAs compute these sums, because the operands $\widehat{\mathbf{Q}}_{[t]} = \mathbf{Q}_{[t]} \odot \boldsymbol{\Gamma}_{[t]}$ and $\overline{\mathbf{K}}_{[t]} = \mathbf{K}_{[t]} \odot \boldsymbol{\Gamma}_{[t]}^{-1}$ of Section~\ref{sec:chunking} carry one half of every ratio, and the epilogue applies the other half, $\boldsymbol{\Gamma}_{[t]}$ or $\boldsymbol{\Gamma}_{[t]}^{-1}$, to the finished sum.

Since the MMA is done in 16 bits, operands $(d\mathbf{A}_{[t]})_{ij}$ and one of the scaled key or query are 16-bit. For the diagonal term, this becomes problematic because it needs no decay ratio, so sending it through the operands would round $(d\mathbf{A}_{[t]})_{ii}$ to 16 bits and multiply it into $k_{ic} / \Gamma_{ic}$, a 16-bit number that can be orders of magnitude larger than $k_{ic}$, only for the epilogue to scale the product back down by $\Gamma_{ic}$. This can lead to numerical instability depending on the magnitude of $(d\mathbf{A}_{[t]})_{ii}$.

Thus, we keep the non-diagonal and diagonal parts separate. The epilogue warp computes $d\mathbf{A}_{[t]} = \big(d\mathbf{O}_{[t]}\mathbf{U}_{[t]}^{\trans}\big) \odot \mathbf{1}^{\ge}$ in registers after its register MMAs, and copies the sixteen diagonal entries into a small separate fp32 ring in shared memory. Then, it zeroes the corresponding entries in the $d\mathbf{A}_{[t]}$ buffer, and stores only the strict lower triangle in the 16-bit data type of the two MMAs. The gradient epilogue warps, which already hold the raw $\mathbf{q}_i$ and $\mathbf{k}_i$ rows, add $(d\mathbf{A}_{[t]})_{ii}\,\mathbf{k}_i$ to $d\mathbf{q}_i$ and $(d\mathbf{A}_{[t]})_{ii}\,\mathbf{q}_i$ to $d\mathbf{k}_i$ as they assemble the rows, so the diagonal term is exact to fp32 and does not have a round trip of scales.

\subsection{Checkpointing the State Every $n$ Tokens}
\label{sec:ckpt}

The backward pass needs the forward's per-chunk states for the backward recurrence. As such, our forward kernels support checkpointing the recurrent state $\mathbf{S}$ on a caller-chosen cadence during the forward pass. Given \texttt{checkpoint\_every\_n\_tokens} $= n$, where $n$ is any positive multiple of the chunk size (counted in sub-token rows for GDP), the kernel emits an extra output $\mathbf{S}_{\text{ckpt}}$, the checkpoint series, whose entry $\mathbf{S}_{\text{ckpt}}[j]$ is the running state $\mathbf{S}$ entering token $j \cdot n$. Entry $\mathbf{S}_{\text{ckpt}}[0]$ is the incoming state, zero unless the caller passed an initial state. The end state remains a separate output buffer.

\begin{figure}[H]
\centering
\begin{tikzpicture}[x=0.62cm, y=0.7cm, every node/.style={font=\tiny}]
    \foreach \i in {0,...,11} {
        \draw[fill=qcolor!10, draw=qcolor!60] (\i, 0) rectangle (\i+0.92, 0.7);
    }
    \node[anchor=east, font=\scriptsize] at (-0.3, 0.35) {chunks ($C$ tokens)};
    \draw[very thick, statecolor, ->] (0, 0.8) -- (0, 1.5);
    \node[fill=statecolor!15, draw=statecolor, rounded corners=1pt, minimum width=0.7cm, anchor=south, font=\tiny] at (0, 1.5) {$\mathbf{S}_{\text{ckpt}}[0]$};
    \foreach \i/\j in {3/1, 7/2} {
        \draw[very thick, statecolor, ->] (\i+0.92, 0.8) -- (\i+0.92, 1.5);
        \node[fill=statecolor!15, draw=statecolor, rounded corners=1pt, minimum width=0.7cm, anchor=south, font=\tiny] at (\i+0.92, 1.5) {$\mathbf{S}_{\text{ckpt}}[\j]$};
    }
    \draw[very thick, vcolor, ->] (11.92, -0.1) -- (11.92, -0.8);
    \node[fill=vcolor!15, draw=vcolor, rounded corners=1pt, anchor=north, font=\tiny] at (11.92, -0.8) {final state (fp32 or bf16)};
\end{tikzpicture}
\caption{Checkpointing every $n = 4C$ tokens. $\mathbf{S}_{\text{ckpt}}[0]$ is the incoming state, and the final state is a separate output.}
\label{fig:ckpt}
\end{figure}
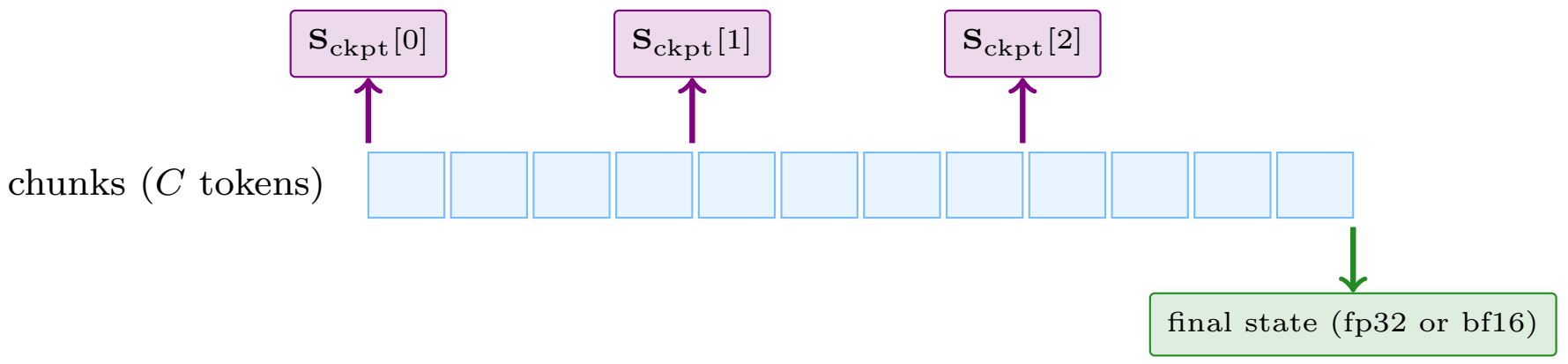

With $n = C$, this is precisely the per-chunk state series the backward kernel consumes. When the forward already ran at the per-chunk cadence, the backward path carries that series and the original inputs across. Thus, the user can choose between two settings.
\begin{itemize}
    \item \textbf{Stash} ($n = C$ at forward time): the forward pays the checkpoint stores, and the backward consumes the series directly.
    \item \textbf{Recompute} (no checkpoints at forward time): the backward first regenerates the series at $n = C$ into workspace with a dedicated state-only recompute kernel, then proceeds identically.
\end{itemize}
For training, stashing trades some extra memory for higher throughput. The recompute path exists for callers who do not have enough memory to hold the state series between the forward and backward passes, and the two paths should produce bitwise-identical gradients. Coarser cadences ($n = 2C, 4C, \ldots$) shrink the stash proportionally and serve callers who want periodic states for their own purposes like prefix caching. The backward kernel still needs the per-chunk series, so it will regenerate it under any coarser cadence.

\section{One Fused Kernel}
\label{sec:fused}

\subsection{Kernel Breakdown}
\label{sec:breakdown}

As hinted at earlier in Section~\ref{sec:prior}, FLA, FlashQLA, and FlashKDA follow a trend toward more aggressive fusion. In general, FLA runs every step of the chunked algorithm as its own kernel or kernels. FlashQLA and FlashKDA each fuse several of those steps into one warp-specialized kernel, but they still leave some as separate launches. We continue that trend and go one step further. Every step of the forward runs in one kernel, and every step of the backward in another.

Per chunk, our GDN forward kernel runs the gate scan that turns the raw logits into $\boldsymbol{\Gamma}_{[t]}$, $\boldsymbol{\Lambda}_{[t]}$, and $\gamma_{[t]}$, the $64 \times 64$ block inverse that turns $\mathbf{M}_{[t]}$ into $\mathbf{T}_{[t]}$, and seven MMAs, namely $\mathbf{K}_{[t]}\mathbf{K}_{[t]}^{\trans}$ and $\mathbf{A}_{[t]} = (\mathbf{Q}_{[t]}\mathbf{K}_{[t]}^{\trans}) \odot \mathbf{L}_{[t]}$, the state read $\mathbf{K}_{[t]}\mathbf{S}_{[t-1]}$, the inter-chunk output $\mathbf{Q}_{[t]}\mathbf{S}_{[t-1]}$, the correction $\mathbf{U}_{[t]} = \mathbf{T}_{[t]}\mathbf{Y}_{[t]}$ from $\mathbf{Y}_{[t]} = \mathbf{V}_{[t]} - (\boldsymbol{\Gamma}_{[t]} \odot \mathbf{K}_{[t]})\mathbf{S}_{[t-1]}$, the intra-chunk output $\mathbf{A}_{[t]}\mathbf{U}_{[t]}$, and the state update $\mathbf{K}_{[t]}^{\trans}\mathbf{U}'_{[t]}$, with the $\mathbf{Y}$, $\mathbf{O}$, and $\mathbf{S}$ epilogues between them. As mentioned previously, the KDA and GDN-2 kernels add the per-channel decay operands before the MMA and compute the $16 \times 16$ outputs in separate register-MMA warps inside the same kernel.

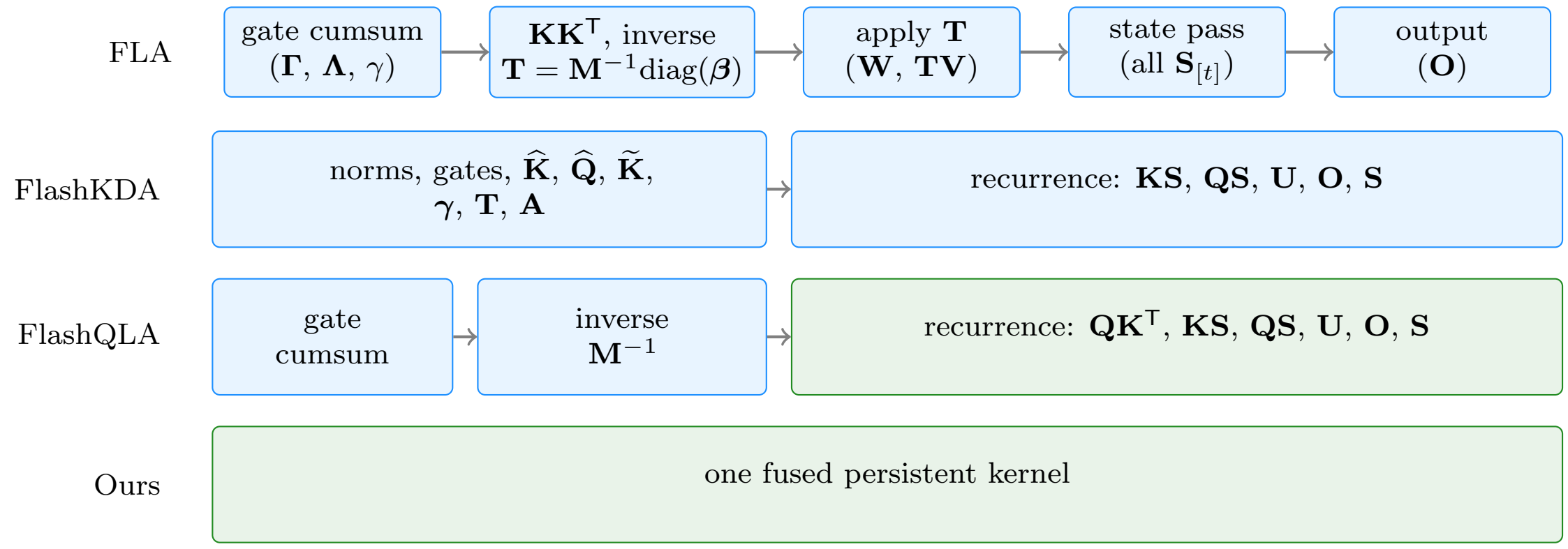
\begin{figure}[H]
\centering
\begin{tikzpicture}[
    kbox/.style={rectangle, draw=qcolor, fill=qcolor!10, rounded corners=2pt, minimum height=0.85cm, minimum width=2.05cm, font=\scriptsize, align=center, inner sep=3pt},
    fbox/.style={rectangle, draw=vcolor, fill=vcolor!10, rounded corners=2pt, minimum height=0.85cm, font=\scriptsize, align=center, inner sep=3pt},
    hbm/.style={->, thick, gray}
]
    \node[kbox] (f1) {gate cumsum\\($\boldsymbol{\Gamma}$, $\boldsymbol{\Lambda}$, $\gamma$)};
    \node[kbox, right=0.45cm of f1] (f2) {$\mathbf{K}\mathbf{K}^{\trans}$, inverse\\$\mathbf{T} = \mathbf{M}^{-1}\diag(\boldsymbol{\beta})$};
    \node[kbox, right=0.45cm of f2] (f3) {apply $\mathbf{T}$\\($\mathbf{W}$, $\mathbf{T}\mathbf{V}$)};
    \node[kbox, right=0.45cm of f3] (f4) {state pass\\(all $\mathbf{S}_{[t]}$)};
    \node[kbox, right=0.45cm of f4] (f5) {output\\($\mathbf{O}$)};
    \draw[hbm] (f1) -- (f2);
    \draw[hbm] (f2) -- (f3);
    \draw[hbm] (f3) -- (f4);
    \draw[hbm] (f4) -- (f5);
    \node[left=0.35cm of f1, font=\scriptsize, anchor=east] {FLA};

    \node[kbox, inner sep=3pt, fit={([yshift=-1.3cm]f1.west) ([yshift=-1.3cm]f2.east)}, minimum height=1.1cm] (k1) {norms, gates, $\widehat{\mathbf{K}}$, $\widehat{\mathbf{Q}}$, $\widetilde{\mathbf{K}}$,\\$\boldsymbol{\gamma}$, $\mathbf{T}$, $\mathbf{A}$};
    \node[kbox, inner sep=3pt, fit={([yshift=-1.3cm]f3.west) ([yshift=-1.3cm]f5.east)}, minimum height=1.1cm] (k2) {recurrence: $\mathbf{K}\mathbf{S}$, $\mathbf{Q}\mathbf{S}$, $\mathbf{U}$, $\mathbf{O}$, $\mathbf{S}$};
    \draw[hbm] (k1) -- (k2);
    \node[left=0.35cm of k1.west, font=\scriptsize, anchor=east] {FlashKDA};

    \node[kbox, inner sep=3pt, fit={([yshift=-2.7cm]f1.west) ([yshift=-2.7cm]f1.east)}, minimum height=1.1cm] (q1) {gate\\cumsum};
    \node[kbox, inner sep=3pt, fit={([yshift=-2.7cm]f2.west) ([yshift=-2.7cm]f2.east)}, minimum height=1.1cm] (q2) {inverse\\$\mathbf{M}^{-1}$};
    \node[fbox, inner sep=3pt, fit={([yshift=-2.7cm]f3.west) ([yshift=-2.7cm]f5.east)}, minimum height=1.1cm] (q3) {recurrence: $\mathbf{Q}\mathbf{K}^{\trans}$, $\mathbf{K}\mathbf{S}$, $\mathbf{Q}\mathbf{S}$, $\mathbf{U}$, $\mathbf{O}$, $\mathbf{S}$};
    \draw[hbm] (q1) -- (q2);
    \draw[hbm] (q2) -- (q3);
    \node[left=0.35cm of q1.west, font=\scriptsize, anchor=east] {FlashQLA};

    \node[fbox, inner sep=3pt, fit={([yshift=-4.1cm]f1.west) ([yshift=-4.1cm]f5.east)}, minimum height=1.1cm] (g1) {one fused persistent kernel};
    \node[left=0.35cm of g1.west, font=\scriptsize, anchor=east] {Ours};
\end{tikzpicture}
\caption{Forward kernel breakdown. FLA runs five steps that communicate through global memory. FlashKDA runs a chunk-parallel preparation kernel and a recurrence kernel. FlashQLA runs one fused kernel after a separate gate cumsum and inverse kernel. Ours is a single persistent kernel with the gate scan, the inverse, and every per-chunk MMA inside of it.}
\label{fig:fwd-breakdown}
\end{figure}

One CTA owns one \texttt{(batch, head)} pair, walks its chunks in sequence order, and updates the state as it moves along. $\mathbf{S}$ always lives in tensor memory (TMEM) in fp32 for the entire sequence and is updated in place. Figure~\ref{fig:fwd-breakdown} shows the kernel breakdowns for each of the implementations.

\begin{figure}[H]
\centering
\begin{tikzpicture}[
    kbox/.style={rectangle, draw=qcolor, fill=qcolor!10, rounded corners=2pt, minimum height=0.85cm, minimum width=1.5cm, font=\tiny, align=center, inner sep=2pt},
    fbox/.style={rectangle, draw=vcolor, fill=vcolor!10, rounded corners=2pt, minimum height=0.85cm, font=\scriptsize, align=center, inner sep=3pt},
    obox/.style={rectangle, draw=vcolor, fill=vcolor!5, dashed, rounded corners=2pt, minimum height=0.85cm, font=\scriptsize, align=center, inner sep=3pt},
    hbm/.style={->, thick, gray}
]
    \node[kbox] (b1) {recompute\\$\mathbf{W}$, $\mathbf{T}\mathbf{V}$};
    \node[kbox, right=0.25cm of b1] (b2) {recompute\\all $\mathbf{S}_{[t]}$};
    \node[kbox, right=0.25cm of b2] (b3) {$d\mathbf{V}$\\(intra-chunk)};
    \node[kbox, right=0.25cm of b3] (b4) {$d\mathbf{S}$ pass\\(all $d\mathbf{S}_{[t]}$, $d\mathbf{V}$)};
    \node[kbox, right=0.25cm of b4] (b5) {$d\mathbf{Q}$, $d\mathbf{K}$,\\$d\mathbf{W}$, $d\mathbf{g}$};
    \node[kbox, right=0.25cm of b5] (b6) {inverse backward\\($d\mathbf{K}$, $d\mathbf{V}$, $d\beta$, $d\mathbf{g}$)};
    \node[kbox, right=0.25cm of b6] (b7) {$d\mathbf{g}$ reverse\\cumsum};
    \draw[hbm] (b1) -- (b2);
    \draw[hbm] (b2) -- (b3);
    \draw[hbm] (b3) -- (b4);
    \draw[hbm] (b4) -- (b5);
    \draw[hbm] (b5) -- (b6);
    \draw[hbm] (b6) -- (b7);
    \node[left=0.35cm of b1, font=\scriptsize, anchor=east] {FLA};

    \node[kbox, font=\scriptsize, inner sep=2pt, fit={([xshift=2pt, yshift=-1.7cm]b1.west) ([xshift=-2pt, yshift=-1.7cm]b2.east)}, minimum height=0.75cm] (r1) {recompute\\all $\mathbf{S}_{[t]}$};
    \node[fbox, inner sep=2pt, fit={([xshift=2pt, yshift=-1.7cm]b3.west) ([xshift=-2pt, yshift=-1.7cm]b6.east)}, minimum height=0.75cm] (r2) {one fused backward kernel:\\$d\mathbf{Q}$, $d\mathbf{K}$, $d\mathbf{V}$, $d\beta$, $d\mathbf{g}$, $d\mathbf{S}_{\text{init}}$};
    \node[kbox, font=\scriptsize, inner sep=2pt, fit={([xshift=2pt, yshift=-1.7cm]b7.west) ([xshift=-2pt, yshift=-1.7cm]b7.east)}, minimum height=0.75cm] (r3) {$d\mathbf{g}$ reverse\\cumsum};
    \draw[hbm] (r1) -- (r2);
    \draw[hbm] (r2) -- (r3);
    \node[left=0.35cm of r1.west, font=\scriptsize, anchor=east] {FlashQLA};

    \node[obox, inner sep=2pt, fit={([xshift=2pt, yshift=-3.5cm]b1.west) ([xshift=-2pt, yshift=-3.5cm]b2.east)}, minimum height=1.0cm] (o1) {state series recompute\\(if recompute)};
    \node[fbox, inner sep=2pt, fit={([xshift=2pt, yshift=-3.5cm]b3.west) ([xshift=-2pt, yshift=-3.5cm]b7.east)}, minimum height=1.0cm] (o2) {one fused persistent kernel:\\recompute $\mathbf{T}$, $\mathbf{A}$, $\mathbf{Y}$, $\mathbf{U}$, then $d\mathbf{Q}$, $d\mathbf{K}$, $d\mathbf{V}$, $d\beta$, $d\mathbf{g}$, $d\mathbf{S}_{\text{init}}$ in place};
    \draw[hbm] (o1) -- (o2);
    \node[left=0.35cm of o1.west, font=\scriptsize, anchor=east] {Ours};
\end{tikzpicture}
\caption{Backward kernel breakdown. FLA runs seven steps, two of which recompute forward intermediates. FlashQLA fuses the main backward pass into one kernel with a separate state recompute kernel and a reverse cumsum kernel. Ours fuses the state recompute and all eighteen backward MMAs into a single persistent kernel. FlashKDA has no backward.}
\label{fig:bwd-breakdown}
\end{figure}
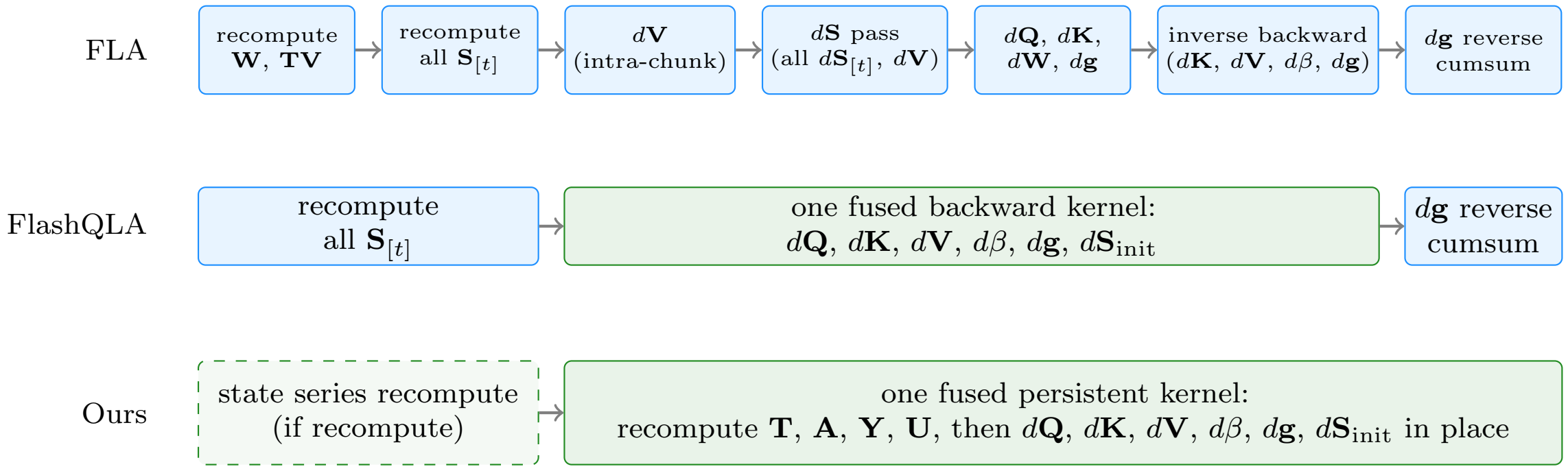

The backward kernel follows a similar design, but runs backward in time. It walks the chunks from the last to the first and carries the state gradient $d\mathbf{S}$ in TMEM in fp32. Per chunk, it first recomputes the forward intermediates in place. Then, it issues the gradient MMAs. Figure~\ref{fig:bwd-breakdown} shows the backward breakdowns for each of the implementations.

\subsection{Pipeline Details}
\label{sec:pipeline}

All the kernels are warp specialized. In this section, we discuss the roles of each warp and how we share data through shared and tensor memory, focusing on the prefill kernels. The backward kernels are warp specialized in a similar way, but we leave the details to Appendix~\ref{app:bwdpipe}.

\subsubsection{GDN and GDP}

The GDN and GDP forward runs twelve warps with the function of each warp discussed in Table~\ref{tab:warps-fwd}.

\begin{table}[H]
\centering
\small
\begin{tabular}{@{}l p{13.2cm}@{}}
\toprule
Warps & Role \\
\midrule
0--3 & compute group 0: runs the $\mathbf{K}\mathbf{K}^{\trans}$ and $\mathbf{Q}\mathbf{K}^{\trans}$ epilogues and the $64 \times 64$ blockwise inverse that yields $\mathbf{T}$ \\
4--7 & compute group 1: computes $\mathbf{Y} = \mathbf{V} - \mathbf{K}\mathbf{S}$, downcasts $\mathbf{U}$, runs the $\mathbf{O}$ epilogue, and rescales $\mathbf{S}$ \\
8 & gate warp: loads the gate logits and $\boldsymbol{\beta}$ and runs the prefix scan into $\boldsymbol{\Gamma}$, $\boldsymbol{\Lambda}$, and $\gamma$ \\
9 & TMA warp: loads the $\mathbf{Q}$, $\mathbf{K}$, and $\mathbf{V}$ tiles \\
10 & MMA warp: issues every \texttt{tcgen05} MMA \\
11 & epilogue warp: stores $\mathbf{O}$ and the state checkpoints \\
\bottomrule
\end{tabular}
\caption{Warp roles of the GDN and GDP forward kernel.}
\label{tab:warps-fwd}
\end{table}

\begin{figure}[H]
\centering
\begin{tikzpicture}[x=0.85cm, y=0.8cm, every node/.style={font=\tiny}]
    \draw[dashed, gray] (0, -0.6) -- (0, 2.6);
    \draw[dashed, gray] (17.5, -0.6) -- (17.5, 2.6);
    \node[anchor=south, font=\scriptsize, text=gray] at (8.75, 2.6) {one pair iteration (chunks $t$ and $t{+}1$)};
    \node[anchor=east, font=\scriptsize\itshape] at (-0.25, 2) {CG0};
    \node[anchor=east, font=\scriptsize\itshape] at (-0.25, 1) {CG1};
    \node[anchor=east, font=\scriptsize\itshape] at (-0.25, 0) {MMA};
    \draw[fill=red!10, draw=red!40, dashed] (0,-0.32) rectangle (1.1,0.32);   \node[text=red!70!black, align=center] at (0.55,0) {wait\\for $\mathbf{S}$};
    \draw[fill=qcolor!15, draw=qcolor] (1.1,-0.32) rectangle (2.5,0.32);   \node at (1.8,0)   {$\mathbf{S}^{\trans}\mathbf{K}^{\trans}$};
    \draw[fill=qcolor!15, draw=qcolor] (2.5,-0.32) rectangle (3.9,0.32);   \node at (3.2,0)   {$\mathbf{S}^{\trans}\mathbf{Q}^{\trans}$};
    \draw[fill=qcolor!15, draw=qcolor] (3.9,-0.32) rectangle (4.9,0.32);   \node at (4.4,0)   {$\mathbf{Y}^{\trans}\mathbf{T}^{\trans}$};
    \draw[fill=gray!20, draw=gray]     (4.9,-0.32) rectangle (6.5,0.32);   \node[align=center] at (5.7,0) {$[\mathbf{K};\mathbf{Q}]\mathbf{K}^{\trans}$\\$t{+}2$};
    \draw[fill=qcolor!15, draw=qcolor] (6.5,-0.32) rectangle (7.7,0.32);   \node at (7.1,0)   {$\mathbf{U}^{\trans}\mathbf{A}^{\trans}$};
    \draw[fill=qcolor!15, draw=qcolor] (7.7,-0.32) rectangle (8.7,0.32);   \node at (8.2,0)   {$\mathbf{U}'^{\trans}\mathbf{K}$};
    \draw[fill=gray!20, draw=gray]     (8.7,-0.32) rectangle (9.9,0.32);   \node[align=center] at (9.3,0) {$[\mathbf{K};\mathbf{Q}]\mathbf{K}^{\trans}$\\$t{+}3$};
    \draw[fill=scolor!20, draw=scolor] (9.9,-0.32) rectangle (11.3,0.32);  \node at (10.6,0)  {$\mathbf{S}^{\trans}\mathbf{K}^{\trans}$};
    \draw[fill=scolor!20, draw=scolor] (11.3,-0.32) rectangle (12.7,0.32); \node at (12.0,0)  {$\mathbf{S}^{\trans}\mathbf{Q}^{\trans}$};
    \draw[fill=scolor!20, draw=scolor] (12.7,-0.32) rectangle (13.7,0.32); \node at (13.2,0)  {$\mathbf{Y}^{\trans}\mathbf{T}^{\trans}$};
    \draw[fill=red!10, draw=red!40, dashed] (13.7,-0.32) rectangle (15.3,0.32); \node[text=red!70!black] at (14.5,0) {wait for $\mathbf{U}$};
    \draw[fill=scolor!20, draw=scolor] (15.3,-0.32) rectangle (16.5,0.32); \node at (15.9,0)  {$\mathbf{U}^{\trans}\mathbf{A}^{\trans}$};
    \draw[fill=scolor!20, draw=scolor] (16.5,-0.32) rectangle (17.5,0.32); \node at (17.0,0)  {$\mathbf{U}'^{\trans}\mathbf{K}$};
    \draw[fill=qcolor!15, draw=qcolor] (0,0.68) rectangle (1.1,1.32);      \node[align=center] at (0.55,1)  {$\mathbf{S}$\\epilogue};
    \draw[fill=qcolor!15, draw=qcolor] (2.5,0.68) rectangle (3.9,1.32);    \node[align=center] at (3.2,1)   {$\mathbf{V} - \mathbf{K}\mathbf{S}$\\epilogue};
    \draw[fill=qcolor!15, draw=qcolor] (3.9,0.68) rectangle (5.3,1.32);    \node at (4.6,1)   {$\mathbf{O}$ scale};
    \draw[fill=qcolor!15, draw=qcolor] (5.3,0.68) rectangle (6.5,1.32);    \node[align=center] at (5.9,1)   {$\mathbf{U}$\\downcast};
    \draw[fill=qcolor!15, draw=qcolor] (6.5,0.68) rectangle (7.3,1.32);    \node at (6.9,1)   {$\mathbf{U}'$};
    \draw[fill=qcolor!15, draw=qcolor] (7.7,0.68) rectangle (8.8,1.32);    \node[align=center] at (8.25,1)  {$\mathbf{O}$\\epilogue};
    \draw[fill=scolor!20, draw=scolor] (8.8,0.68) rectangle (9.9,1.32);    \node[align=center] at (9.35,1)  {$\mathbf{S}$\\epilogue};
    \draw[fill=scolor!20, draw=scolor] (11.3,0.68) rectangle (12.7,1.32);  \node[align=center] at (12.0,1)  {$\mathbf{V} - \mathbf{K}\mathbf{S}$\\epilogue};
    \draw[fill=scolor!20, draw=scolor] (12.7,0.68) rectangle (14.1,1.32);  \node at (13.4,1)  {$\mathbf{O}$ scale};
    \draw[fill=scolor!20, draw=scolor] (14.1,0.68) rectangle (15.3,1.32);  \node[align=center] at (14.7,1)  {$\mathbf{U}$\\downcast};
    \draw[fill=scolor!20, draw=scolor] (15.3,0.68) rectangle (16.1,1.32);  \node at (15.7,1)  {$\mathbf{U}'$};
    \draw[fill=scolor!20, draw=scolor] (16.5,0.68) rectangle (17.5,1.32);  \node[align=center] at (17.0,1)  {$\mathbf{O}$\\epilogue};
    \draw[fill=statecolor!25, draw=statecolor] (0,2.02) rectangle (1.9,2.32);    \node at (0.95,2.17) {steps 1--4, $t$};
    \draw[fill=statecolor!40, draw=statecolor] (0,1.68) rectangle (1.9,1.98);    \node at (0.95,1.83) {steps 1--4, $t{+}1$};
    \draw[fill=statecolor!25, draw=statecolor] (1.9,2.02) rectangle (3.9,2.32);  \node at (2.9,2.17) {step 5, $t$};
    \draw[fill=statecolor!40, draw=statecolor] (1.9,1.68) rectangle (3.9,1.98);  \node at (2.9,1.83) {step 5, $t{+}1$};
    \draw[fill=scolor!20, draw=scolor] (3.9,2.02) rectangle (6.5,2.32);        \node at (5.2,2.17) {$\mathbf{Q}\mathbf{K}^{\trans}$ epilogue $t{+}1$};
    \draw[fill=qcolor!15, draw=qcolor] (3.9,1.68) rectangle (6.5,1.98);        \node at (5.2,1.83) {$\mathbf{Q}\mathbf{K}^{\trans}$ epilogue $t$};
    \draw[fill=gray!20, draw=gray] (6.5,1.68) rectangle (9.9,2.32);            \node[align=center] at (8.2,2)   {mask\\$t{+}2$, $t{+}3$};
    \draw[fill=gray!20, draw=gray] (9.9,2.02) rectangle (12.5,2.32);           \node at (11.2,2.17) {$\mathbf{K}\mathbf{K}^{\trans}$ epilogue $t{+}2$};
    \draw[fill=gray!20, draw=gray] (9.9,1.68) rectangle (12.5,1.98);           \node at (11.2,1.83) {$\mathbf{K}\mathbf{K}^{\trans}$ epilogue $t{+}3$};
    \draw[fill=statecolor!25, draw=statecolor] (12.5,2.02) rectangle (17.5,2.32); \node at (15.0,2.17) {steps 1--4, W0--1, $t{+}2$};
    \draw[fill=statecolor!40, draw=statecolor] (12.5,1.68) rectangle (17.5,1.98); \node at (15.0,1.83) {steps 1--4, W2--3, $t{+}3$};
\end{tikzpicture}
\caption{The GDN and GDP forward pipeline over a chunk pair. Blue is chunk $t$, orange chunk $t{+}1$, and gray the next pair. Purple is the blockwise inverse of Section~\ref{sec:pairinv}, labeled with its steps. The MMA warp issues the next pair's two fused $[\mathbf{K};\mathbf{Q}]\mathbf{K}^{\trans}$ MMAs after the $\mathbf{U}$ MMA of chunk $t$ and after its state update, where it would otherwise wait for the $\mathbf{U}$ downcast and for the $\mathbf{S}$ epilogue that restages the state. Compute group 0 works one pair ahead, as it is on the critical path.}
\label{fig:pipeline}
\end{figure}
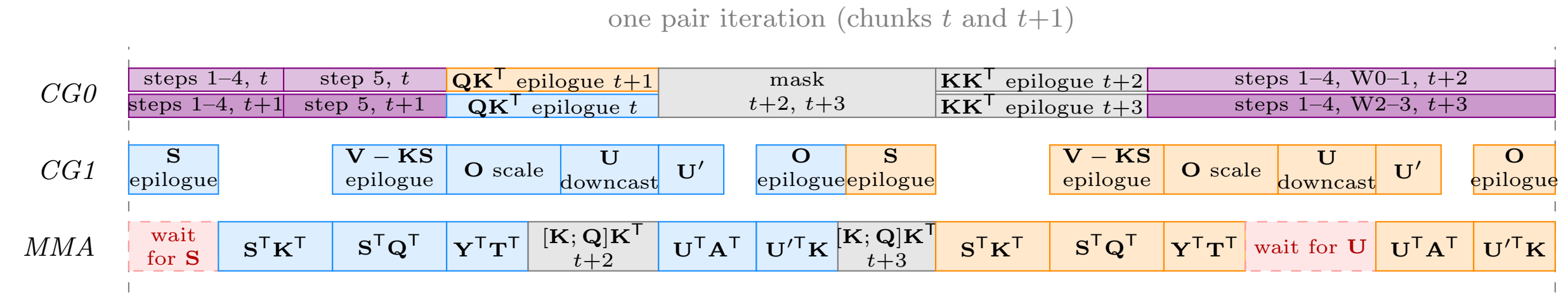

Using this warp configuration, we settled on the pipeline shown below. In general, the ultimate goal of the pipeline is to keep the MMA warp busy and hide any elementwise operations behind it. Figure~\ref{fig:pipeline} shows the schedule of the GDN and GDP forward over a chunk pair, with the improvements discussed in the previous sections, in particular the pairing of inverse computations from Section~\ref{sec:pairinv}.

In a straightforward schedule that handles one chunk at a time, the compute group that inverts $\mathbf{M}_{[t]}$ sits on the critical path, and the MMA warp waits for $\mathbf{T}_{[t]}$ and then for $\mathbf{U}_{[t]}$ in every chunk. Therefore, as we discussed previously, we run chunks in pairs to run the inverse more efficiently, and we only assign compute group 0 to that work. The two halves of compute group 0 invert the two chunks' $\mathbf{M}$ concurrently. Note that this also allows the MMA warp to fill the gaps with the next pair's fused $[\mathbf{K};\mathbf{Q}]\mathbf{K}^{\trans}$ MMAs.

\subsubsection{KDA and GDN-2}

The KDA and GDN-2 forward runs sixteen warps with their functions discussed in Table~\ref{tab:warps-fwd-kda}. The deviation from GDN is motivated by the extra scaling that needs to happen prior to the MMAs and by the register MMAs that the smaller chunk size calls for, as described in Section~\ref{sec:regwarp}.

\begin{table}[H]
\centering
\small
\begin{tabular}{@{}l p{13.2cm}@{}}
\toprule
Warps & Role \\
\midrule
0--7 & compute group 0, two ping-pong warpgroups: runs the per-channel gate prefix scan, computes the decay-folded operands $\widehat{\mathbf{Q}}$, $\widehat{\mathbf{K}}$, $\overline{\mathbf{K}}$, and $\widetilde{\mathbf{K}}$, and runs the $\mathbf{S}$ epilogue for the first half of the key channels \\
8--11 & compute group 1: stages $\mathbf{Y}$ and $\mathbf{U}$, runs the $\mathbf{S}$ epilogue for the second half of the key channels, and runs the $\mathbf{O}$ epilogue \\
12 & register-MMA warp: computes $\widehat{\mathbf{K}}\overline{\mathbf{K}}^{\trans}$ and the blockwise inverse that yields $\mathbf{T}$ \\
13 & MMA warp: issues every \texttt{tcgen05} MMA \\
14 & TMA warp: loads the $\mathbf{Q}$, $\mathbf{K}$, $\mathbf{V}$, and gate tiles and stages $\boldsymbol{\beta}$ one chunk ahead \\
15 & epilogue warp: computes $\mathbf{A}$ on register MMAs and stores $\mathbf{O}$ and the state checkpoints \\
\bottomrule
\end{tabular}
\caption{Warp roles of the KDA and GDN-2 forward kernel.}
\label{tab:warps-fwd-kda}
\end{table}

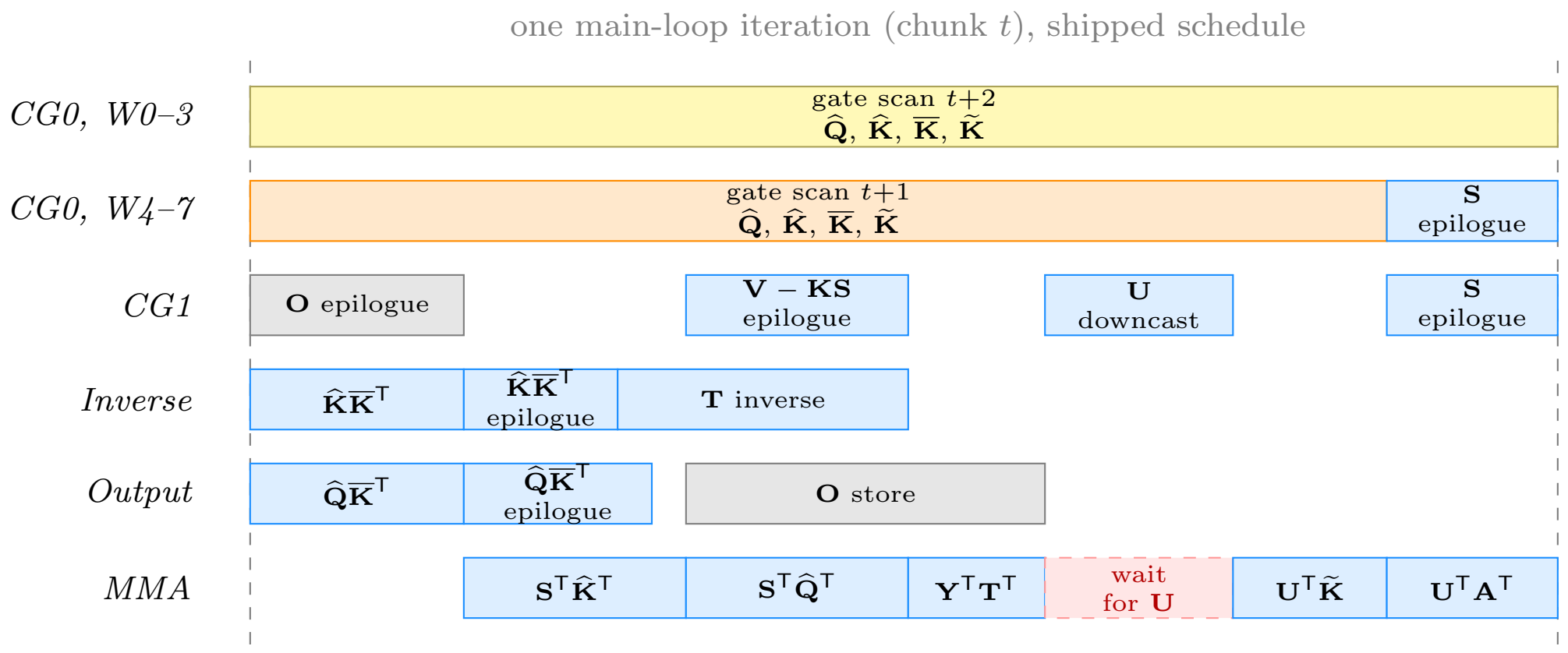
\begin{figure}[H]
\centering
\begin{tikzpicture}[x=1.45cm, y=0.8cm, every node/.style={font=\tiny}]
    \draw[dashed, gray] (4.05, -0.6) -- (4.05, 5.6);
    \draw[dashed, gray] (11.7, -0.6) -- (11.7, 5.6);
    \node[anchor=south, font=\scriptsize, text=gray] at (7.9, 5.6) {one main-loop iteration (chunk $t$), shipped schedule};
    \node[anchor=east, font=\scriptsize\itshape] at (3.8, 5) {CG0, W0--3};
    \node[anchor=east, font=\scriptsize\itshape] at (3.8, 4) {CG0, W4--7};
    \node[anchor=east, font=\scriptsize\itshape] at (3.8, 3) {CG1};
    \node[anchor=east, font=\scriptsize\itshape] at (3.8, 2) {Inverse};
    \node[anchor=east, font=\scriptsize\itshape] at (3.8, 1) {Output};
    \node[anchor=east, font=\scriptsize\itshape] at (3.8, 0) {MMA};
    \draw[fill=yellow!35, draw=yellow!60!black] (4.05,4.68) rectangle (11.7,5.32); \node[align=center] at (7.875,5)  {gate scan $t{+}2$\\$\widehat{\mathbf{Q}}$, $\widehat{\mathbf{K}}$, $\overline{\mathbf{K}}$, $\widetilde{\mathbf{K}}$};
    \draw[fill=scolor!20, draw=scolor] (4.05,3.68) rectangle (10.7,4.32); \node[align=center] at (7.375,4) {gate scan $t{+}1$\\$\widehat{\mathbf{Q}}$, $\widehat{\mathbf{K}}$, $\overline{\mathbf{K}}$, $\widetilde{\mathbf{K}}$};
    \draw[fill=qcolor!15, draw=qcolor] (10.7,3.68) rectangle (11.7,4.32); \node[align=center] at (11.2,4)   {$\mathbf{S}$\\epilogue};
    \draw[fill=gray!20, draw=gray]     (4.05,2.68) rectangle (5.3,3.32);  \node at (4.675,3)  {$\mathbf{O}$ epilogue};
    \draw[fill=qcolor!15, draw=qcolor] (6.6,2.68) rectangle (7.9,3.32);   \node[align=center] at (7.25,3)   {$\mathbf{V} - \mathbf{K}\mathbf{S}$\\epilogue};
    \draw[fill=qcolor!15, draw=qcolor] (8.7,2.68) rectangle (9.8,3.32);   \node[align=center] at (9.25,3)   {$\mathbf{U}$\\downcast};
    \draw[fill=qcolor!15, draw=qcolor] (10.7,2.68) rectangle (11.7,3.32); \node[align=center] at (11.2,3)   {$\mathbf{S}$\\epilogue};
    \draw[fill=qcolor!15, draw=qcolor] (4.05,1.68) rectangle (5.3,2.32);  \node at (4.675,2)  {$\widehat{\mathbf{K}}\overline{\mathbf{K}}^{\trans}$};
    \draw[fill=qcolor!15, draw=qcolor] (5.3,1.68) rectangle (6.2,2.32);   \node[align=center] at (5.75,2)   {$\widehat{\mathbf{K}}\overline{\mathbf{K}}^{\trans}$\\epilogue};
    \draw[fill=qcolor!15, draw=qcolor] (6.2,1.68) rectangle (7.9,2.32);   \node at (7.05,2)   {$\mathbf{T}$ inverse};
    \draw[fill=qcolor!15, draw=qcolor] (4.05,0.68) rectangle (5.3,1.32);  \node at (4.675,1)  {$\widehat{\mathbf{Q}}\overline{\mathbf{K}}^{\trans}$};
    \draw[fill=qcolor!15, draw=qcolor] (5.3,0.68) rectangle (6.4,1.32);   \node[align=center] at (5.85,1)   {$\widehat{\mathbf{Q}}\overline{\mathbf{K}}^{\trans}$\\epilogue};
    \draw[fill=gray!20, draw=gray]     (6.6,0.68) rectangle (8.7,1.32);   \node at (7.65,1)   {$\mathbf{O}$ store};
    \draw[fill=qcolor!15, draw=qcolor] (5.3,-0.32) rectangle (6.6,0.32);  \node at (5.95,0)   {$\mathbf{S}^{\trans}\widehat{\mathbf{K}}^{\trans}$};
    \draw[fill=qcolor!15, draw=qcolor] (6.6,-0.32) rectangle (7.9,0.32);  \node at (7.25,0)   {$\mathbf{S}^{\trans}\widehat{\mathbf{Q}}^{\trans}$};
    \draw[fill=qcolor!15, draw=qcolor] (7.9,-0.32) rectangle (8.7,0.32);  \node at (8.3,0)    {$\mathbf{Y}^{\trans}\mathbf{T}^{\trans}$};
    \draw[fill=red!10, draw=red!40, dashed] (8.7,-0.32) rectangle (9.8,0.32); \node[text=red!70!black, align=center] at (9.25,0) {wait\\for $\mathbf{U}$};
    \draw[fill=qcolor!15, draw=qcolor] (9.8,-0.32) rectangle (10.7,0.32);  \node at (10.25,0)  {$\mathbf{U}^{\trans}\widetilde{\mathbf{K}}$};
    \draw[fill=qcolor!15, draw=qcolor] (10.7,-0.32) rectangle (11.7,0.32); \node at (11.2,0)   {$\mathbf{U}^{\trans}\mathbf{A}^{\trans}$};
\end{tikzpicture}
\caption{The KDA and GDN-2 forward pipeline over a single main-loop iteration. Blue is chunk $t$, gray chunk $t{-}1$, orange chunk $t{+}1$, and yellow chunk $t{+}2$, and the red dashed box is a wait of the MMA warp. Compute group 0 is drawn as its two ping-pong groups of four warps, which alternate chunks. In this iteration, warps 4 to 7 finish chunk $t{+}1$'s gate scan and decay-folded operands, and warps 0 to 3 run chunk $t{+}2$'s.}
\label{fig:pipeline-kda}
\end{figure}

Figure~\ref{fig:pipeline-kda} shows one main-loop iteration of the KDA and GDN-2 forward. We should note three main differences compared to the GDN schedule. First, compute group 0 has eight warps in two ping-pong groups that alternate chunks and work one to two chunks ahead, since the per-channel gate scan and constructing the decay-folded operands are on the critical path and block any MMAs from that iteration from starting. Second, as we alluded to previously, the $16 \times 16$ matrices $\mathbf{T}$ and $\mathbf{A}$ are created in separate register-MMA warps, while the MMAs that apply them stay on the MMA warp. Third, to decrease register pressure, we split the $\mathbf{S}$ epilogue between the two compute groups, rather than leaving all of it to compute group 1 as in GDN and GDP.

\section{Filling the GPU}
\label{sec:regimes}

While fusion provides many benefits, it does present its own challenges. A unique challenge of linear attention is that the grid size is rather small. A standard quadratic attention implementation, like \cudnn{} or \textsc{FlashAttention-4}~\cite{zadouri2026flashattention4}, parallelizes over query tiles, because every output tile of softmax attention is independent of the others. Thus, each CTA fixes one $\mathbf{Q}$ tile and runs a KV loop that streams the $\mathbf{K}$ and $\mathbf{V}$ tiles of the whole sequence over it. Therefore, the grid has $B \cdot H \cdot s / C$ CTAs, where $C$ is the tile size, and easily fills the SMs. A fused linear attention kernel cannot do this. Given the serialization of the recurrence, one CTA must walk all the chunks of a \texttt{(batch, head)} pair in order. Its loop is a QKV loop that consumes a fresh $\mathbf{Q}$, $\mathbf{K}$, and $\mathbf{V}$ tile at every step, unlike quadratic attention with a fixed $\mathbf{Q}$, as Figure~\ref{fig:loops} contrasts.

\begin{figure}[H]
\centering
\begin{minipage}[t]{0.48\textwidth}
\centering
\textbf{Quadratic Attention}
\begin{lstlisting}
# grid = B * H * (s / tile)
i = blockIdx  # one Q tile
# KV loop
for j in range(s_kv / tile):
    S = Q[i] @ K[j].T
    O[i] += softmax(S) @ V[j]
\end{lstlisting}
\end{minipage}\hfill
\begin{minipage}[t]{0.48\textwidth}
\centering
\textbf{Linear Attention}
\begin{lstlisting}
# grid = B * H
S = 0
# QKV loop
for t in range(s / C):
    O[t] = intra(Q, K, V, S, t)
         + inter(Q, S, t)
    S = update(S, K, V, t)
\end{lstlisting}
\end{minipage}
\caption{The loop structure of quadratic attention and fused linear attention.}
\label{fig:loops}
\end{figure}

For linear attention, the natural grid is one CTA per \texttt{(batch, head)}, so it does not scale with sequence length. Unfused approaches can better make use of the GPU because chunk-parallel parts can launch up to $B \cdot H \cdot s / C$ CTAs like in quadratic attention. However, for our fully fused approach, we can only use $B \cdot H$ CTAs. For a GB200 GPU with 152 SMs, a layer with $B \cdot H = 16$ occupies 16 of them and leaves $(152 - 16) / 152 \approx 89\%$ of the machine idle. Furthermore, even if the number of CTAs filled the SMs, we would have far fewer waves, leading to more serious wave quantization issues. In this section, we will attempt to address both of these issues.

If the GPU is underfilled, we would like to split each tile into multiple tiles of work. Cutting every tile into $P$ segments of work produces a grid of $P \cdot B \cdot H$ CTAs, and that grid still fits in one wave only if $P \cdot B \cdot H \le \#\text{SMs}$. Thus, the number of segments we can afford per tile is defined by the following quantity.
\[
P \;=\; \left\lfloor \frac{\#\text{SMs}}{B \cdot H} \right\rfloor
\]
$P$ is a pure function of the shapes, so the plan is fixed when the graph is built. As we will see, the best strategy differs with $P$, so this section discusses three remedies, one for each range of $P$. First, intra-card context parallelism (intra-card CP) cuts a long sequence into segments, summarizes every segment exactly, chains the segment states with a small kernel, and runs the fused kernel again from the corrected states. For the prefill pass, its summary pass costs about as much as the fused kernel itself, so it pays off only from three segments per tile on. Second, split-$d_v$ splits the tile along the value head dimension. We assign two CTAs per \texttt{(batch, head)} pair, with each owning half of the value head columns. Finally, split-K cuts the sequence only where the decay gates allow it and needs no chain, so it remains available when there is no spare CTA for an even split.

All three descend from FlashQLA~\cite{qwen2026flashqla}, with some adaptations. FlashQLA's intra-card context parallelism of Section~\ref{sec:fqla} combines an exact chain with a gate-driven skip in one scheme. We take it apart into its two halves and use each alone for different $P$. The exact chain, with no warmup and no skipping, becomes our intra-card CP, and the gate-driven skip, with no chain, becomes our split-K. We also extend the intra-card CP and split-$d_v$ to all of GDN, GDP, KDA, and GDN-2. The split-K method is limited to GDN and GDP only, for reasons we make precise in Section~\ref{sec:splitk}. These three remedies apply when the tiles do not fill the machine (i.e., $P \ge 1$). When the tiles outnumber the SMs and $P < 1$, the tail method of Section~\ref{sec:waves} runs the full waves as they are and applies the same remedies only to the leftover tiles of the partial last wave, in the spirit of Stream-K~\cite{osama2023streamk}. Table~\ref{tab:plans} lists which strategy we use for every value of $P$.

\begin{table}[H]
\centering
\footnotesize
\begin{tabular}{@{}l@{\hspace{1em}}l@{\hspace{1em}}l@{}}
\toprule
$P$ & forward & backward \\
\midrule
$\ge 3$ & intra-card CP (Section~\ref{sec:chain}) & intra-card CP \\
$2$, $d_v = 128$ & split-$d_v$ (Section~\ref{sec:dvsplit}) & intra-card CP \\
$2$, $d_v = 64$ & split-K (Section~\ref{sec:splitk}) & intra-card CP \\
$1$ & split-K & split-K \\
$< 1$ & tail (Section~\ref{sec:waves}) & same \\
\bottomrule
\end{tabular}
\caption{The plan per number of segments $P$. Split-K applies to GDN and GDP only, and KDA and GDN-2 run uncut where it is listed.}
\label{tab:plans}
\end{table}

\subsection{$P \ge 3$: Intra-Card CP}
\label{sec:chain}

Note that every recurrence in the linear attention family is affine in the state. This allows us to easily cut along the sequence length. We can write the state update as follows.
\[
\mathbf{S}_t = \mathbf{F}_t \mathbf{S}_{t-1} + \mathbf{k}_t^\trans \tilde{\mathbf{v}}_t \qquad
\mathbf{F}_t = \left(\mathbf{I} - \mathbf{k}_t^\trans\left(\boldsymbol{\beta}_t \odot \mathbf{k}_t\right)\right)\diag\left(\boldsymbol{\alpha}_t\right)
\]
Here, $\tilde{\mathbf{v}}_t$ is the written value ($\beta_t \mathbf{v}_t$ for GDN, GDP, and KDA, $\mathbf{w}_t \odot \mathbf{v}_t$ for GDN-2). Let $\mathbf{S}_t$ be the state at token $t$. Then, over a span of tokens $(a, b]$, $\mathbf{S}_b$ can be written as an affine function of the state entering the span $\mathbf{S}_a$.
\[
\mathbf{S}_b = \mathbf{M}\,\mathbf{S}_a + \mathbf{H} \qquad \mathbf{M} = \mathbf{F}_b \mathbf{F}_{b-1} \cdots \mathbf{F}_{a+1} \in \mathbb{R}^{d_{qk} \times d_{qk}}
\]
Here, $\mathbf{H}$ is the state the span builds from an initial zero state. Note that neither $\mathbf{M}$ nor $\mathbf{H}$ depends on $\mathbf{S}_a$, so the spans of a sequence are independent of one another. Furthermore, $\mathbf{M}$ can also be produced by the recurrence itself by setting the initial state to $\mathbf{S} = \mathbf{I}$ and feeding $\mathbf{V} = \mathbf{0}$. Intra-card CP cuts the sequence of every tile into $P$ segments and decomposes into three stages: the summaries, the state chain, and the main kernel.

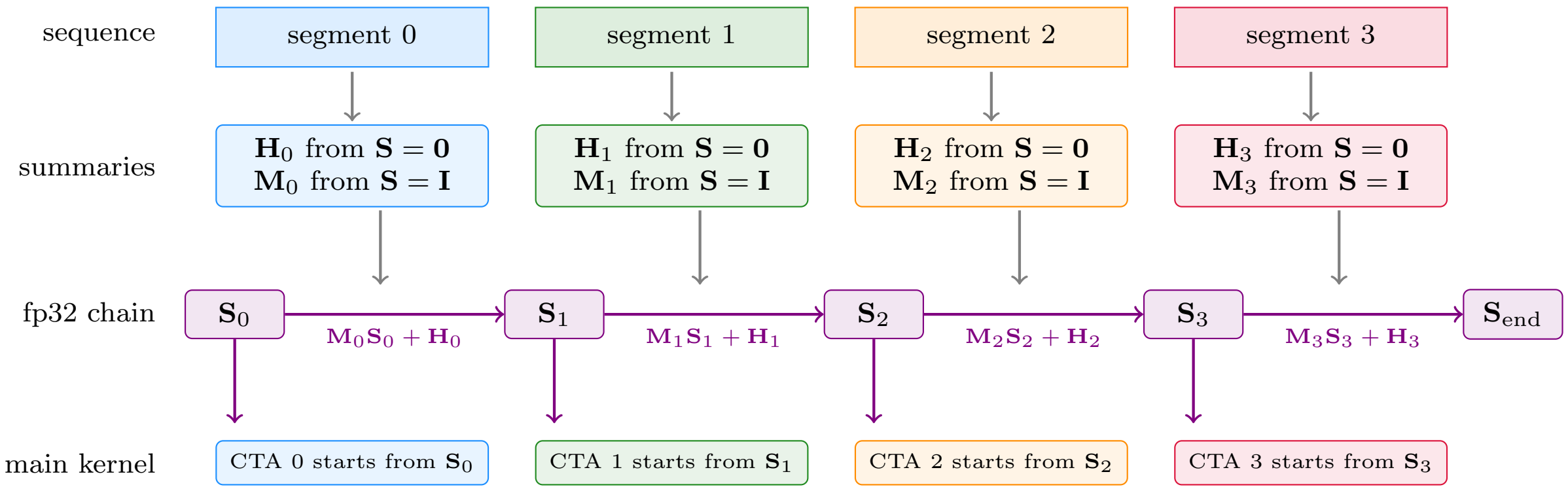
\begin{figure}[H]
\centering
\begin{tikzpicture}[x=0.95cm, y=1cm, every node/.style={font=\scriptsize}]
  \node[anchor=east] at (-0.5, 4.3) {sequence};
  \node[anchor=east] at (-0.5, 3.0) {summaries};
  \node[anchor=east] at (-0.5, 1.5) {fp32 chain};
  \node[anchor=east] at (-0.5, 0.0) {main kernel};
  \foreach \j/\c in {0/qcolor, 1/vcolor, 2/scolor, 3/kcolor} {
    \draw[fill=\c!15, draw=\c] (3.4*\j, 4.0) rectangle (3.4*\j+2.9, 4.6);
    \node at (3.4*\j+1.45, 4.3) {segment $\j$};
    \draw[->, thick, gray] (3.4*\j+1.45, 3.95) -- (3.4*\j+1.45, 3.45);
    \node[draw=\c, fill=\c!10, rounded corners=2pt, align=center, minimum width=2.75cm, minimum height=0.8cm] at (3.4*\j+1.45, 3.0) {$\mathbf{H}_{\j}$ from $\mathbf{S} = \mathbf{0}$ \\ $\mathbf{M}_{\j}$ from $\mathbf{S} = \mathbf{I}$};
    \draw[->, thick, gray] (3.4*\j+1.75, 2.55) -- (3.4*\j+1.75, 1.8);
    \node[draw=statecolor, fill=statecolor!10, rounded corners=2pt, minimum width=1.0cm] (x\j) at (3.4*\j+0.2, 1.5) {$\mathbf{S}_{\j}$};
    \draw[->, thick, statecolor] (x\j) -- (3.4*\j+0.2, 0.4);
    \node[draw=\c, fill=\c!10, rounded corners=2pt, minimum width=2.75cm, minimum height=0.42cm, font=\tiny] at (3.4*\j+1.45, 0.0) {CTA $\j$ starts from $\mathbf{S}_{\j}$};
  }
  \node[draw=statecolor, fill=statecolor!10, rounded corners=2pt, minimum width=1.0cm] (x4) at (13.8, 1.5) {$\mathbf{S}_{\text{end}}$};
  \foreach \j/\k in {0/1, 1/2, 2/3, 3/4} {
    \draw[->, thick, statecolor] (x\j) -- node[below, font=\tiny] {$\mathbf{M}_{\j}\mathbf{S}_{\j} + \mathbf{H}_{\j}$} (x\k);
  }
\end{tikzpicture}
\caption{Intra-card CP in the forward pass. Every segment is summarized from a zero initial state and from an identity initial state, the summaries are chained in fp32 to recover each segment's true incoming state $\mathbf{S}_j$, and the main kernel then runs the segments as independent sequences.}
\label{fig:chain}
\end{figure}

\subsubsection{Summaries}

First, we execute a fused summary kernel, with one CTA per segment, that computes both $(\mathbf{H}_j, \mathbf{M}_j)$ for every segment $j$. This amounts to the state-only recompute kernel of Section~\ref{sec:ckpt} carrying two states side by side on one stream of $\mathbf{K}$, gates, and $\beta$. Chain $\mathbf{H}$ uses the true $\mathbf{V}$ and an initial state of zeros, while chain $\mathbf{M}$ feeds zeros for $\mathbf{V}$ and an initial identity state.

Note that the two chains share everything that does not depend on the state, namely the loads, the gate scan, and the inverse of $\mathbf{M}_{[t]}$. Only the MMAs that depend on the state are not shared. For GDN and GDP, we choose to precompute $\mathbf{T}_{[t]}$ in a separate pre-processing kernel before the fused summary kernel to lessen TMEM pressure and make space for the two chains to run concurrently. The main kernel later loads the same $\mathbf{T}_{[t]}$ instead of inverting again. KDA and GDN-2 compute their $16 \times 16$ inverse on the register-MMA warp of Section~\ref{sec:regwarp} and do not have as much TMEM pressure. Furthermore, for KDA and GDN-2, the inverse is very small relative to the other MMAs due to the smaller chunk size, so we do not find it beneficial to precompute $\mathbf{T}_{[t]}$.

\subsubsection{State Chain}

Once these summaries have been computed, a small chain kernel walks the segments in order, accumulating the states in fp32. We choose to use fp32 for the states to maintain the numerical precision. We observe that using smaller bitwidths leads to unrecoverable accuracy degradation. We follow the recurrence below.
\[
\mathbf{S}_0 = \mathbf{S}_{\text{init}} \qquad \mathbf{S}_{j+1} = \mathbf{M}_j \mathbf{S}_j + \mathbf{H}_j
\]
Every value column of the state chains independently. Thus, for better utilization, the chain kernel cuts the $d_v$ columns into $n_s$ slices, and launches one CTA per \texttt{(batch, head)} pair and slice. For each CTA, a step is then a $d_{qk} \times d_{qk}$ by $d_{qk} \times d_v / n_s$ product of fused multiply-adds.

\subsubsection{Main Kernel}

Finally, we run the fused prefill described in Section~\ref{sec:fused} on each segment as if it were its own sequence with initial state $\mathbf{S}_j$, as we have now calculated the corrected state at the start of each segment. The summary kernel does about a prefill's worth of work per token, so a chain of $P$ segments does roughly twice the arithmetic of the uncut run, but it is spread over more CTAs. As a result, at $P = 2$ this strategy is roughly break-even for the forward pass, since the summary and main kernels are roughly the same amount of work as just a single main kernel on the uncut sequence. Therefore, we use this only for $P \ge 3$ in the forward pass. However, we can actually use this for $P \ge 2$ in the backward pass, as we will later discuss.

FlashQLA skips the summary pass for segments whenever the gates saturate in some warmup period, at the price of a warmup pass per segment. In our implementation, we remove this optional skip and warmup. At $P \ge 3$, the summary kernel is already a single wave that lasts as long as its longest segment whether or not some segments skip. A skip by every segment of every head would skip the kernel entirely, but we find this unlikely.

\subsection{$P = 2$: Split-$d_v$}
\label{sec:dvsplit}

As we discussed above, we roughly break even with intra-card CP at $P = 2$. For this case, we again follow FlashQLA's lead of splitting along the value head dimension, with some slight modifications for KDA and GDN-2.

\begin{figure}[H]
\centering
\begin{tikzpicture}[scale=0.85, every node/.style={font=\scriptsize}]
  \draw[fill=qcolor!15, draw=qcolor, thick] (0,1.5) rectangle (3.4,3.0);
  \node[anchor=west] at (0.15,2.25) {$\mathbf{S}^{\trans}[0{:}64]$};
  \node[anchor=east] at (3.25,2.25) {CTA 0};
  \draw[fill=vcolor!15, draw=vcolor, thick] (0,0) rectangle (3.4,1.5);
  \node[anchor=west] at (0.15,0.75) {$\mathbf{S}^{\trans}[64{:}128]$};
  \node[anchor=east] at (3.25,0.75) {CTA 1};
  \draw[<->] (-0.3,0) -- (-0.3,3.0);
  \node[rotate=90, anchor=south] at (-0.6,1.5) {$M = d_v = 128$};
  \draw[<->] (0,-0.3) -- (3.4,-0.3);
  \node at (1.7,-0.6) {$K = d_{qk}$};
  \draw[fill=kcolor!12, draw=kcolor, thick] (3.8,3.4) rectangle (5.0,6.6);
  \node[rotate=90] at (4.4,5.0) {$\mathbf{K}^{\trans}$, $\mathbf{Q}^{\trans}$, $\mathbf{T}^{\trans}$, $\mathbf{A}^{\trans}$};
  \node[anchor=west, align=left] at (5.3,5.0) {shared operands \\ (loaded or computed by both CTAs)};
  \draw[<->] (3.8,6.9) -- (5.0,6.9);
  \node at (4.4,7.2) {$N = C$};
  \draw[fill=qcolor!15, draw=qcolor, thick] (3.8,1.5) rectangle (5.0,3.0);
  \node at (4.4,2.25) {CTA 0};
  \draw[fill=vcolor!15, draw=vcolor, thick] (3.8,0) rectangle (5.0,1.5);
  \node at (4.4,0.75) {CTA 1};
  \node[anchor=west, align=left] at (5.3,1.5) {$(\mathbf{K}\mathbf{S})^{\trans}$, $\mathbf{V}^{\trans}$, $\mathbf{O}^{\trans}$ \\ (each CTA owns 64 of the 128 rows)};
\end{tikzpicture}
\caption{Split-$d_v$ in swap-AB orientation. The two CTAs own the two halves of the value rows of $\mathbf{S}^{\trans}$ and of the value-side tiles, and both load or compute the shared operands.}
\label{fig:dvsplit}
\end{figure}
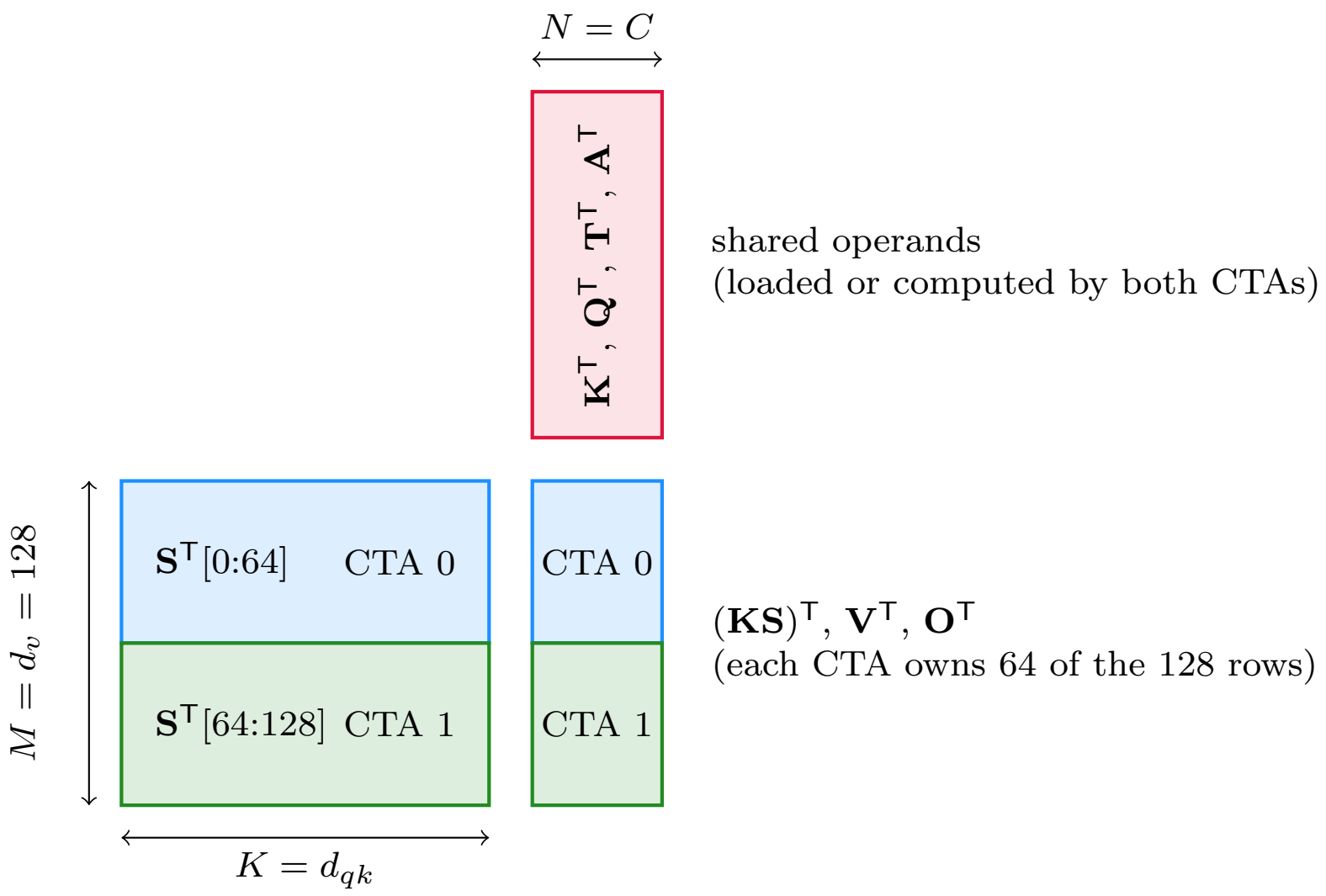

\subsubsection{Split-$d_v$}

The state is $d_{qk} \times d_v$, and every value column of it evolves independently of the others. Thus, we split the tile among two CTAs, each owning 64 of the 128 value columns of the same \texttt{(batch, head)} pair. $\mathbf{V}$, $\mathbf{O}$, $\mathbf{Y}$, $\mathbf{U}$, the state, and its checkpoints split by column, while $\mathbf{Q}$, $\mathbf{K}$, the gates, $\beta$, $\mathbf{K}\mathbf{K}^{\trans}$, $\mathbf{T}_{[t]}$, and $\mathbf{A}_{[t]}$ are recomputed by both. Table~\ref{tab:dvsplit} lists the resulting MMA shapes.

\begin{table}[H]
\centering
\small
\begin{tabular}{lcccl}
\toprule
MMA & $M$ & $N$ & $K$ & \\
\midrule
$\mathbf{K} \;@\; \mathbf{K}^{\trans}$ & $C$ & $C$ & $d_{qk}$ & shared \\
$\mathbf{Q} \;@\; \mathbf{K}^{\trans}$ & $C$ & $C$ & $d_{qk}$ & shared \\
$\mathbf{S}^{\trans} \;@\; \mathbf{K}^{\trans}$ & $d_v / 2$ & $C$ & $d_{qk}$ & split \\
$\mathbf{S}^{\trans} \;@\; \mathbf{Q}^{\trans}$ & $d_v / 2$ & $C$ & $d_{qk}$ & split \\
$\mathbf{Y}^{\trans} \;@\; \mathbf{T}^{\trans}$ & $d_v / 2$ & $C$ & $C$ & split \\
$\mathbf{U}^{\trans} \;@\; \mathbf{A}^{\trans}$ & $d_v / 2$ & $C$ & $C$ & split \\
$\mathbf{U}'^{\trans} \;@\; \mathbf{K}$ & $d_v / 2$ & $d_{qk}$ & $C$ & split \\
\bottomrule
\end{tabular}
\caption{The seven MMAs of Table~\ref{tab:swapab} per CTA under split-$d_v$.}
\label{tab:dvsplit}
\end{table}

Note that the five value-side MMAs drop to $M = d_v / 2 = 64$, so they fill only half the tensor core lanes and take as long as they did at $M = 128$. Furthermore, the two score MMAs are unchanged and issued by both CTAs, so the split does not help much with respect to the tensor core. Nevertheless, we observe a speedup because it halves the rest of the chunk work, namely $\mathbf{Y} = \mathbf{V} - \mathbf{K}\mathbf{S}$, the $\mathbf{U}$ downcast, the $\mathbf{O}$ epilogue, and the state rescale and restage. As we saw in Figure~\ref{fig:pipeline}, the state rescale can cause a bubble in the MMA pipeline, which this strategy shortens.

\subsubsection{Key-Side Preparation for KDA and GDN-2}

For KDA and GDN-2, we make one additional enhancement by moving the decay scaling work off the serial chain altogether. Instead, we launch a chunk-parallel preparation kernel that computes every chunk's decay operands, in the spirit of FlashKDA. Then, we have a recurrence kernel with two CTAs per \texttt{(batch, head)} that consumes them as described above. Our preparation kernel writes five intermediate tensors per token chunk, namely $\widehat{\mathbf{K}}_{[t]}$ and $\widehat{\mathbf{Q}}_{[t]}$, $\mathbf{T}_{[t]}^{\trans}\widetilde{\mathbf{K}}_{[t]}$, $\mathbf{A}_{[t]}\mathbf{T}_{[t]}$, and $\boldsymbol{\gamma}_{[t]}$. The prefill kernel then loads the intermediates straight into the shared memory slots the ordinary prefill fills itself, and its recurrence issues the four remaining MMAs per chunk.

\subsection{$P = 1$: Split-K}
\label{sec:splitk}

\begin{figure}[H]
\centering
\begin{tikzpicture}[
    token/.style={rectangle, draw=gray!30, fill=gray!5, minimum width=1.0cm, minimum height=0.6cm, font=\footnotesize\ttfamily, rounded corners=2pt},
    reset/.style={fill=red!10, draw=red!30},
    alpha_val/.style={font=\scriptsize\bfseries, text=black!70},
    alpha_reset/.style={font=\scriptsize\bfseries, text=red!80!black}
]
    \node[token] (t1) {The};
    \node[token, right=0.06cm of t1] (t2) {cat};
    \node[token, right=0.06cm of t2] (t3) {in};
    \node[token, right=0.06cm of t3] (t4) {the};
    \node[token, right=0.06cm of t4] (t5) {hat};
    \node[token, right=0.06cm of t5, reset] (t6) {.};
    \node[token, right=0.06cm of t6] (t7) {A};
    \node[token, right=0.06cm of t7] (t8) {dog};
    \node[token, right=0.06cm of t8] (t9) {barks};

    \node[below=0.15cm of t1, alpha_val] (a1) {0.99};
    \node[below=0.15cm of t2, alpha_val] (a2) {0.98};
    \node[below=0.15cm of t3, alpha_val] (a3) {0.97};
    \node[below=0.15cm of t4, alpha_val] (a4) {0.96};
    \node[below=0.15cm of t5, alpha_val] (a5) {0.99};
    \node[below=0.15cm of t6, alpha_reset] (a6) {0.05};
    \node[below=0.15cm of t7, alpha_val] (a7) {0.98};
    \node[below=0.15cm of t8, alpha_val] (a8) {0.97};
    \node[below=0.15cm of t9, alpha_val] (a9) {0.99};

    \draw[dashed, very thick, red!70] ([xshift=1pt, yshift=8pt]t6.north east) -- ([xshift=1pt, yshift=-48pt]a6.south -| t6.east);
    \node[font=\scriptsize\bfseries, text=red!80!black] at ([xshift=1pt, yshift=14pt]t6.north east) {cut};

    \draw[<->, thick, qcolor] ([yshift=-8pt]a1.south -| t1.west) -- ([yshift=-8pt]a1.south -| t6.east) node[midway, below, font=\scriptsize\bfseries, text=qcolor] {CTA 0 Outputs};

    \node[font=\scriptsize\bfseries, text=scolor, anchor=east] at ([yshift=-32pt, xshift=-3pt]a1.south -| t4.west) {$\mathbf{S} = \mathbf{0}$};
    \draw[<->, thick, dashed, scolor] ([yshift=-32pt]a1.south -| t4.west) -- ([yshift=-32pt]a1.south -| t6.east) node[midway, below, font=\scriptsize\bfseries, text=scolor] {Warmup, $\prod_r \alpha_r \le \epsilon$};
    \draw[<->, thick, vcolor] ([yshift=-32pt]a1.south -| t7.west) -- ([yshift=-32pt]a1.south -| t9.east) node[midway, below, font=\scriptsize\bfseries, text=vcolor] {CTA 1 Outputs};

    \node[left=0.2cm of a1, font=\scriptsize, anchor=east] {Gate $\alpha_t$};
\end{tikzpicture}
\caption{Split-K in the forward pass on a toy sequence. CTA 0 runs the tokens before the cut and writes their outputs. CTA 1 starts from a zero state at the beginning of the warmup, only accumulates the state through it, and writes outputs from the cut onward.}
\label{fig:splitk}
\end{figure}
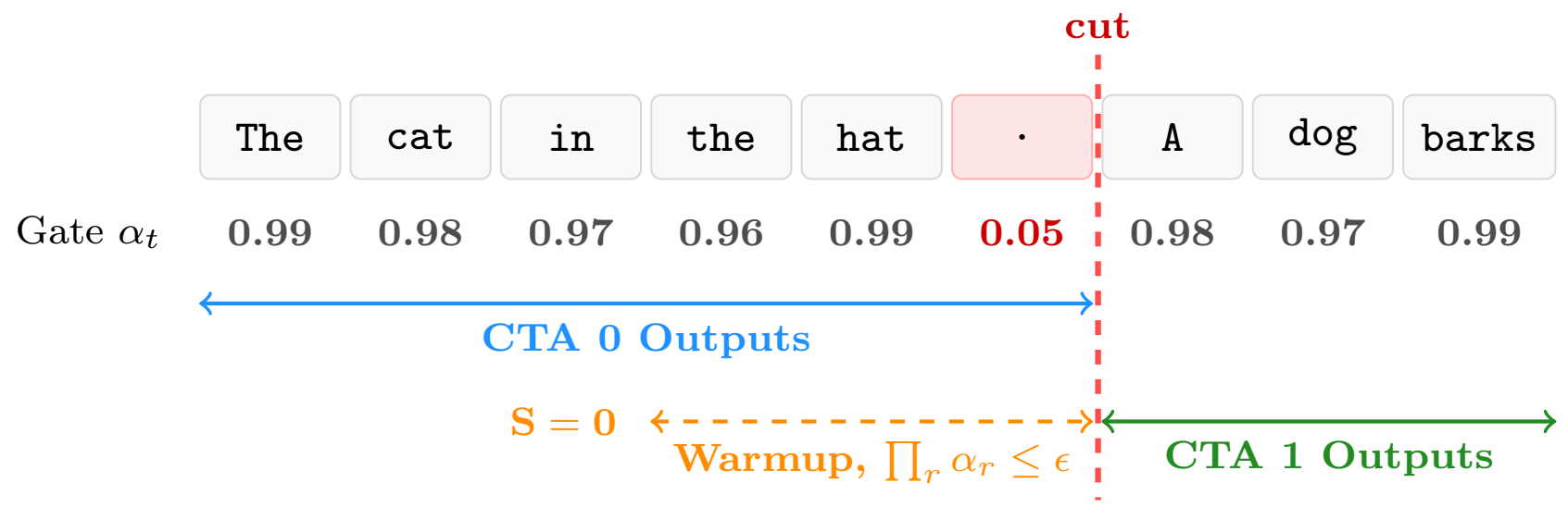

At $P = 1$, the grid still underfills the SMs, but by less than half. Neither of the previous two strategies applies. Thus, we resort to the gate-driven skip strategy on its own, which we will call ``split-K''. In this scheme, we first ``propose'' some cuts along the sequence length similar to the intra-card CP strategy. Then, for every proposed cut, we check whether the product of the gates over a warmup window of chunks ending at the cut has decayed below a threshold $\epsilon$. That is, $\prod_r \alpha_r \le \epsilon$ over the tokens $r$ of the warmup, with $\epsilon = e^{-10}$. If so, we cut at that point and run the segment after the cut with zero initial state starting at the warmup. In the warmup, the segment only accumulates the state and does not write outputs, since the segment before the cut already produced them. Figure~\ref{fig:splitk} illustrates the separation of work. Unlike the intra-card CP method, no state needs to be exchanged between segments, since the states are already correct up to some negligible error.

To show that this works, consider the affine form of Section~\ref{sec:chain}. Let the warmup be the span $(a, b]$ with the cut at $b$. Then, consider the difference between the state of the segment with the cut $\mathbf{S}'_t$ and the state of the segment that would have run without the cut $\mathbf{S}_t$. At token $a$, we start with the state $\mathbf{S}'_a = \mathbf{0}$, while the true state there is $\mathbf{S}_a$. Both runs have the same keys, values, and gates from token $a + 1$ onward, so the $\mathbf{k}_t^\trans\tilde{\mathbf{v}}_t$ terms cancel. Thus, for every output token $t > b$ of the segment, we have the following.
\[
\mathbf{S}_t - \mathbf{S}'_t = \mathbf{F}_t \cdots \mathbf{F}_{b+1}\,\mathbf{M}\,\mathbf{S}_a \qquad
\mathbf{o}_t - \mathbf{o}'_t = \mathbf{q}_t\left(\mathbf{S}_t - \mathbf{S}'_t\right)
\]
Here, $\mathbf{M} = \mathbf{F}_b \cdots \mathbf{F}_{a+1}$ is the same transfer matrix from Section~\ref{sec:chain}. Intra-card CP computes $\mathbf{M}$ and $\mathbf{H}$ exactly. Split-K instead looks for spans where $\mathbf{M}$ is negligible, so that the correction can be dropped, and we need not calculate $\mathbf{M}$. For GDN and GDP, the decay is a scalar, so $\diag(\boldsymbol{\alpha}_t) = \alpha_t\mathbf{I}$ commutes with the erase $\mathbf{E}_t = \mathbf{I} - \beta_t\mathbf{k}_t^\trans\mathbf{k}_t$. Therefore, the transfer matrix factors nicely into the gates and the erases.
\[
\mathbf{M} = \Bigg(\prod_{r=a+1}^{b} \alpha_r\Bigg)\,\mathbf{E}_b \cdots \mathbf{E}_{a+1}
\]
Each $\mathbf{E}_r$ is symmetric with eigenvalue $1 - \beta_r\lVert\mathbf{k}_r\rVert^2$ along $\mathbf{k}_r$ and $1$ on everything orthogonal to $\mathbf{k}_r$. Assuming the queries and keys are L2 normalized and the write strength is the result of a sigmoid, we have that $\lVert \mathbf{q}_t \rVert_2 = \lVert \mathbf{k}_r \rVert_2 = 1$ and $\beta_r \in (0, 1)$. Thus, the eigenvalue along $\mathbf{k}_r$ lies in $(0, 1)$, and every erase has spectral norm one. From this, we can deduce two bounds. First, every transition $\mathbf{F}_r = \alpha_r \mathbf{E}_r$ has spectral norm $\alpha_r \le 1$, so the product $\mathbf{F}_t \cdots \mathbf{F}_{b+1}$ of the transitions after the cut has norm at most one. Second, the transfer matrix $\mathbf{M}$ is such that $\lVert \mathbf{M} \rVert_2 \le \prod_{r=a+1}^{b} \alpha_r$, since its erases have norm one. Applying both of these inequalities and using the threshold $\prod_{r=a+1}^{b} \alpha_r \le \epsilon$, we can bound the state error relative to $\mathbf{S}_a$, the true state entering the warmup, as follows.
\[
\lVert \mathbf{S}_t - \mathbf{S}'_t \rVert_2 \;\le\; \lVert \mathbf{F}_t \cdots \mathbf{F}_{b+1} \rVert_2\,\lVert \mathbf{M} \rVert_2\,\lVert \mathbf{S}_a \rVert_2 \;\le\; \Bigg(\prod_{r=a+1}^{b} \alpha_r\Bigg) \lVert \mathbf{S}_a \rVert_2 \;\le\; \epsilon\,\lVert \mathbf{S}_a \rVert_2
\]
Thus, at most $\epsilon$ of the initial state would have reached any token after the cut.

For KDA and GDN-2, the decay is a per-channel vector, so $\diag(\boldsymbol{\alpha}_t)$ no longer commutes with $\mathbf{E}_t$ and $\mathbf{M}$ does not factor into a gate product and an erase product. Intuitively, the erase can mix state between channels that decay at different rates. We have that $\lVert \diag(\boldsymbol{\alpha}_r) \rVert_2 = \max_c \alpha_{r,c}$. Thus, we can give the following bound on the transfer matrix $\mathbf{M}$.
\[
\lVert \mathbf{M} \rVert_2 \;\le\; \prod_{r=a+1}^{b} \lVert \mathbf{E}_r \rVert_2 \max_{c} \alpha_{r,c}
\]
This bound is much weaker than the GDN and GDP one though. The gate factor is the product of each token's slowest channel. Per-channel gates exist so that some channels can decay slower than others, so this product decays far more slowly than any single channel. Therefore, a cut is much less likely to be accepted and just adds overhead for the required pre-processing. As a result, we do not use this method for KDA and GDN-2.

\begin{figure}[H]
\centering
\begin{tikzpicture}[x=0.5cm, y=0.36cm, every node/.style={font=\tiny}]
  \begin{scope}[shift={(0,0)}]
    \node[anchor=south west, font=\scriptsize] at (0,4.15) {in sequence order};
    \foreach \c in {0,...,3} { \node[anchor=east] at (-0.2,{3-\c+0.5}) {CTA \c}; }
    \foreach \c/\s/\l in {0/0/2, 1/0/2, 2/0/2, 3/0/2, 0/2/2, 1/2/2} {
      \fill[qcolor!25] (\s,{3-\c}) rectangle ({\s+\l},{3-\c+0.9}); \draw[qcolor!70] (\s,{3-\c}) rectangle ({\s+\l},{3-\c+0.9});
      \node at ({\s+\l/2},{3-\c+0.45}) {\l};
    }
    \fill[kcolor!30] (2,1) rectangle (10,1.9); \draw[kcolor] (2,1) rectangle (10,1.9); \node at (6,1.45) {8};
    \fill[scolor!35] (2,0) rectangle (8,0.9);  \draw[scolor] (2,0) rectangle (8,0.9);  \node at (5,0.45) {6};
    \draw[dashed, thick, kcolor] (10,-0.3) -- (10,4.0); \node[anchor=south, text=kcolor] at (10,4.05) {done at 10};
    \draw[->] (0,-0.5) -- (11,-0.5); \foreach \t in {0,2,4,6,8,10} { \draw (\t,-0.4) -- (\t,-0.6); \node[anchor=north] at (\t,-0.6) {\t}; }
  \end{scope}
  \begin{scope}[shift={(14,0)}]
    \node[anchor=south west, font=\scriptsize] at (0,4.15) {longest first (LPT)};
    \foreach \c in {0,...,3} { \node[anchor=east] at (-0.2,{3-\c+0.5}) {CTA \c}; }
    \fill[kcolor!30] (0,3) rectangle (8,3.9); \draw[kcolor] (0,3) rectangle (8,3.9); \node at (4,3.45) {8};
    \fill[scolor!35] (0,2) rectangle (6,2.9); \draw[scolor] (0,2) rectangle (6,2.9); \node at (3,2.45) {6};
    \foreach \c/\s in {2/0, 2/2, 2/4, 3/0, 3/2, 3/4} {
      \fill[qcolor!25] (\s,{3-\c}) rectangle ({\s+2},{3-\c+0.9}); \draw[qcolor!70] (\s,{3-\c}) rectangle ({\s+2},{3-\c+0.9});
      \node at ({\s+1},{3-\c+0.45}) {2};
    }
    \draw[dashed, thick, vcolor] (8,-0.3) -- (8,4.0); \node[anchor=south, text=vcolor!70!black] at (8,4.05) {done at 8};
    \draw[->] (0,-0.5) -- (11,-0.5); \foreach \t in {0,2,4,6,8,10} { \draw (\t,-0.4) -- (\t,-0.6); \node[anchor=north] at (\t,-0.6) {\t}; }
  \end{scope}
\end{tikzpicture}
\caption{LPT scheduling on a toy example. Suppose we have eight work items of lengths 2, 2, 2, 2, 2, 2, 8, and 6 chunks on four CTAs. Each CTA takes the next item as it is free. In the naive order, the two long items start last, so one CTA finishes at 10 and another at 8 while the other two idle from 4. Sorted longest first (LPT), the long items start at the beginning. The total time is now the length of the longest item, which is optimal for this case.}
\label{fig:lpt}
\end{figure}
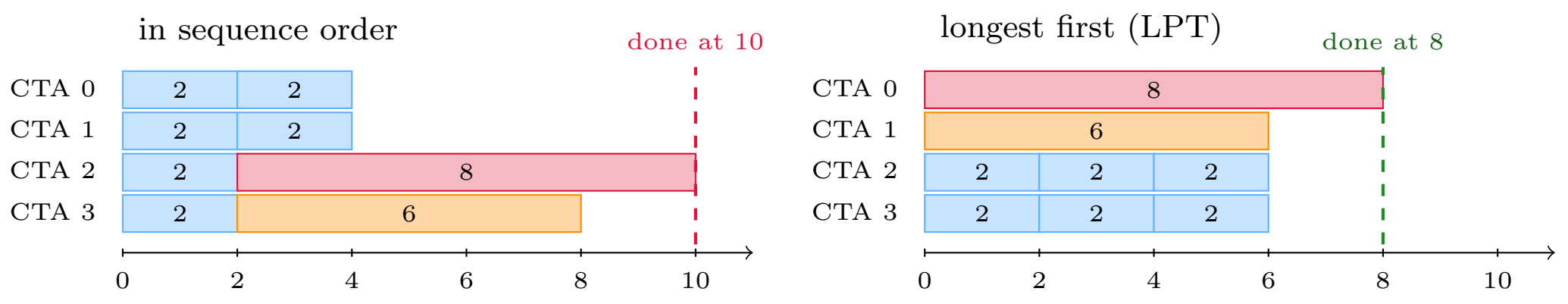

We should highlight that this method is data-dependent and may not result in any accepted splits at all. As we alluded to, it requires that the gates are ``well-behaved'', such that there are no heads that decay very little across the token horizon. Thus, we only use this strategy at $P = 1$, where neither of the above remedies applies. Furthermore, we should note that this strategy may launch more waves than we started off with. However, this is intentional. Cutting the tiles more aggressively allows us to reach a tile number that is closer to a multiple of the number of SMs, allowing us to fully utilize the GPU. Furthermore, since not all cuts may be accepted, we may end up with many segments of varying lengths. Thus, we sort the work items in descending order of length, adopting a longest-processing-time-first (LPT) scheduling design. This design also helps with variable sequence lengths. Figure~\ref{fig:lpt} illustrates a scenario where such LPT scheduling is useful.

\subsection{$P < 1$: Multiple Waves}
\label{sec:waves}

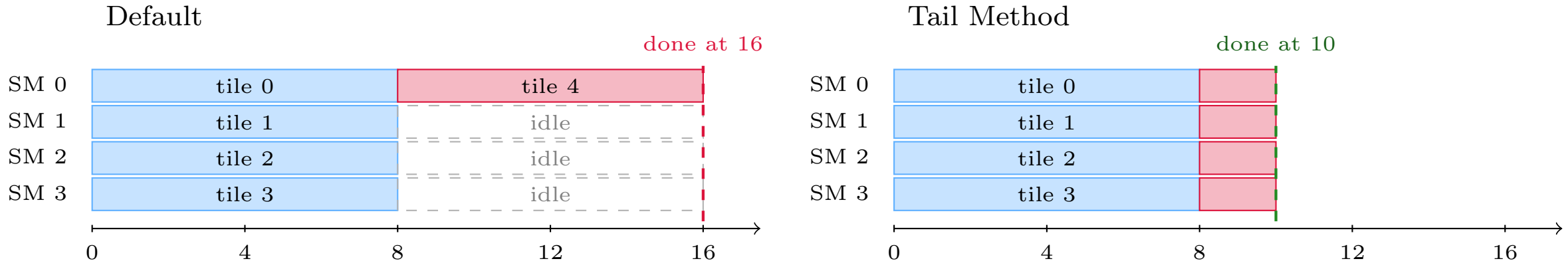
\begin{figure}[H]
\centering
\begin{tikzpicture}[x=0.38cm, y=0.36cm, every node/.style={font=\tiny}]
  \begin{scope}[shift={(0,0)}]
    \node[anchor=south west, font=\scriptsize] at (0,4.75) {Default};
    \foreach \c in {0,...,3} { \node[anchor=east] at (-0.3,{3-\c+0.5}) {SM \c}; }
    \foreach \c in {0,...,3} {
      \fill[qcolor!25] (0,{3-\c}) rectangle (8,{3-\c+0.9}); \draw[qcolor!70] (0,{3-\c}) rectangle (8,{3-\c+0.9});
      \node at (4,{3-\c+0.45}) {tile \c};
    }
    \fill[kcolor!30] (8,3) rectangle (16,3.9); \draw[kcolor] (8,3) rectangle (16,3.9); \node at (12,3.45) {tile 4};
    \foreach \c in {1,...,3} {
      \draw[dashed, gray!60] (8,{3-\c}) rectangle (16,{3-\c+0.9}); \node[text=gray] at (12,{3-\c+0.45}) {idle};
    }
    \draw[dashed, thick, kcolor] (16,-0.3) -- (16,4.0); \node[anchor=south, text=kcolor] at (16,4.05) {done at 16};
    \draw[->] (0,-0.5) -- (17.5,-0.5); \foreach \t in {0,4,8,12,16} { \draw (\t,-0.4) -- (\t,-0.6); \node[anchor=north] at (\t,-0.6) {\t}; }
  \end{scope}
  \begin{scope}[shift={(21,0)}]
    \node[anchor=south west, font=\scriptsize] at (0,4.75) {Tail Method};
    \foreach \c in {0,...,3} { \node[anchor=east] at (-0.3,{3-\c+0.5}) {SM \c}; }
    \foreach \c in {0,...,3} {
      \fill[qcolor!25] (0,{3-\c}) rectangle (8,{3-\c+0.9}); \draw[qcolor!70] (0,{3-\c}) rectangle (8,{3-\c+0.9});
      \node at (4,{3-\c+0.45}) {tile \c};
      \fill[kcolor!30] (8,{3-\c}) rectangle (10,{3-\c+0.9}); \draw[kcolor] (8,{3-\c}) rectangle (10,{3-\c+0.9});
    }
    \draw[dashed, thick, vcolor] (10,-0.3) -- (10,4.0); \node[anchor=south, text=vcolor!70!black] at (10,4.05) {done at 10};
    \draw[->] (0,-0.5) -- (17.5,-0.5); \foreach \t in {0,4,8,12,16} { \draw (\t,-0.4) -- (\t,-0.6); \node[anchor=north] at (\t,-0.6) {\t}; }
  \end{scope}
\end{tikzpicture}
\caption{Wave quantization on a toy machine with four SMs and five tiles of eight chunks. \textbf{Left}: With one CTA per tile, the second wave keeps one SM busy while three idle. \textbf{Right}: The four tiles of the full wave run together, and the leftover tile is cut into four segments by the plan of Table~\ref{tab:plans}.}
\label{fig:waves}
\end{figure}

Finally, when $B \cdot H > \#\text{SMs}$, the grid runs more than one wave, but we might run into some wave quantization issues. With $N = B \cdot H$ tiles on $S$ SMs, the persistent kernel runs $\lceil N / S \rceil$ waves, and the last wave causes wave quantization whenever $N$ is not a multiple of $S$, which can be extreme. For example, at 200 tiles on 152 SMs the second wave runs 48 tiles while 104 SMs idle, and at 160 tiles the second wave runs 8 tiles while 144 SMs idle, so the kernel takes two full waves for $1.05$ waves of work. The left half of Figure~\ref{fig:waves} shows the effect on a toy machine with four SMs and five tiles, where the second wave keeps one SM busy and the kernel takes two waves for $1.25$ waves of work.

Instead, we would like the kernel to run a fractional number of waves, in the spirit of Stream-K~\cite{osama2023streamk}. We adopt a similar idea and call it the ``tail'' method. We round the tile count down to the nearest multiple of the SM count, $W S$ with $W = \lfloor N / S \rfloor$, and run those $W$ full waves together. For these waves, we can assign one CTA per tile, since they fill the machine by construction. The $R = N - W S$ leftover tiles, which would have formed the partial last wave, are then exactly the problem of the previous sections with $P = \lfloor S / R \rfloor$. Thus, we handle them in the same manner as specified in Table~\ref{tab:plans}. Note that at $P = 1$ the leftover tiles cannot be split evenly either. Thus, for GDN and GDP, we apply split-K over the whole batch rather than just the last wave.

\subsection{Backward}
\label{sec:bwd_cp}

For the backward pass, all of the strategies discussed also apply. Since $\mathbf{S}_b = \mathbf{M}\mathbf{S}_a + \mathbf{H}$, the state gradient $d\mathbf{S}_a$ that a span hands to the segment on its left is likewise an affine function of the state gradient $d\mathbf{S}_b$ that enters it from the right.
\[
d\mathbf{S}_a = \mathbf{M}^{\trans} d\mathbf{S}_b + \mathbf{G}
\]
Here, $\mathbf{G}$ is the $d\mathbf{S}_a$ the span produces on its own when it starts from $d\mathbf{S}_b = \mathbf{0}$, the backward counterpart of $\mathbf{H}$.

Previously, for the forward pass, we set a threshold of $P > 2$ for intra-card CP under the assumption that the summary kernel did roughly as much work as the prefill itself. However, for the backward pass, the summary kernel does much less work because we do not need to compute any of the gradients. We write a backward summary kernel by taking the fused backward kernel from Section~\ref{sec:fused} and removing everything except the $d\mathbf{S}$ recurrence. Then, the same chain kernel from the forward pass runs right to left with $\mathbf{M}_j$ transposed. Finally, we run the fused backward kernel again with the corrected state gradients. We illustrate this below.

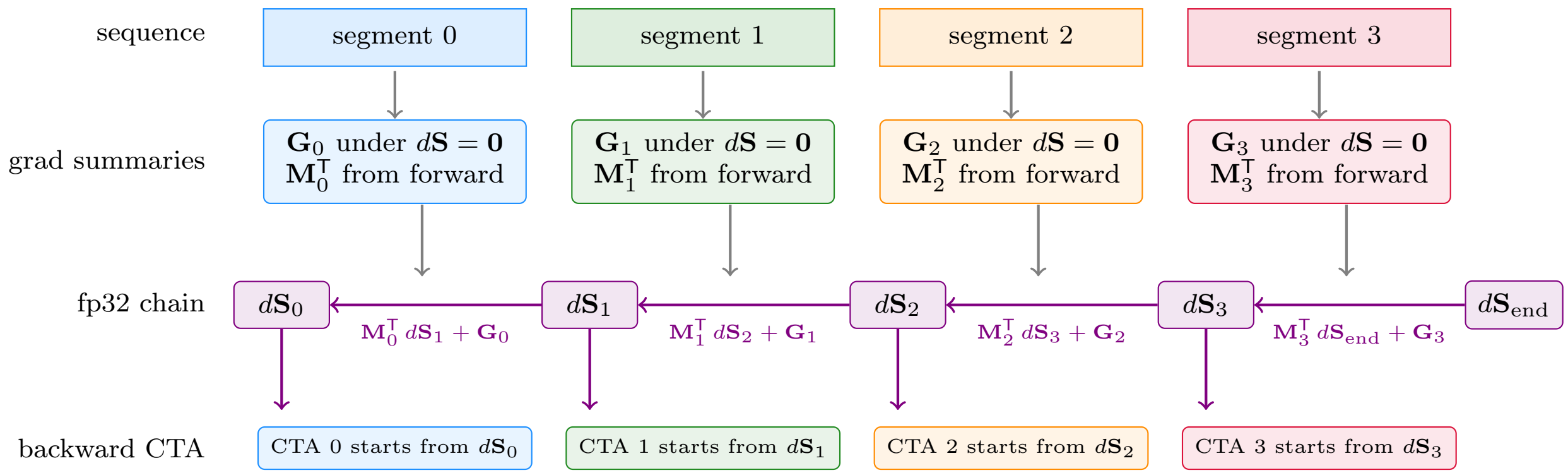
\begin{figure}[H]
\centering
\begin{tikzpicture}[x=0.95cm, y=1cm, every node/.style={font=\scriptsize}]
  \node[anchor=east] at (-0.5, 4.3) {sequence};
  \node[anchor=east] at (-0.5, 3.0) {grad summaries};
  \node[anchor=east] at (-0.5, 1.5) {fp32 chain};
  \node[anchor=east] at (-0.5, 0.0) {backward CTA};
  \foreach \j/\c in {0/qcolor, 1/vcolor, 2/scolor, 3/kcolor} {
    \draw[fill=\c!15, draw=\c] (3.4*\j, 4.0) rectangle (3.4*\j+2.9, 4.6);
    \node at (3.4*\j+1.45, 4.3) {segment $\j$};
    \draw[->, thick, gray] (3.4*\j+1.45, 3.95) -- (3.4*\j+1.45, 3.45);
    \node[draw=\c, fill=\c!10, rounded corners=2pt, align=center, minimum width=2.75cm, minimum height=0.8cm] at (3.4*\j+1.45, 3.0) {$\mathbf{G}_{\j}$ under $d\mathbf{S} = \mathbf{0}$ \\ $\mathbf{M}_{\j}^{\trans}$ from forward};
    \draw[->, thick, gray] (3.4*\j+1.75, 2.55) -- (3.4*\j+1.75, 1.8);
    \node[draw=statecolor, fill=statecolor!10, rounded corners=2pt, minimum width=1.0cm] (y\j) at (3.4*\j+0.2, 1.5) {$d\mathbf{S}_{\j}$};
    \draw[->, thick, statecolor] (y\j) -- (3.4*\j+0.2, 0.4);
    \node[draw=\c, fill=\c!10, rounded corners=2pt, minimum width=2.75cm, minimum height=0.42cm, font=\tiny] at (3.4*\j+1.45, 0.0) {CTA $\j$ starts from $d\mathbf{S}_{\j}$};
  }
  \node[draw=statecolor, fill=statecolor!10, rounded corners=2pt, minimum width=1.0cm] (y4) at (13.8, 1.5) {$d\mathbf{S}_{\text{end}}$};
  \foreach \j/\k/\l in {3/4/\text{end}, 2/3/3, 1/2/2, 0/1/1} {
    \draw[->, thick, statecolor] (y\k) -- node[below, font=\tiny] {$\mathbf{M}_{\j}^{\trans}\,d\mathbf{S}_{\l} + \mathbf{G}_{\j}$} (y\j);
  }
\end{tikzpicture}
\caption{Intra-card CP in the backward pass. Each segment's backward summary kernel produces $\mathbf{G}_j$ from a zero outgoing gradient. $\mathbf{M}_j$ is reused from the forward to recover the true $d\mathbf{S}_{j}$. Then, we run the main kernel started from $d\mathbf{S}_{j}$.}
\label{fig:chainbwd}
\end{figure}

Thus, we change the threshold to $P \ge 2$ for intra-card CP in the backward pass, removing the need for split-$d_v$. The split-K and tail methods still apply in the same way.

\subsection{Inter-Card CP}
\label{sec:intercp}

For context parallelism across GPUs (inter-card CP), we can reuse the same algebra laid out above. In \cudnn{}, we offer \texttt{*\_summary} ops, one per variant, to run the fused summary kernel over a rank's local shard from a zero state and return $(\mathbf{H}, \mathbf{M})$ per \texttt{(batch, head)} pair. The framework only needs to all-gather these summaries. The main op of every rank will accept the summaries of all the ranks before it, in rank order, and compose its own incoming state from them on the device using the following recurrence.
\[
\mathbf{S}_{\text{final}}^{\trans} = \mathbf{S}_{\text{init}}^{\trans}\,\mathbf{M}^{\trans} + \mathbf{H}^{\trans}
\]
As discussed above, the backward is similar. The \texttt{*\_summary\_bwd} ops run the backward summary kernel over the shard from a zero outgoing gradient and return $\mathbf{G}$ together with the transfer matrix. After the same all-gather, the backward op of every rank accepts the summaries of all the ranks after it and composes its outgoing state gradient the same way.
\[
d\mathbf{S}_{\text{init}}^{\trans} = d\mathbf{S}_{\text{final}}^{\trans}\,\mathbf{M} + \mathbf{G}^{\trans}
\]

The two levels of CP, intra-card and inter-card, compose nicely to get optimal performance. Calling either the summary or the main op on a shard that leaves SMs idle can run intra-card CP itself. The shard is further cut into segments to fully utilize the GPU.

\section{L2 Normalization, Safe Gates, and Beta Sigmoid}
\label{sec:block}

In the spirit of all prior work, we also extended the operator to support all the common surrounding operations including L2 normalization, safe gates, and beta sigmoid. A typical layer projects the input and applies some convolution yielding $\mathbf{q}$, $\mathbf{k}$, $\mathbf{v}$, a scalar or per-channel gate logit $a_t$, and a write-strength logit $b_t$. Before the main recurrence, it applies the following elementwise transforms.
\begin{align*}
\hat{\mathbf{q}}_t &= \mathbf{q}_t / \lVert \mathbf{q}_t \rVert_2 \qquad \hat{\mathbf{k}}_t = \mathbf{k}_t / \lVert \mathbf{k}_t \rVert_2 && \text{row L2 normalization} \\
\beta_t &= \sigma(b_t) && \text{write strength ($2\sigma(b_t)$ if negative eigenvalues~\cite{grazzi2024negative} are used)} \\
g_t &= -\exp(A_{\log})\,\mathrm{softplus}(a_t + b_{\Delta}) && \text{without safe gate (GDN, GDP)} \\
\mathbf{g}_t &= \ell \cdot \sigma\big(\exp(A_{\log})\,(\mathbf{a}_t + \mathbf{b}_{\Delta})\big) && \text{with safe gate (KDA, GDN-2, $\ell = -5$)}
\end{align*}
Here, $\alpha_t = \exp(g_t)$ is the decay the recurrence uses, $A_{\log}$ is a learned log-amplitude per head, and $b_{\Delta}$ is a learned bias per head (GDN, GDP) or per head and channel (KDA, GDN-2). The normalized query is also typically scaled by some constant.

Our ops take the inputs and apply each transform with its own argument. The normalization is selected by \texttt{use\_qk\_l2norm\_in\_kernel}, the write strength by \texttt{use\_beta\_sigmoid\_in\_kernel}, with \texttt{allow\_neg\_eigval} selecting $2\sigma$, and the gate transform of the family by \texttt{safe\_gate}. The gate parameters $A_{\log}$ and $b_{\Delta}$ are passed through the arguments \texttt{a\_log} and \texttt{dt\_bias} respectively, and the query scalar through \texttt{scale}. The backward returns gradients with respect to the same inputs, before any transform, plus $dA_{\log}$ and $db_{\Delta}$.

\begin{table}[H]
\centering
\small
\begin{tabular}{@{}lllllll@{}}
\toprule
 & & & & & \multicolumn{2}{c}{Ours} \\
\cmidrule(l){6-7}
Transform & Pass & FLA & FlashQLA & FlashKDA & GDN, GDP & KDA, GDN-2 \\
\midrule
L2 normalization & forward & separate & separate & preparation & separate & fused \\
 & backward & separate & separate & -- & fused & fused \\
write strength & forward & separate & -- & preparation & fused & fused \\
 & backward & separate & -- & -- & fused & fused \\
gate & forward & cumsum & -- & -- & fused & -- \\
 & backward & separate & -- & -- & fused & -- \\
safe gate & forward & cumsum & -- & preparation & -- & fused \\
 & backward & separate & -- & -- & -- & fused \\
\bottomrule
\end{tabular}
\caption{Location of each transform in each implementation and pass. Separate means kernels of its own around the recurrence, fused means inside the recurrence kernels, and preparation and cumsum name FlashKDA's preparation kernel and FLA's gate cumsum kernel. A dash marks a transform the implementation does not provide.}
\label{tab:transforms}
\end{table}

\section{Results}
\label{sec:perf}

In this section, we present the performance results from the above optimization work. All measurements in this section are on GB200 with bf16 inputs. The L2 normalization is on for every backend, the write strengths are random, and the decay gates are all ones, which is the worst case for any data-dependent method. Each reported time is the device time of one pass. For stability, we ran 20 iterations, discarded the first five, and reported the median of the remaining 15. Before every iteration, we flushed the L2 cache by zeroing a 256 MB buffer. The baselines are FLA's Triton kernels for all five variants, FlashQLA for the GDN forward and backward, FlashKDA for the KDA forward, and mamba\_ssm~\cite{dao2024transformersssmsgeneralizedmodels} for the Mamba-2 forward and backward. GDN and Mamba-2 group $H_{qk}$ query and key heads over $H_v = H$ value heads. We swept each variant in three ways.
\begin{enumerate}[(1)]
    \item Fix $B = 1$ and $H = 16$ and sweep over $s$ to profile the low $B \cdot H$ regime for TP configurations.
    \item Fix $B = 4$ and sweep over $s$ to show performance across different CP configurations.
    \item Fix $s = 8192$ and sweep over $B$ to show performance across different DP configurations.
\end{enumerate}
The complete set of charts, including the batch sweep, the DRAM bandwidth view, and sibling GB300 runs, is on the benchmark page at \url{https://github.com/NVIDIA/cudnn-frontend/tree/develop/benchmark/linear_attention}. We used the latest versions of each library at the time of writing. Table~\ref{tab:versions} lists the version and commit of each library we used.

\begin{table}[H]
\centering
\small
\begin{tabular}{lll}
\toprule
Library & Version & Commit \\
\midrule
FLA & 0.6.0 & \texttt{9f38d249} \\
mamba\_ssm & 2.3.2.post1 & \texttt{e9594ce1} \\
FlashQLA & 0.1.3 & \texttt{da06429d} \\
FlashKDA & \texttt{master} & \texttt{7afb9f45} \\
\cudnn{} Frontend & 1.34.0 & \texttt{develop} \\
\bottomrule
\end{tabular}
\caption{The library versions for the performance numbers below.}
\label{tab:versions}
\end{table}

Our kernels appear twice in every chart to show the performance with and without the state checkpoints described in Section~\ref{sec:ckpt}. The ``state on'' runs the forward at the per-chunk checkpoint cadence of Section~\ref{sec:ckpt} and lets the backward consume the series. The default runs without checkpoints and recomputes the state in the backward pass. The state-on forward pays for the stores of the series, and the default backward pays for the state-only recompute kernel. Note that the other baselines also stash intermediates between the forward and backward passes. FLA's autograd keeps the inputs, the chunk-local gate sums, and the per-chunk inverse $\mathbf{T}_{[t]}$. FlashQLA keeps the same tensors, plus the segment states and transfer matrices of its context parallelism when enabled.

\subsection{Low $B \cdot H$ Region}

First, we profile the low $B \cdot H$ region, where the optimizations in Section~\ref{sec:regimes} shine. We set $B \cdot H = 16$, so we use the intra-card CP approach at $P = 9$. Figures~\ref{fig:res-fwd-lowbh} and~\ref{fig:res-bwd-lowbh} show the forward and backward sweeps. The speedup over FLA grows with the sequence length, from $1.5\times$ to $2.8\times$ at $s = 2048$ to $4.9\times$ to $8.7\times$ at $s = 32768$ in the forward, and in the backward from $1.8\times$ to $3.9\times$ up to $2.8\times$ to $6.1\times$ without checkpoints and from $2.4\times$ to $5.4\times$ up to $3.7\times$ to $8.1\times$ with the checkpoint series. Against the other baselines in this regime, the GDN forward runs $1.4\times$ to $2.2\times$ FlashQLA's, and the GDN backward runs $1.1\times$ to $1.3\times$ FlashQLA's without checkpoints and $1.7\times$ to $1.8\times$ with them. The KDA forward runs $2.5\times$ to $6.1\times$ FlashKDA's, and the Mamba-2 forward $1.3\times$ to $3.3\times$ and backward $2.2\times$ to $5.5\times$ mamba\_ssm's. The performance improves as the sequence length grows, since the overhead cost of the summary and chain stages is amortized over more chunks. Mamba-2, with no inverse to compute and $C = 128$, climbs the highest from a pure TFLOPS perspective.

\begin{figure}[H]
\centering
\input{figures/results_fwd_lowbh}
\caption{Forward throughput at $B = 1$ and $H = 16$ on GB200 sweeping over the sequence length in tokens. $d_{qk} = 128$ throughout, and $d_v = 128$ except for GDP ($n_h = 3$) and Mamba-2 at $d_v = 64$. The label above each group is the speedup of Ours over FLA.}
\label{fig:res-fwd-lowbh}
\end{figure}
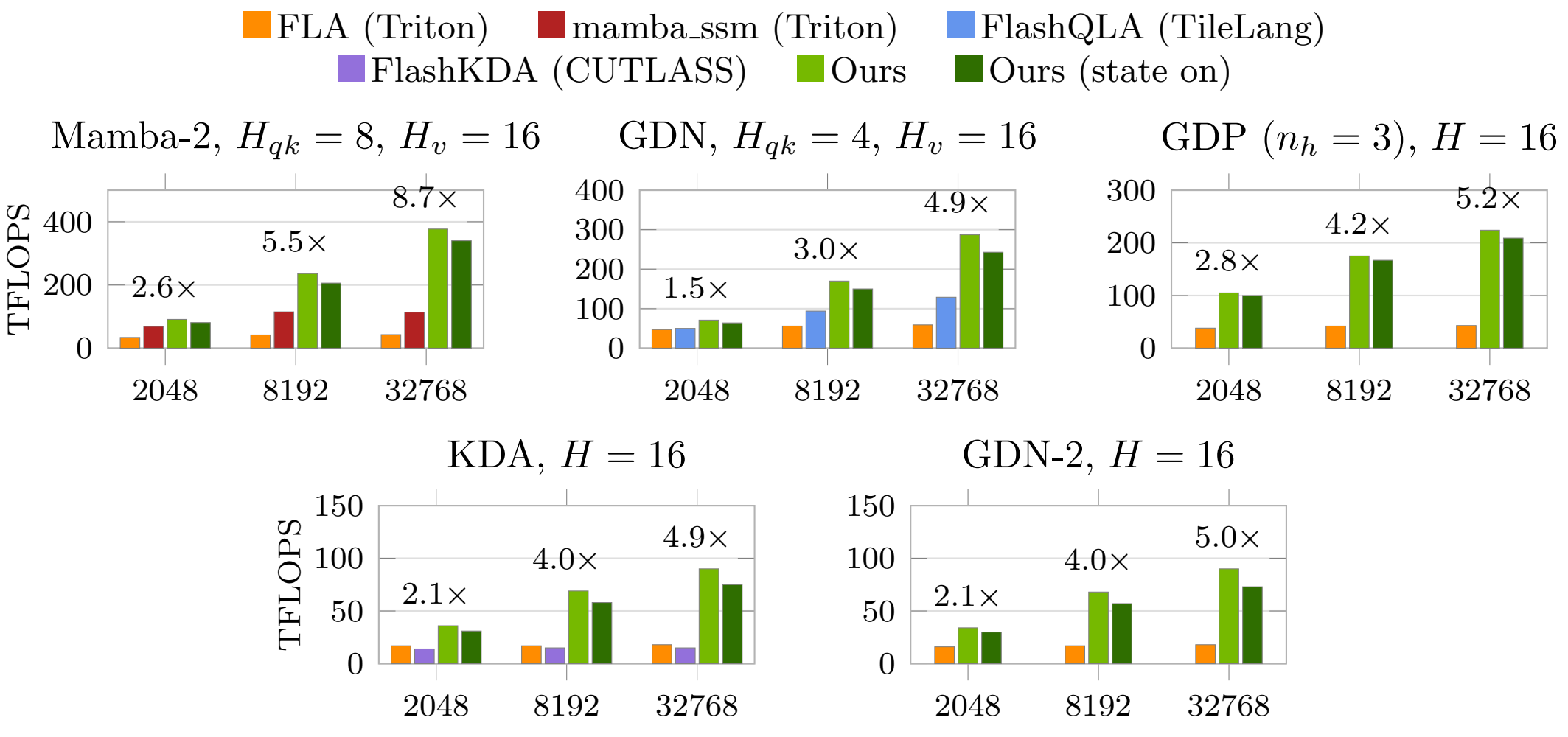

\begin{figure}[H]
\centering
\input{figures/results_bwd_lowbh}
\caption{Backward throughput at $B = 1$ and $H = 16$ on GB200 sweeping over the sequence length in tokens, at the same head dimensions as Figure~\ref{fig:res-fwd-lowbh}. The label above each group is its speedup over FLA.}
\label{fig:res-bwd-lowbh}
\end{figure}
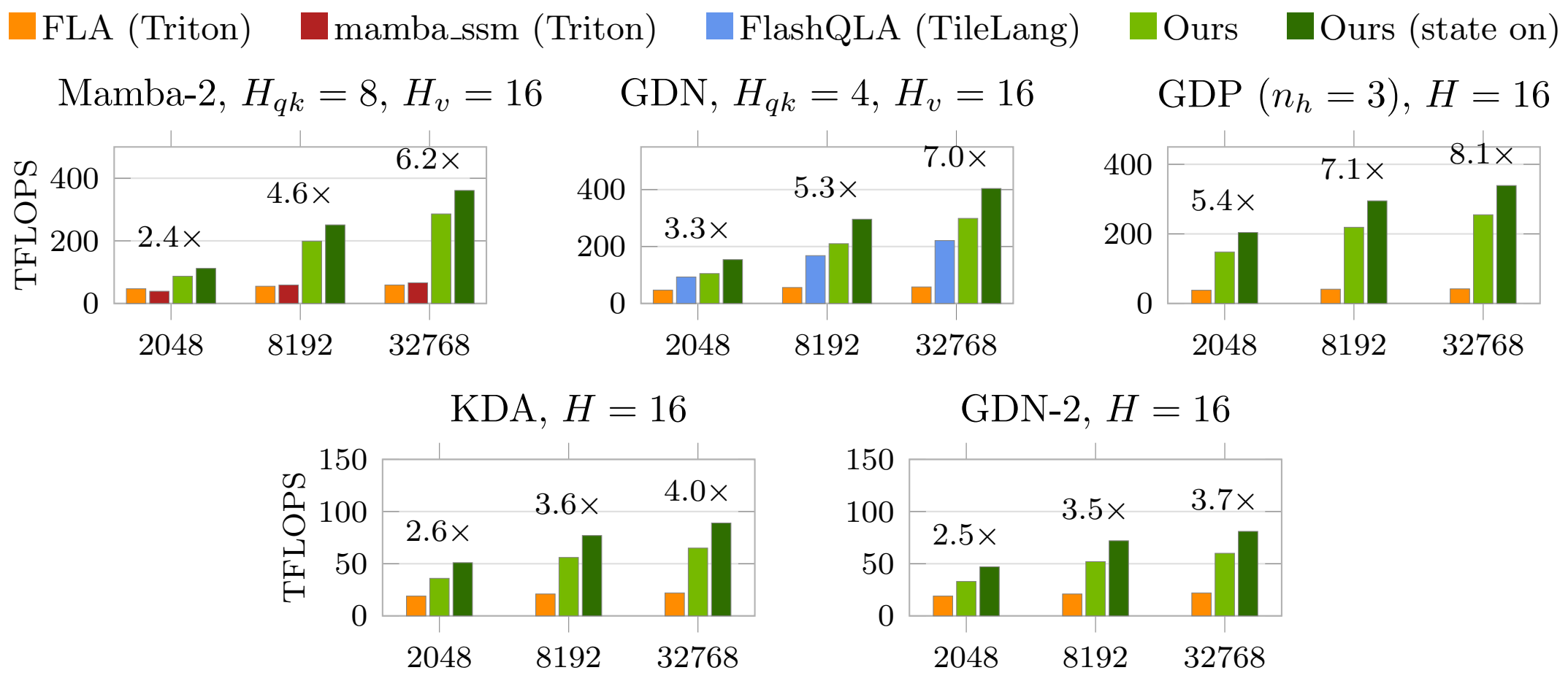

\subsection{Fully Saturated Region}

\begin{figure}[H]
\centering
\input{figures/results_fwd_b4}
\caption{Forward throughput at $B = 4$ on GB200 sweeping over the sequence length in tokens. $H = 64$ for Mamba-2 ($H_{qk} = 8$), GDN ($H_{qk} = 16$), and GDN-2, $H = 40$ for GDP, and $H = 96$ for KDA, at the head dimensions of Figure~\ref{fig:res-fwd-lowbh}. The label above each group is the speedup of Ours over FLA.}
\label{fig:res-fwd-b4}
\end{figure}
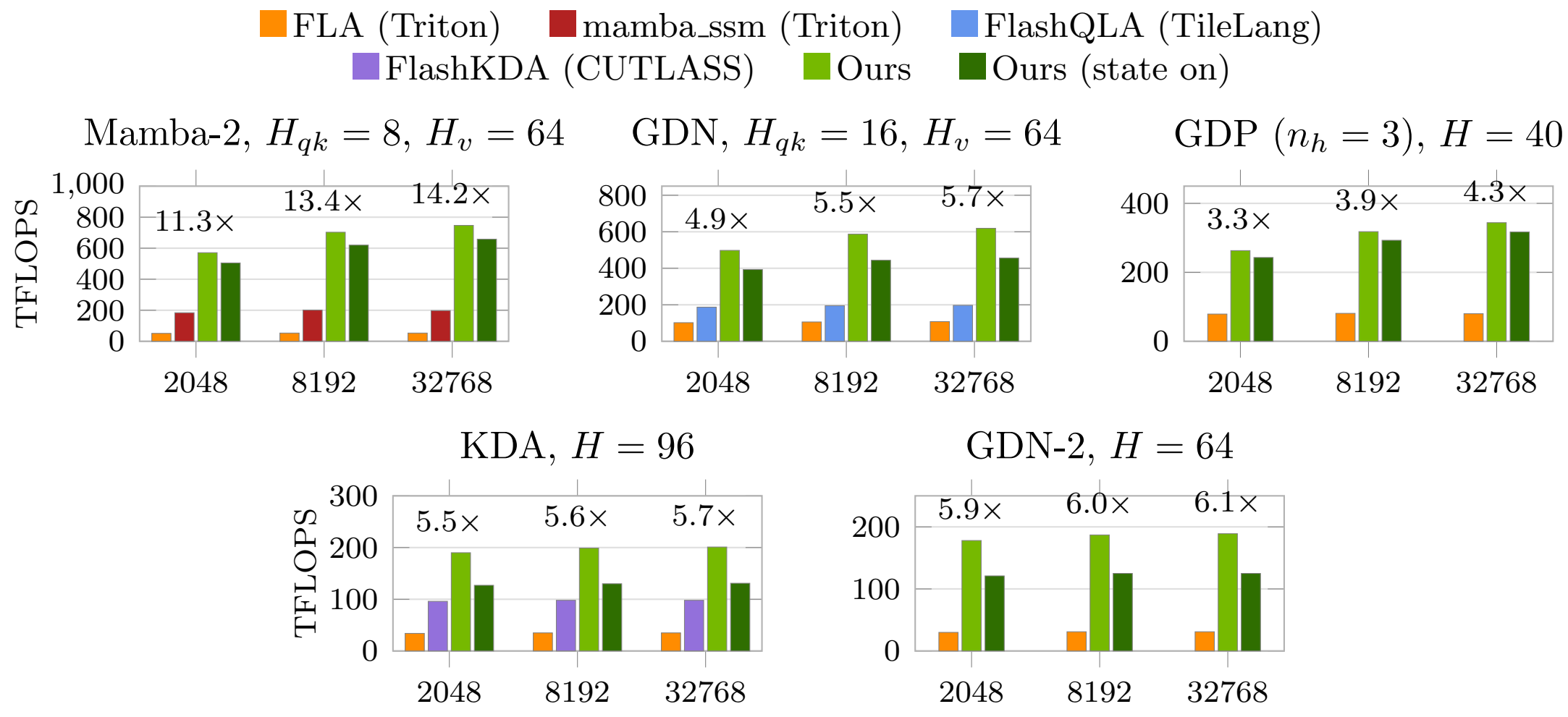

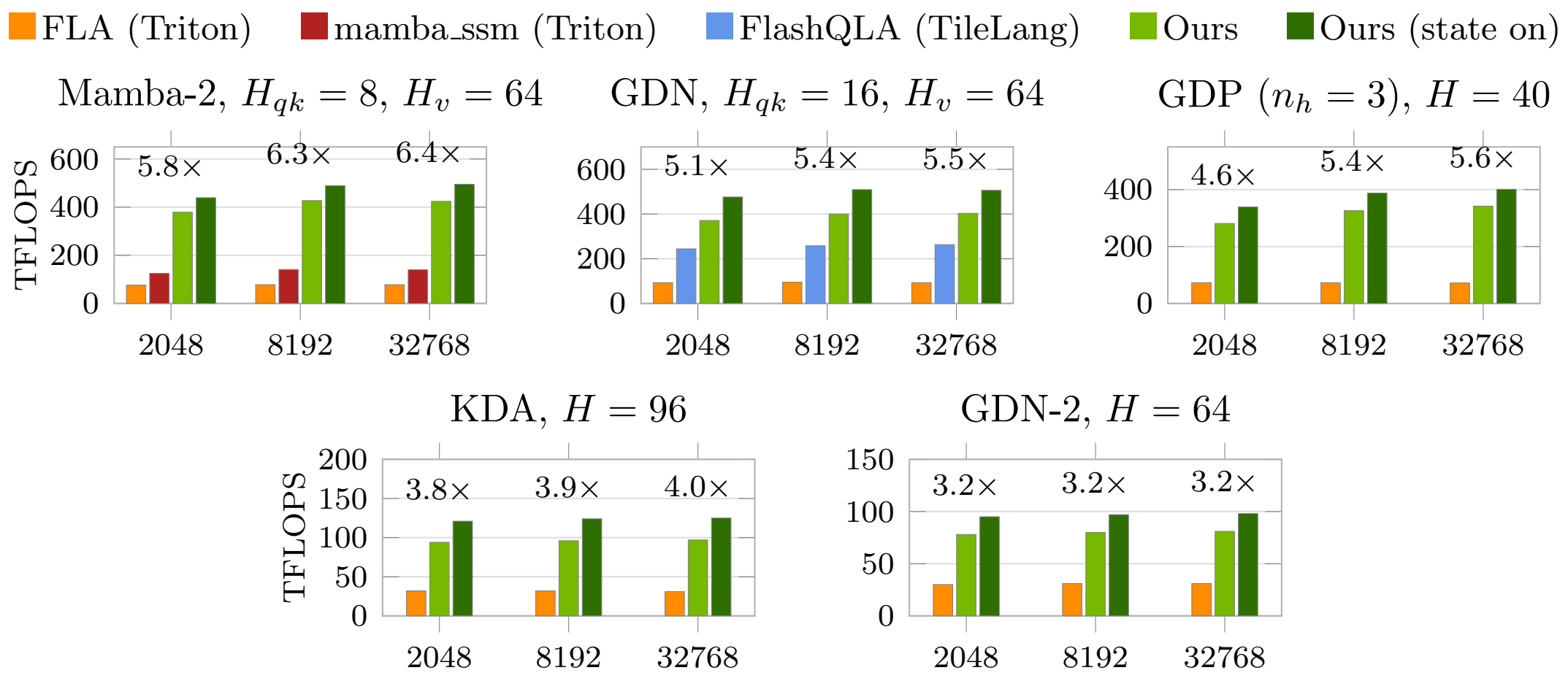
\begin{figure}[H]
\centering
\input{figures/results_bwd_b4}
\caption{Backward throughput at $B = 4$ on GB200 sweeping over the sequence length in tokens, at the same head counts and dimensions as Figure~\ref{fig:res-fwd-b4}. The label above each group is its speedup over FLA.}
\label{fig:res-bwd-b4}
\end{figure}

Next, we profile at shapes where the GPU is fully saturated, highlighting the performance of our main kernels. Figures~\ref{fig:res-fwd-b4} and~\ref{fig:res-bwd-b4} show the sweep over $s$ at $B = 4$. Across this sweep and the batch sweep at $s = 8192$, our forward runs $3.0\times$ to $15.9\times$ FLA's and the backward $2.1\times$ to $5.8\times$ without checkpoints and $2.9\times$ to $6.9\times$ with them. Against the other baselines, the GDN forward runs $2.7\times$ to $3.2\times$ FlashQLA's, and the GDN backward $1.4\times$ to $1.5\times$ without checkpoints and $1.9\times$ to $2.0\times$ with them. The KDA forward runs $1.8\times$ to $2.2\times$ FlashKDA's. Finally, the Mamba-2 forward runs $1.9\times$ to $4.2\times$ mamba\_ssm's, and the Mamba-2 backward $2.4\times$ to $3.2\times$ without checkpoints and $3.0\times$ to $3.8\times$ with them.

\subsection{Integration}

The optimized kernels are reachable through two entry points. The PyTorch custom ops located in \texttt{cudnn.linear\_attention.ops} are registered through \texttt{torch.library}, so they play nicely with autograd and \texttt{torch.compile}. Underneath them, the same operations are exposed through ordinary \cudnn{} graph nodes that are framework-agnostic and available directly to any user who builds their own graphs. We have also integrated the \cudnn{} engines as a backend of FlashInfer's linear attention APIs~\cite{ye2025flashinfer}, and the kernels have been integrated into Attention Gym as well.

\section{Conclusion and Future Work}
\label{sec:conclusion}

In this paper, we presented the \cudnn{} linear attention kernels for Mamba-2, GDN, GDP, KDA, and GDN-2, covering the training forward and backward passes on Blackwell. The design starts from one persistent kernel per pass that keeps the recurrent state in tensor memory in fp32 for the whole sequence, removing round trips through global memory.

We believe that further work could push the chunk size $C$ for the KDA and GDN-2 kernels to $32$ or higher, in a similar fashion to FLA. Section~\ref{sec:chunk} fixed $C = 16$ for the two per-channel variants because their decay-folded operands carry $\boldsymbol{\Gamma}_{[t]}$ and its reciprocal, and the cumulative decay of a channel is counted from the first token of the chunk. To further increase the chunk size, we can extend the anchor idea from FLA, which measures the cumulative decay within each diagonal $16 \times 16$ block from the middle row of that block rather than from its first token. Then, every row of a 32-token chunk lies within 16 tokens of the anchor, so every operand stays inside the same envelope. Increasing the chunk size should allow us to better saturate the MMAs for the forward and backward passes. Furthermore, it helps to reduce the memory traffic for the backward pass, as we only need the states at the chunk cadence. With Rubin's increased shared memory capacity, it should be possible to fit all intermediates in shared memory, while maintaining one fused kernel.

Furthermore, we believe that future work could explore using lower-precision data types for the MMA operations. All MMAs in this paper use 16-bit operands. In addition, the projections around the recurrence are large GEMMs that could overlap with the fused kernel on the SMs it leaves idle, instead of the strategies discussed in Section~\ref{sec:regimes}.

\section*{Acknowledgments}

We thank Sai Kishan Pampana for proposing the fused $[\mathbf{K};\mathbf{Q}]\mathbf{K}^{\trans}$ MMA idea. We thank Chaofan Yu for suggesting the reordering of the $d\mathbf{Q}$ MMA in the KDA and GDN-2 backward pipeline and other improvements. We thank Eric Alcaide Aldeano for various GDP performance improvements. We thank Driss Guessous for integrating the kernels into Attention Gym, so that they can be used in the PyTorch ecosystem. We would also like to thank Ali Hatamizadeh, Harris Nover, Shlok Agarwal, and members of the NVIDIA Fast Kernels and cuDNN teams for their thoughtful discussions and feedback on early drafts of the paper. Finally, we thank the GitHub community for their help with bug fixes and adoption of the kernels.

\bibliographystyle{plain}
\bibliography{references}

\clearpage
\appendix

\section{Backward Derivations}
\label{app:bwdmath}

This section of the appendix spells out the backward pass of the chunked forms from Section~\ref{sec:chunking}, as the fused backward kernels compute them. The walk runs from the last chunk to the first and carries $d\mathbf{S}_{[t]}$, the gradient with respect to the state leaving chunk $t$, starting from the caller's $d\mathbf{S}_{\text{end}}$ or from zero. The gates are parameterized in the log domain, $g_r = \log \alpha_r$ per token for GDN and GDP and $\mathbf{g}_r = \log \boldsymbol{\alpha}_r$ per token and key channel for KDA and GDN-2. Throughout, $\mathbf{e}_{C-1}$ is the indicator of the last row of the chunk, $\mathbf{1}_{n}$ is the all-ones vector of length $n$, and $\boldsymbol{\beta}_{[t]}^{-1}$ and $\boldsymbol{\Gamma}_{[t]}^{-1}$ are elementwise reciprocals.

\subsection{GDN and GDP}

With $\mathbf{T}_{[t]}$, $\mathbf{M}_{[t]}$, $\mathbf{A}_{[t]}$, and $\mathbf{L}_{[t]}$ as in Section~\ref{sec:chunking}, the backward equations for GDN and GDP are as follows.
\begin{align*}
d\mathbf{U}_{[t]} &= \mathbf{A}_{[t]}^{\trans} d\mathbf{O}_{[t]} + \boldsymbol{\Lambda}_{[t]} \odot \left(\mathbf{K}_{[t]}\, d\mathbf{S}_{[t]}\right) \\
d\mathbf{Y}_{[t]} &= \mathbf{T}_{[t]}^{\trans} d\mathbf{U}_{[t]} \\
d\mathbf{V}_{[t]} &= d\mathbf{Y}_{[t]} \\
d\mathbf{A}_{[t]} &= d\mathbf{O}_{[t]}\mathbf{U}_{[t]}^{\trans} \\
d\mathbf{M}_{[t]} &= \left(d\mathbf{Y}_{[t]}\mathbf{U}_{[t]}^{\trans}\right) \odot \mathbf{1}^{>} \\
\mathbf{E}_{[t]} &= d\mathbf{M}_{[t]} \odot \mathbf{L}_{[t]} \odot \mathbf{K}_{[t]}\mathbf{K}_{[t]}^{\trans} \\
d\mathbf{Q}_{[t]} &= \boldsymbol{\Gamma}_{[t]} \odot \left(d\mathbf{O}_{[t]}\mathbf{S}_{[t-1]}^{\trans}\right) + \left(d\mathbf{A}_{[t]} \odot \mathbf{L}_{[t]}\right)\mathbf{K}_{[t]} \\
d\mathbf{K}^{\text{output}}_{[t]} &= \left(d\mathbf{A}_{[t]} \odot \mathbf{L}_{[t]}\right)^{\trans}\mathbf{Q}_{[t]} \\
d\mathbf{K}^{\text{update}}_{[t]} &= \boldsymbol{\Lambda}_{[t]} \odot \left(\mathbf{U}_{[t]}\, d\mathbf{S}_{[t]}^{\trans}\right) \\
d\mathbf{K}_{[t]} &= d\mathbf{K}^{\text{output}}_{[t]} + d\mathbf{K}^{\text{update}}_{[t]} - \left(d\mathbf{M}_{[t]} \odot \mathbf{L}_{[t]} + \left(d\mathbf{M}_{[t]} \odot \mathbf{L}_{[t]}\right)^{\trans}\right)\mathbf{K}_{[t]} - \boldsymbol{\Gamma}_{[t]} \odot \left(d\mathbf{Y}_{[t]}\mathbf{S}_{[t-1]}^{\trans}\right) \\
d\boldsymbol{\beta}_{[t]} &= \boldsymbol{\beta}_{[t]}^{-1} \odot \left[\left(d\mathbf{Y}_{[t]} \odot \mathbf{Y}_{[t]}\right)\mathbf{1}_{d_v} - \mathbf{E}_{[t]}\mathbf{1}_{C}\right] \\
dg^{\text{last}}_{[t]} &= \gamma_{[t]}\, \tr\left(\mathbf{S}_{[t-1]}^{\trans} d\mathbf{S}_{[t]}\right) + \mathbf{1}_{C}^{\trans}\left(\mathbf{K}_{[t]} \odot d\mathbf{K}^{\text{update}}_{[t]}\right)\mathbf{1}_{d_{qk}} \\
d\mathbf{b}_{[t]} &= \left(\mathbf{Q}_{[t]} \odot d\mathbf{Q}_{[t]}\right)\mathbf{1}_{d_{qk}} - \left(\mathbf{K}_{[t]} \odot \left(d\mathbf{K}^{\text{output}}_{[t]} + d\mathbf{K}^{\text{update}}_{[t]}\right)\right)\mathbf{1}_{d_{qk}} \\
&\qquad - \left(\left(\mathbf{V}_{[t]} - \mathbf{Y}_{[t]}\right) \odot d\mathbf{Y}_{[t]}\right)\mathbf{1}_{d_v} - \mathbf{E}_{[t]}\mathbf{1}_{C} + \mathbf{E}_{[t]}^{\trans}\mathbf{1}_{C} + dg^{\text{last}}_{[t]}\, \mathbf{e}_{C-1} \\
d\mathbf{g}_{[t]} &= \left(\mathbf{1}^{\ge}\right)^{\trans} d\mathbf{b}_{[t]} \\
d\mathbf{S}_{[t-1]} &= \gamma_{[t]}\, d\mathbf{S}_{[t]} + \mathbf{Q}_{[t]}^{\trans}\left(\boldsymbol{\Gamma}_{[t]} \odot d\mathbf{O}_{[t]}\right) - \mathbf{K}_{[t]}^{\trans}\left(\boldsymbol{\Gamma}_{[t]} \odot d\mathbf{Y}_{[t]}\right)
\end{align*}

\subsection{KDA}

We use the same notation as introduced in Section~\ref{sec:chunking}. The KDA backward equations are as follows.
\begin{align*}
d\mathbf{U}_{[t]} &= \mathbf{A}_{[t]}^{\trans} d\mathbf{O}_{[t]} + \widetilde{\mathbf{K}}_{[t]}\, d\mathbf{S}_{[t]} \\
d\mathbf{Y}_{[t]} &= \mathbf{T}_{[t]}^{\trans} d\mathbf{U}_{[t]} \\
d\mathbf{V}_{[t]} &= d\mathbf{Y}_{[t]} \\
d\mathbf{A}_{[t]} &= d\mathbf{O}_{[t]}\mathbf{U}_{[t]}^{\trans} \\
d\mathbf{M}_{[t]} &= \left(d\mathbf{Y}_{[t]}\mathbf{U}_{[t]}^{\trans}\right) \odot \mathbf{1}^{>} \\
d\widehat{\mathbf{Q}}_{[t]} &= d\mathbf{O}_{[t]}\mathbf{S}_{[t-1]}^{\trans} + \left(d\mathbf{A}_{[t]} \odot \mathbf{1}^{\ge}\right)\overline{\mathbf{K}}_{[t]} \\
d\overline{\mathbf{K}}_{[t]} &= \left(d\mathbf{A}_{[t]} \odot \mathbf{1}^{\ge}\right)^{\trans}\widehat{\mathbf{Q}}_{[t]} - d\mathbf{M}_{[t]}^{\trans}\widehat{\mathbf{K}}_{[t]} \\
d\widehat{\mathbf{K}}_{[t]} &= -\left(d\mathbf{Y}_{[t]}\mathbf{S}_{[t-1]}^{\trans} + d\mathbf{M}_{[t]}\overline{\mathbf{K}}_{[t]}\right) \\
d\widetilde{\mathbf{K}}_{[t]} &= \mathbf{U}_{[t]}\, d\mathbf{S}_{[t]}^{\trans} \\
d\mathbf{Q}_{[t]} &= \boldsymbol{\Gamma}_{[t]} \odot d\widehat{\mathbf{Q}}_{[t]} \\
d\mathbf{K}_{[t]} &= \boldsymbol{\Gamma}_{[t]} \odot d\widehat{\mathbf{K}}_{[t]} + \boldsymbol{\Gamma}_{[t]}^{-1} \odot d\overline{\mathbf{K}}_{[t]} + \boldsymbol{\Lambda}_{[t]} \odot d\widetilde{\mathbf{K}}_{[t]} \\
d\boldsymbol{\gamma}_{[t]} &= \left(d\mathbf{S}_{[t]} \odot \mathbf{S}_{[t-1]}\right)\mathbf{1}_{d_v} \\
d\boldsymbol{\beta}_{[t]} &= \boldsymbol{\beta}_{[t]}^{-1} \odot \left[\left(d\mathbf{Y}_{[t]} \odot \mathbf{Y}_{[t]}\right)\mathbf{1}_{d_v} - \left(d\mathbf{M}_{[t]} \odot \widehat{\mathbf{K}}_{[t]}\overline{\mathbf{K}}_{[t]}^{\trans}\right)\mathbf{1}_{C}\right] \\
d\mathbf{g}^{\text{last}}_{[t]} &= \boldsymbol{\gamma}_{[t]} \odot d\boldsymbol{\gamma}_{[t]} + \left(\mathbf{K}_{[t]} \odot \boldsymbol{\Lambda}_{[t]} \odot d\widetilde{\mathbf{K}}_{[t]}\right)^{\trans}\mathbf{1}_{C} \\
d\mathbf{b}_{[t]} &= \mathbf{Q}_{[t]} \odot \boldsymbol{\Gamma}_{[t]} \odot d\widehat{\mathbf{Q}}_{[t]} + \mathbf{K}_{[t]} \odot \boldsymbol{\Gamma}_{[t]} \odot d\widehat{\mathbf{K}}_{[t]} - \mathbf{K}_{[t]} \odot \boldsymbol{\Gamma}_{[t]}^{-1} \odot d\overline{\mathbf{K}}_{[t]} - \mathbf{K}_{[t]} \odot \boldsymbol{\Lambda}_{[t]} \odot d\widetilde{\mathbf{K}}_{[t]} \\
&\qquad + \mathbf{e}_{C-1}\left(d\mathbf{g}^{\text{last}}_{[t]}\right)^{\trans} \\
d\mathbf{g}_{[t]} &= \left(\mathbf{1}^{\ge}\right)^{\trans} d\mathbf{b}_{[t]} \\
d\mathbf{S}_{[t-1]} &= \diag\left(\boldsymbol{\gamma}_{[t]}\right) d\mathbf{S}_{[t]} + \widehat{\mathbf{Q}}_{[t]}^{\trans} d\mathbf{O}_{[t]} - \widehat{\mathbf{K}}_{[t]}^{\trans} d\mathbf{Y}_{[t]}
\end{align*}

\subsection{GDN-2}

GDN-2 carries the per-channel erase gate inside $\widehat{\mathbf{K}}_{[t]} = \boldsymbol{\beta}_{[t]} \odot \mathbf{K}_{[t]} \odot \boldsymbol{\Gamma}_{[t]}$, writes $\mathbf{w}_{[t]} \odot \mathbf{V}_{[t]}$ in place of $\mathbf{V}_{[t]}$, and inverts the plain $\mathbf{M}_{[t]} = \mathbf{I} + \mathbf{1}^{>} \odot \widehat{\mathbf{K}}_{[t]}\overline{\mathbf{K}}_{[t]}^{\trans}$, so $\mathbf{T}_{[t]} = \mathbf{M}_{[t]}^{-1}$. Let $\mathbf{K}^{+}_{[t]} = \mathbf{K}_{[t]} \odot \boldsymbol{\Gamma}_{[t]}$. The GDN-2 backward equations are as follows.
\begin{align*}
d\mathbf{U}_{[t]} &= \mathbf{A}_{[t]}^{\trans} d\mathbf{O}_{[t]} + \widetilde{\mathbf{K}}_{[t]}\, d\mathbf{S}_{[t]} \\
d\mathbf{Y}_{[t]} &= \mathbf{T}_{[t]}^{\trans} d\mathbf{U}_{[t]} \\
d\mathbf{V}_{[t]} &= \mathbf{w}_{[t]} \odot d\mathbf{Y}_{[t]} \\
d\mathbf{w}_{[t]} &= \mathbf{V}_{[t]} \odot d\mathbf{Y}_{[t]} \\
d\mathbf{A}_{[t]} &= d\mathbf{O}_{[t]}\mathbf{U}_{[t]}^{\trans} \\
d\mathbf{M}_{[t]} &= \left(d\mathbf{Y}_{[t]}\mathbf{U}_{[t]}^{\trans}\right) \odot \mathbf{1}^{>} \\
d\widehat{\mathbf{Q}}_{[t]} &= d\mathbf{O}_{[t]}\mathbf{S}_{[t-1]}^{\trans} + \left(d\mathbf{A}_{[t]} \odot \mathbf{1}^{\ge}\right)\overline{\mathbf{K}}_{[t]} \\
d\overline{\mathbf{K}}_{[t]} &= \left(d\mathbf{A}_{[t]} \odot \mathbf{1}^{\ge}\right)^{\trans}\widehat{\mathbf{Q}}_{[t]} - d\mathbf{M}_{[t]}^{\trans}\widehat{\mathbf{K}}_{[t]} \\
d\widehat{\mathbf{K}}_{[t]} &= -\left(d\mathbf{Y}_{[t]}\mathbf{S}_{[t-1]}^{\trans} + d\mathbf{M}_{[t]}\overline{\mathbf{K}}_{[t]}\right) \\
d\widetilde{\mathbf{K}}_{[t]} &= \mathbf{U}_{[t]}\, d\mathbf{S}_{[t]}^{\trans} \\
d\boldsymbol{\beta}_{[t]} &= \mathbf{K}^{+}_{[t]} \odot d\widehat{\mathbf{K}}_{[t]} \\
d\mathbf{Q}_{[t]} &= \boldsymbol{\Gamma}_{[t]} \odot d\widehat{\mathbf{Q}}_{[t]} \\
d\mathbf{K}_{[t]} &= \left(\boldsymbol{\beta}_{[t]} \odot \boldsymbol{\Gamma}_{[t]}\right) \odot d\widehat{\mathbf{K}}_{[t]} + \boldsymbol{\Gamma}_{[t]}^{-1} \odot d\overline{\mathbf{K}}_{[t]} + \boldsymbol{\Lambda}_{[t]} \odot d\widetilde{\mathbf{K}}_{[t]} \\
d\boldsymbol{\gamma}_{[t]} &= \left(d\mathbf{S}_{[t]} \odot \mathbf{S}_{[t-1]}\right)\mathbf{1}_{d_v} \\
d\mathbf{g}^{\text{last}}_{[t]} &= \boldsymbol{\gamma}_{[t]} \odot d\boldsymbol{\gamma}_{[t]} + \left(\mathbf{K}_{[t]} \odot \boldsymbol{\Lambda}_{[t]} \odot d\widetilde{\mathbf{K}}_{[t]}\right)^{\trans}\mathbf{1}_{C} \\
d\mathbf{b}_{[t]} &= \mathbf{Q}_{[t]} \odot \boldsymbol{\Gamma}_{[t]} \odot d\widehat{\mathbf{Q}}_{[t]} + \widehat{\mathbf{K}}_{[t]} \odot d\widehat{\mathbf{K}}_{[t]} - \overline{\mathbf{K}}_{[t]} \odot d\overline{\mathbf{K}}_{[t]} - \mathbf{K}_{[t]} \odot \boldsymbol{\Lambda}_{[t]} \odot d\widetilde{\mathbf{K}}_{[t]} \\
&\qquad + \mathbf{e}_{C-1}\left(d\mathbf{g}^{\text{last}}_{[t]}\right)^{\trans} \\
d\mathbf{g}_{[t]} &= \left(\mathbf{1}^{\ge}\right)^{\trans} d\mathbf{b}_{[t]} \\
d\mathbf{S}_{[t-1]} &= \diag\left(\boldsymbol{\gamma}_{[t]}\right) d\mathbf{S}_{[t]} + \widehat{\mathbf{Q}}_{[t]}^{\trans} d\mathbf{O}_{[t]} - \widehat{\mathbf{K}}_{[t]}^{\trans} d\mathbf{Y}_{[t]}
\end{align*}
Note that every line that does not involve $\boldsymbol{\beta}_{[t]}$ or $\mathbf{w}_{[t]}$ is the same as KDA. $d\boldsymbol{\beta}_{[t]} \in \mathbb{R}^{C \times d_{qk}}$ and $d\mathbf{w}_{[t]} \in \mathbb{R}^{C \times d_v}$ are now per-token vectors, and $d\mathbf{M}_{[t]}$ needs no $\boldsymbol{\beta}_{[t]}$ since $\mathbf{T}_{[t]}$ is the plain inverse.

\section{Backward Pipelines}
\label{app:bwdpipe}

The backward kernels are warp specialized in the same way as the prefill kernels discussed in Section~\ref{sec:pipeline}. One CTA walks the chunks of one \texttt{(batch, head)} pair from the last to the first, carries $d\mathbf{S}$ in TMEM in fp32, reloads each chunk's entering state from the checkpoint series, recomputes $\mathbf{T}_{[t]}$, $\mathbf{A}_{[t]}$, $\mathbf{Y}_{[t]}$, and $\mathbf{U}_{[t]}$ from it, and then issues the gradient MMAs of Appendix~\ref{app:bwdmath}.

\subsection{GDN and GDP}

The GDN and GDP backward runs sixteen warps with their functions discussed in Table~\ref{tab:warps-bwd}.

\begin{table}[H]
\centering
\small
\begin{tabular}{@{}l p{13.2cm}@{}}
\toprule
Warps & Role \\
\midrule
0--3 & compute group 0: runs the $\mathbf{K}\mathbf{K}^{\trans}$ and $\mathbf{Q}\mathbf{K}^{\trans}$ epilogues and the $64 \times 64$ blockwise inverse that yields $\mathbf{T}$, and reduces the $\mathbf{M}$ terms of $d\boldsymbol{\beta}$ and $d\mathbf{g}$ and the state term of $d\mathbf{g}$ \\
4--7 & compute group 1: rescales $d\mathbf{S}$, runs the $d\mathbf{V}$ and $d\mathbf{Q}$ epilogues, and reduces the $\mathbf{Y}$ terms of $d\boldsymbol{\beta}$ and $d\mathbf{g}$ and the $\mathbf{Q}$ terms of $d\mathbf{g}$ \\
8--11 & compute group 2: rescales the inter-chunk $d\mathbf{K}$, reduces the $\mathbf{K}$ terms of $d\mathbf{g}$, and runs the $d\mathbf{K}$ epilogue \\
12 & MMA warp: issues every \texttt{tcgen05} MMA \\
13 & TMA warp: loads the $\mathbf{Q}$, $\mathbf{K}$, $\mathbf{V}$, and $d\mathbf{O}$ tiles and the entering state from the checkpoint series \\
14 & gate warp: loads the gate logits and $\boldsymbol{\beta}$ and stores $d\mathbf{g}$ and $d\boldsymbol{\beta}$ \\
15 & epilogue warp: stores $d\mathbf{Q}$, $d\mathbf{K}$, and $d\mathbf{V}$ \\
\bottomrule
\end{tabular}
\caption{Warp roles of the GDN and GDP backward kernel.}
\label{tab:warps-bwd}
\end{table}

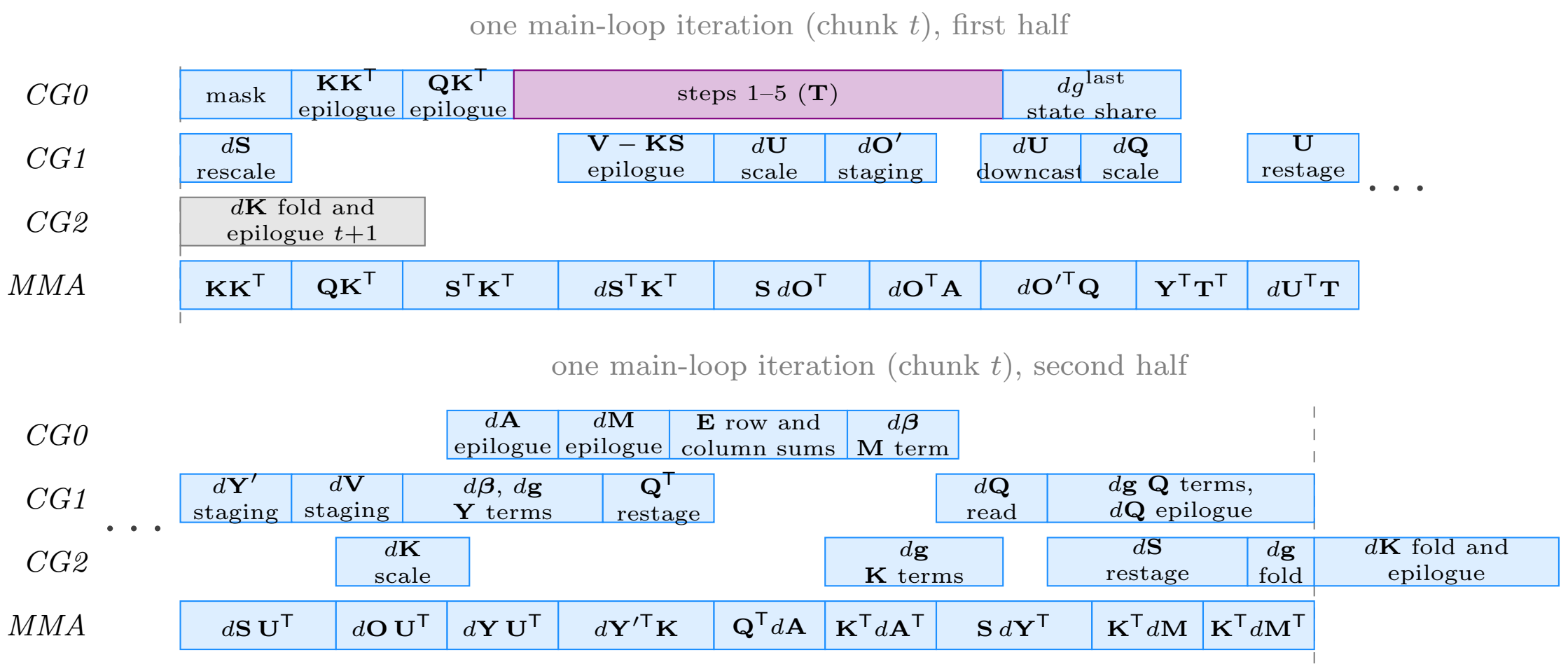
\begin{figure}[H]
\centering
\begin{tikzpicture}[x=1.050cm, y=0.60cm, every node/.style={font=\tiny}]
    \path (-1.667, -0.6) rectangle (13.164, 4.05);
    \draw[dashed, gray] (0.000, -0.6) -- (0.000, 3.60);
    \node[anchor=south, font=\scriptsize, text=gray] at (5.300, 3.60) {one main-loop iteration (chunk $t$), first half};
    \node[anchor=west, font=\Large, inner sep=0pt, text=gray!50!black] at (10.650, 1.50) {$\cdots$};
    \node[anchor=east, font=\scriptsize\itshape] at (-0.714, 3) {CG0};
    \node[anchor=east, font=\scriptsize\itshape] at (-0.714, 2) {CG1};
    \node[anchor=east, font=\scriptsize\itshape] at (-0.714, 1) {CG2};
    \node[anchor=east, font=\scriptsize\itshape] at (-0.714, 0) {MMA};
    \draw[fill=qcolor!15, draw=qcolor] (0.000,-0.38) rectangle (1.000,0.38);
    \node[] at (0.500,0) {$\mathbf{K}\mathbf{K}^{\trans}$};
    \draw[fill=qcolor!15, draw=qcolor] (1.000,-0.38) rectangle (2.000,0.38);
    \node[] at (1.500,0) {$\mathbf{Q}\mathbf{K}^{\trans}$};
    \draw[fill=qcolor!15, draw=qcolor] (2.000,-0.38) rectangle (3.400,0.38);
    \node[] at (2.700,0) {$\mathbf{S}^{\trans}\mathbf{K}^{\trans}$};
    \draw[fill=qcolor!15, draw=qcolor] (3.400,-0.38) rectangle (4.800,0.38);
    \node[] at (4.100,0) {$d\mathbf{S}^{\trans}\mathbf{K}^{\trans}$};
    \draw[fill=qcolor!15, draw=qcolor] (4.800,-0.38) rectangle (6.200,0.38);
    \node[] at (5.500,0) {$\mathbf{S}\,d\mathbf{O}^{\trans}$};
    \draw[fill=qcolor!15, draw=qcolor] (6.200,-0.38) rectangle (7.200,0.38);
    \node[] at (6.700,0) {$d\mathbf{O}^{\trans}\mathbf{A}$};
    \draw[fill=qcolor!15, draw=qcolor] (7.200,-0.38) rectangle (8.600,0.38);
    \node[] at (7.900,0) {$d\mathbf{O}'^{\trans}\mathbf{Q}$};
    \draw[fill=qcolor!15, draw=qcolor] (8.600,-0.38) rectangle (9.600,0.38);
    \node[] at (9.100,0) {$\mathbf{Y}^{\trans}\mathbf{T}^{\trans}$};
    \draw[fill=qcolor!15, draw=qcolor] (9.600,-0.38) rectangle (10.600,0.38);
    \node[] at (10.100,0) {$d\mathbf{U}^{\trans}\mathbf{T}$};
    \draw[fill=qcolor!15, draw=qcolor] (0.000,2.62) rectangle (1.000,3.38);
    \node[align=center] at (0.500,3) {mask};
    \draw[fill=qcolor!15, draw=qcolor] (1.000,2.62) rectangle (2.000,3.38);
    \node[align=center] at (1.500,3) {$\mathbf{K}\mathbf{K}^{\trans}$\\epilogue};
    \draw[fill=qcolor!15, draw=qcolor] (2.000,2.62) rectangle (3.000,3.38);
    \node[align=center] at (2.500,3) {$\mathbf{Q}\mathbf{K}^{\trans}$\\epilogue};
    \draw[fill=statecolor!25, draw=statecolor] (3.000,2.62) rectangle (7.400,3.38);
    \node[] at (5.200,3) {steps 1--5 ($\mathbf{T}$)};
    \draw[fill=qcolor!15, draw=qcolor] (7.400,2.62) rectangle (9.000,3.38);
    \node[align=center] at (8.200,3) {$dg^{\text{last}}$\\state share};
    \draw[fill=qcolor!15, draw=qcolor] (0.000,1.62) rectangle (1.000,2.38);
    \node[align=center] at (0.500,2) {$d\mathbf{S}$\\rescale};
    \draw[fill=qcolor!15, draw=qcolor] (3.400,1.62) rectangle (4.800,2.38);
    \node[align=center] at (4.100,2) {$\mathbf{V} - \mathbf{K}\mathbf{S}$\\epilogue};
    \draw[fill=qcolor!15, draw=qcolor] (4.800,1.62) rectangle (5.800,2.38);
    \node[align=center] at (5.300,2) {$d\mathbf{U}$\\scale};
    \draw[fill=qcolor!15, draw=qcolor] (5.800,1.62) rectangle (6.800,2.38);
    \node[align=center] at (6.300,2) {$d\mathbf{O}'$\\staging};
    \draw[fill=qcolor!15, draw=qcolor] (7.200,1.62) rectangle (8.100,2.38);
    \node[align=center] at (7.650,2) {$d\mathbf{U}$\\downcast};
    \draw[fill=qcolor!15, draw=qcolor] (8.100,1.62) rectangle (9.000,2.38);
    \node[align=center] at (8.550,2) {$d\mathbf{Q}$\\scale};
    \draw[fill=qcolor!15, draw=qcolor] (9.600,1.62) rectangle (10.600,2.38);
    \node[align=center] at (10.100,2) {$\mathbf{U}$\\restage};
    \draw[fill=gray!20, draw=gray] (0.000,0.62) rectangle (2.200,1.38);
    \node[align=center] at (1.100,1) {$d\mathbf{K}$ fold and\\epilogue $t{+}1$};
\end{tikzpicture}

\smallskip

\begin{tikzpicture}[x=1.050cm, y=0.60cm, every node/.style={font=\tiny}]
    \path (-1.667, -0.6) rectangle (13.164, 4.05);
    \draw[dashed, gray] (10.200, -0.6) -- (10.200, 3.60);
    \node[anchor=south, font=\scriptsize, text=gray] at (6.200, 3.60) {one main-loop iteration (chunk $t$), second half};
    \node[anchor=east, font=\Large, inner sep=0pt, text=gray!50!black] at (-0.050, 1.50) {$\cdots$};
    \node[anchor=east, font=\scriptsize\itshape] at (-0.714, 3) {CG0};
    \node[anchor=east, font=\scriptsize\itshape] at (-0.714, 2) {CG1};
    \node[anchor=east, font=\scriptsize\itshape] at (-0.714, 1) {CG2};
    \node[anchor=east, font=\scriptsize\itshape] at (-0.714, 0) {MMA};
    \draw[fill=qcolor!15, draw=qcolor] (0.000,-0.38) rectangle (1.400,0.38);
    \node[] at (0.700,0) {$d\mathbf{S}\,\mathbf{U}^{\trans}$};
    \draw[fill=qcolor!15, draw=qcolor] (1.400,-0.38) rectangle (2.400,0.38);
    \node[] at (1.900,0) {$d\mathbf{O}\,\mathbf{U}^{\trans}$};
    \draw[fill=qcolor!15, draw=qcolor] (2.400,-0.38) rectangle (3.400,0.38);
    \node[] at (2.900,0) {$d\mathbf{Y}\,\mathbf{U}^{\trans}$};
    \draw[fill=qcolor!15, draw=qcolor] (3.400,-0.38) rectangle (4.800,0.38);
    \node[] at (4.100,0) {$d\mathbf{Y}'^{\trans}\mathbf{K}$};
    \draw[fill=qcolor!15, draw=qcolor] (4.800,-0.38) rectangle (5.800,0.38);
    \node[] at (5.300,0) {$\mathbf{Q}^{\trans}d\mathbf{A}$};
    \draw[fill=qcolor!15, draw=qcolor] (5.800,-0.38) rectangle (6.800,0.38);
    \node[] at (6.300,0) {$\mathbf{K}^{\trans}d\mathbf{A}^{\trans}$};
    \draw[fill=qcolor!15, draw=qcolor] (6.800,-0.38) rectangle (8.200,0.38);
    \node[] at (7.500,0) {$\mathbf{S}\,d\mathbf{Y}^{\trans}$};
    \draw[fill=qcolor!15, draw=qcolor] (8.200,-0.38) rectangle (9.200,0.38);
    \node[] at (8.700,0) {$\mathbf{K}^{\trans}d\mathbf{M}$};
    \draw[fill=qcolor!15, draw=qcolor] (9.200,-0.38) rectangle (10.200,0.38);
    \node[] at (9.700,0) {$\mathbf{K}^{\trans}d\mathbf{M}^{\trans}$};
    \draw[fill=qcolor!15, draw=qcolor] (2.400,2.62) rectangle (3.400,3.38);
    \node[align=center] at (2.900,3) {$d\mathbf{A}$\\epilogue};
    \draw[fill=qcolor!15, draw=qcolor] (3.400,2.62) rectangle (4.400,3.38);
    \node[align=center] at (3.900,3) {$d\mathbf{M}$\\epilogue};
    \draw[fill=qcolor!15, draw=qcolor] (4.400,2.62) rectangle (6.000,3.38);
    \node[align=center] at (5.200,3) {$\mathbf{E}$ row and\\column sums};
    \draw[fill=qcolor!15, draw=qcolor] (6.000,2.62) rectangle (7.000,3.38);
    \node[align=center] at (6.500,3) {$d\boldsymbol{\beta}$\\$\mathbf{M}$ term};
    \draw[fill=qcolor!15, draw=qcolor] (0.000,1.62) rectangle (1.000,2.38);
    \node[align=center] at (0.500,2) {$d\mathbf{Y}'$\\staging};
    \draw[fill=qcolor!15, draw=qcolor] (1.000,1.62) rectangle (2.000,2.38);
    \node[align=center] at (1.500,2) {$d\mathbf{V}$\\staging};
    \draw[fill=qcolor!15, draw=qcolor] (2.000,1.62) rectangle (3.800,2.38);
    \node[align=center] at (2.900,2) {$d\boldsymbol{\beta}$, $d\mathbf{g}$\\$\mathbf{Y}$ terms};
    \draw[fill=qcolor!15, draw=qcolor] (3.800,1.62) rectangle (4.800,2.38);
    \node[align=center] at (4.300,2) {$\mathbf{Q}^{\trans}$\\restage};
    \draw[fill=qcolor!15, draw=qcolor] (6.800,1.62) rectangle (7.800,2.38);
    \node[align=center] at (7.300,2) {$d\mathbf{Q}$\\read};
    \draw[fill=qcolor!15, draw=qcolor] (7.800,1.62) rectangle (10.200,2.38);
    \node[align=center] at (9.000,2) {$d\mathbf{g}$ $\mathbf{Q}$ terms,\\$d\mathbf{Q}$ epilogue};
    \draw[fill=qcolor!15, draw=qcolor] (1.400,0.62) rectangle (2.600,1.38);
    \node[align=center] at (2.000,1) {$d\mathbf{K}$\\scale};
    \draw[fill=qcolor!15, draw=qcolor] (5.800,0.62) rectangle (7.400,1.38);
    \node[align=center] at (6.600,1) {$d\mathbf{g}$\\$\mathbf{K}$ terms};
    \draw[fill=qcolor!15, draw=qcolor] (7.800,0.62) rectangle (9.600,1.38);
    \node[align=center] at (8.700,1) {$d\mathbf{S}$\\restage};
    \draw[fill=qcolor!15, draw=qcolor] (9.600,0.62) rectangle (10.200,1.38);
    \node[align=center] at (9.900,1) {$d\mathbf{g}$\\fold};
    \draw[fill=qcolor!15, draw=qcolor] (10.200,0.62) rectangle (12.400,1.38);
    \node[align=center] at (11.300,1) {$d\mathbf{K}$ fold and\\epilogue};
\end{tikzpicture}
\caption{The GDN and GDP backward pipeline. Blue is chunk $t$, gray chunk $t{+}1$, and purple the blockwise inverse.}
\label{fig:pipeline-bwd}
\end{figure}

Figure~\ref{fig:pipeline-bwd} shows one main-loop iteration of its schedule. The MMA warp issues the eighteen MMAs of a chunk in the same swap-AB orientation presented in Section~\ref{sec:lanes}. Again, the goal is to keep the MMA warp as busy as possible.

Similar to prefill, compute group 0 builds the $\mathbf{L}_{[t]}$ masks, scales $\mathbf{K}\mathbf{K}^{\trans}$, $\mathbf{Q}\mathbf{K}^{\trans}$, $d\mathbf{A}$, and $d\mathbf{M}$ into their shared memory operands, computes the inverse, and reduces the state share of $dg^{\text{last}}_{[t]}$ and the tile $\mathbf{E}_{[t]}$ of Appendix~\ref{app:bwdmath} for $d\boldsymbol{\beta}$ and $d\mathbf{g}$. Note that we do not invert two chunks at a time as described in Section~\ref{sec:pairinv}, because the eighteen MMAs per chunk make it hard to fit a second chunk's accumulators and operands in the TMEM or shared memory capacity. Compute group 1 rescales $d\mathbf{S}$, $d\mathbf{U}$, and $d\mathbf{Q}$ in TMEM, downcasts $d\mathbf{U}$, restages $\mathbf{U}$ over $\mathbf{V}$, stages $d\mathbf{V} = d\mathbf{Y}$, and owns the $\mathbf{Y}$ and $\mathbf{Q}$ terms of $d\boldsymbol{\beta}$ and $d\mathbf{g}$ and the $d\mathbf{Q}$ epilogue. Compute group 2 rescales the inter-chunk $d\mathbf{K}$, owns the $\mathbf{K}$ terms of $d\mathbf{g}$, restages the finished $d\mathbf{S}$ for the next chunk, and folds and stores $d\mathbf{K}$. The gate warp finishes $d\mathbf{g}$ with the reverse cumulative sum and stores $d\mathbf{g}$ and $d\boldsymbol{\beta}$, and the epilogue warp stores $d\mathbf{V}$, $d\mathbf{Q}$, and $d\mathbf{K}$.

\subsection{KDA and GDN-2}
As in the KDA and GDN-2 prefill, the backward keeps the two register-MMA warps and runs three compute groups as shown in Table~\ref{tab:warps-bwd-kda}.

\begin{table}[H]
\centering
\small
\begin{tabular}{@{}l p{13.2cm}@{}}
\toprule
Warps & Role \\
\midrule
0--3 & compute group 0: runs the per-channel gate prefix scan and computes the decay-folded operands $\widehat{\mathbf{Q}}$, $\widehat{\mathbf{K}}$, and $\overline{\mathbf{K}}$ \\
4--7 & compute group 1: rescales $d\mathbf{S}$, stages $\mathbf{Y}$, $d\mathbf{U}$, $\mathbf{U}$, $d\mathbf{Y}$, and $-\boldsymbol{\beta} \odot d\mathbf{Y}$, runs the $d\mathbf{V}$ epilogue (and the $d\mathbf{w}$ epilogue for GDN-2), and reduces and stores $d\boldsymbol{\beta}$ for KDA \\
8--11 & compute group 2: forms the $d\boldsymbol{\gamma}$ dot products and runs the $d\mathbf{Q}$, $d\mathbf{K}$, and $d\mathbf{g}$ epilogues (and the per-channel $d\boldsymbol{\beta}$ epilogue for GDN-2) with the $d\mathbf{A}$ diagonal, the L2-norm backward, and the reverse cumulative sum \\
12 & register-MMA warp: computes $\widehat{\mathbf{K}}\overline{\mathbf{K}}^{\trans}$, the blockwise inverse that yields $\mathbf{T}$, $d\mathbf{M}$, and the $\boldsymbol{\beta}$ term of $d\boldsymbol{\beta}$ \\
13 & MMA warp: issues every \texttt{tcgen05} MMA \\
14 & TMA warp: loads the $\mathbf{Q}$, $\mathbf{K}$, $\mathbf{V}$, $d\mathbf{O}$, and gate tiles and the entering state from the checkpoint series \\
15 & epilogue warp: computes $\mathbf{A}$ and $d\mathbf{A}$ on register MMAs and stores $d\mathbf{Q}$, $d\mathbf{K}$, $d\mathbf{V}$, and $d\mathbf{g}$ (and $d\boldsymbol{\beta}$ and $d\mathbf{w}$ for GDN-2) \\
\bottomrule
\end{tabular}
\caption{Warp roles of the KDA and GDN-2 backward kernel.}
\label{tab:warps-bwd-kda}
\end{table}

\begin{figure}[H]
\centering
\begin{tikzpicture}[x=1.200cm, y=0.55cm, every node/.style={font=\tiny}]
    \path (-1.250, -0.6) rectangle (12.186, 6.05);
    \draw[dashed, gray] (0.000, -0.6) -- (0.000, 5.60);
    \node[anchor=south, font=\scriptsize, text=gray] at (5.800, 5.60) {one main-loop iteration (chunk $t$), first half};
    \node[anchor=west, font=\Large, inner sep=0pt, text=gray!50!black] at (11.650, 2.50) {$\cdots$};
    \node[anchor=east, font=\scriptsize\itshape] at (-0.536, 5) {CG0};
    \node[anchor=east, font=\scriptsize\itshape] at (-0.536, 4) {CG1};
    \node[anchor=east, font=\scriptsize\itshape] at (-0.536, 3) {CG2};
    \node[anchor=east, font=\scriptsize\itshape] at (-0.536, 2) {Inverse};
    \node[anchor=east, font=\scriptsize\itshape] at (-0.536, 1) {Output};
    \node[anchor=east, font=\scriptsize\itshape] at (-0.536, 0) {MMA};
    \draw[fill=qcolor!15, draw=qcolor] (1.300,-0.42) rectangle (2.600,0.42);
    \node[] at (1.950,0) {$\mathbf{S}^{\trans}\widehat{\mathbf{K}}^{\trans}$};
    \draw[fill=qcolor!15, draw=qcolor] (9.000,-0.42) rectangle (10.300,0.42);
    \node[] at (9.650,0) {$\mathbf{S}\,d\mathbf{O}^{\trans}$};
    \draw[fill=qcolor!15, draw=qcolor] (2.600,-0.42) rectangle (3.900,0.42);
    \node[] at (3.250,0) {$d\mathbf{S}'^{\trans}\overline{\mathbf{K}}^{\trans}$};
    \draw[fill=qcolor!15, draw=qcolor] (3.900,-0.42) rectangle (4.700,0.42);
    \node[] at (4.300,0) {$d\mathbf{O}^{\trans}\mathbf{A}$};
    \draw[fill=qcolor!15, draw=qcolor] (4.700,-0.42) rectangle (6.000,0.42);
    \node[] at (5.350,0) {$d\mathbf{O}^{\trans}\widehat{\mathbf{Q}}$};
    \draw[fill=qcolor!15, draw=qcolor] (6.000,-0.42) rectangle (6.800,0.42);
    \node[] at (6.400,0) {$\mathbf{Y}^{\trans}\mathbf{T}^{\trans}$};
    \draw[fill=qcolor!15, draw=qcolor] (6.800,-0.42) rectangle (7.600,0.42);
    \node[] at (7.200,0) {$d\mathbf{U}^{\trans}\mathbf{T}$};
    \draw[fill=qcolor!15, draw=qcolor] (10.300,-0.42) rectangle (11.600,0.42);
    \node[] at (10.950,0) {$d\mathbf{S}'\,\mathbf{U}^{\trans}$};
    \draw[fill=scolor!20, draw=scolor] (0.000,4.58) rectangle (4.300,5.42);
    \node[align=center] at (2.150,5) {gate scan, $\widehat{\mathbf{Q}}$, $\widehat{\mathbf{K}}$, $\overline{\mathbf{K}}$\\$t{-}1$};
    \draw[fill=qcolor!15, draw=qcolor] (2.600,3.58) rectangle (3.900,4.42);
    \node[align=center] at (3.250,4) {$\mathbf{V} - \widehat{\mathbf{K}}\mathbf{S}$\\epilogue};
    \draw[fill=qcolor!15, draw=qcolor] (4.700,3.58) rectangle (5.600,4.42);
    \node[align=center] at (5.150,4) {$d\mathbf{U}$\\downcast};
    \draw[fill=qcolor!15, draw=qcolor] (6.800,3.58) rectangle (7.700,4.42);
    \node[align=center] at (7.250,4) {$\mathbf{U}$\\restage};
    \draw[fill=qcolor!15, draw=qcolor] (7.700,3.58) rectangle (9.600,4.42);
    \node[align=center] at (8.650,4) {$d\mathbf{Y}$ restage\\$-\boldsymbol{\beta} \odot d\mathbf{Y}$ staging};
    \draw[fill=gray!20, draw=gray] (0.000,2.58) rectangle (2.200,3.42);
    \node[align=center] at (1.100,3) {$d\mathbf{Q}$, $d\mathbf{K}$, $d\mathbf{g}$\\epilogue $t{+}1$};
    \draw[fill=qcolor!15, draw=qcolor] (2.200,2.58) rectangle (3.200,3.42);
    \node[align=center] at (2.700,3) {$d\boldsymbol{\gamma}$\\dot};
    \draw[fill=qcolor!15, draw=qcolor] (0.000,1.58) rectangle (1.300,2.42);
    \node[] at (0.650,2) {$\widehat{\mathbf{K}}\overline{\mathbf{K}}^{\trans}$};
    \draw[fill=qcolor!15, draw=qcolor] (1.300,1.58) rectangle (2.100,2.42);
    \node[align=center] at (1.700,2) {$\widehat{\mathbf{K}}\overline{\mathbf{K}}^{\trans}$\\epilogue};
    \draw[fill=qcolor!15, draw=qcolor] (2.100,1.58) rectangle (3.900,2.42);
    \node[] at (3.000,2) {$\mathbf{T}$ inverse};
    \draw[fill=qcolor!15, draw=qcolor] (0.000,0.58) rectangle (1.300,1.42);
    \node[] at (0.650,1) {$\widehat{\mathbf{Q}}\overline{\mathbf{K}}^{\trans}$};
    \draw[fill=qcolor!15, draw=qcolor] (1.300,0.58) rectangle (2.400,1.42);
    \node[align=center] at (1.850,1) {$\widehat{\mathbf{Q}}\overline{\mathbf{K}}^{\trans}$\\epilogue};
    \draw[fill=qcolor!15, draw=qcolor] (7.700,0.58) rectangle (9.000,1.42);
    \node[] at (8.350,1) {$d\mathbf{O}\mathbf{U}^{\trans}$};
    \draw[fill=qcolor!15, draw=qcolor] (9.000,0.58) rectangle (11.100,1.42);
    \node[align=center] at (10.050,1) {$d\mathbf{A}$\\epilogue};
\end{tikzpicture}

\smallskip

\begin{tikzpicture}[x=1.200cm, y=0.55cm, every node/.style={font=\tiny}]
    \path (-1.250, -0.6) rectangle (12.186, 6.05);
    \draw[dashed, gray] (6.800, -0.6) -- (6.800, 5.60);
    \node[anchor=south, font=\scriptsize, text=gray] at (4.500, 5.60) {one main-loop iteration (chunk $t$), second half};
    \node[anchor=east, font=\Large, inner sep=0pt, text=gray!50!black] at (-0.050, 2.50) {$\cdots$};
    \node[anchor=east, font=\scriptsize\itshape] at (-0.536, 5) {CG0};
    \node[anchor=east, font=\scriptsize\itshape] at (-0.536, 4) {CG1};
    \node[anchor=east, font=\scriptsize\itshape] at (-0.536, 3) {CG2};
    \node[anchor=east, font=\scriptsize\itshape] at (-0.536, 2) {Inverse};
    \node[anchor=east, font=\scriptsize\itshape] at (-0.536, 1) {Output};
    \node[anchor=east, font=\scriptsize\itshape] at (-0.536, 0) {MMA};
    \draw[fill=qcolor!15, draw=qcolor] (1.000,-0.42) rectangle (2.300,0.42);
    \node[] at (1.650,0) {$-d\mathbf{Y}^{\trans}\widehat{\mathbf{K}}$};
    \draw[fill=qcolor!15, draw=qcolor] (2.300,-0.42) rectangle (3.100,0.42);
    \node[] at (2.700,0) {$\widehat{\mathbf{Q}}^{\trans}d\mathbf{A}$};
    \draw[fill=qcolor!15, draw=qcolor] (3.100,-0.42) rectangle (3.900,0.42);
    \node[] at (3.500,0) {$\overline{\mathbf{K}}^{\trans}d\mathbf{A}^{\trans}$};
    \draw[fill=qcolor!15, draw=qcolor] (3.900,-0.42) rectangle (5.200,0.42);
    \node[] at (4.550,0) {$\mathbf{S}\,d\mathbf{Y}^{\trans}$};
    \draw[fill=qcolor!15, draw=qcolor] (5.200,-0.42) rectangle (6.000,0.42);
    \node[] at (5.600,0) {$-\widehat{\mathbf{K}}^{\trans}d\mathbf{M}$};
    \draw[fill=qcolor!15, draw=qcolor] (6.000,-0.42) rectangle (6.800,0.42);
    \node[] at (6.400,0) {$\overline{\mathbf{K}}^{\trans}d\mathbf{M}^{\trans}$};
    \draw[fill=qcolor!15, draw=qcolor] (0.000,3.58) rectangle (1.100,4.42);
    \node[align=center] at (0.550,4) {$d\mathbf{V}$\\staging};
    \draw[fill=qcolor!15, draw=qcolor] (1.100,3.58) rectangle (2.000,4.42);
    \node[align=center] at (1.550,4) {$d\boldsymbol{\beta}$\\(KDA)};
    \draw[fill=qcolor!15, draw=qcolor] (2.300,3.58) rectangle (4.100,4.42);
    \node[align=center] at (3.200,4) {$d\mathbf{S}$ rescale\\and restage};
    \draw[fill=qcolor!15, draw=qcolor] (3.900,2.58) rectangle (4.700,3.42);
    \node[align=center] at (4.300,3) {$d\mathbf{Q}$\\read};
    \draw[fill=qcolor!15, draw=qcolor] (6.800,2.58) rectangle (9.000,3.42);
    \node[align=center] at (7.900,3) {$d\mathbf{Q}$, $d\mathbf{K}$, $d\mathbf{g}$\\epilogue};
    \draw[fill=qcolor!15, draw=qcolor] (0.000,1.58) rectangle (1.000,2.42);
    \node[] at (0.500,2) {$d\mathbf{Y}\mathbf{U}^{\trans}$};
    \draw[fill=qcolor!15, draw=qcolor] (1.000,1.58) rectangle (2.500,2.42);
    \node[align=center] at (1.750,2) {$d\mathbf{M}$ epilogue\\($d\boldsymbol{\beta}$ term, KDA)};
    \draw[fill=gray!20, draw=gray] (0.000,0.58) rectangle (2.200,1.42);
    \node[align=center] at (1.100,1) {$d\mathbf{Q}$, $d\mathbf{K}$, $d\mathbf{g}$, $d\mathbf{V}$ stores\\$t{+}1$};
    \draw[fill=gray!20, draw=gray] (2.200,0.58) rectangle (3.600,1.42);
    \node[align=center] at (2.900,1) {$d\boldsymbol{\beta}$, $d\mathbf{w}$ stores\\$t{+}1$ (GDN-2)};
\end{tikzpicture}
\caption{The KDA and GDN-2 backward pipeline. Blue is chunk $t$, gray chunk $t{+}1$, visited before it, and orange chunk $t{-}1$, visited next.}
\label{fig:pipeline-bwd-kda}
\end{figure}
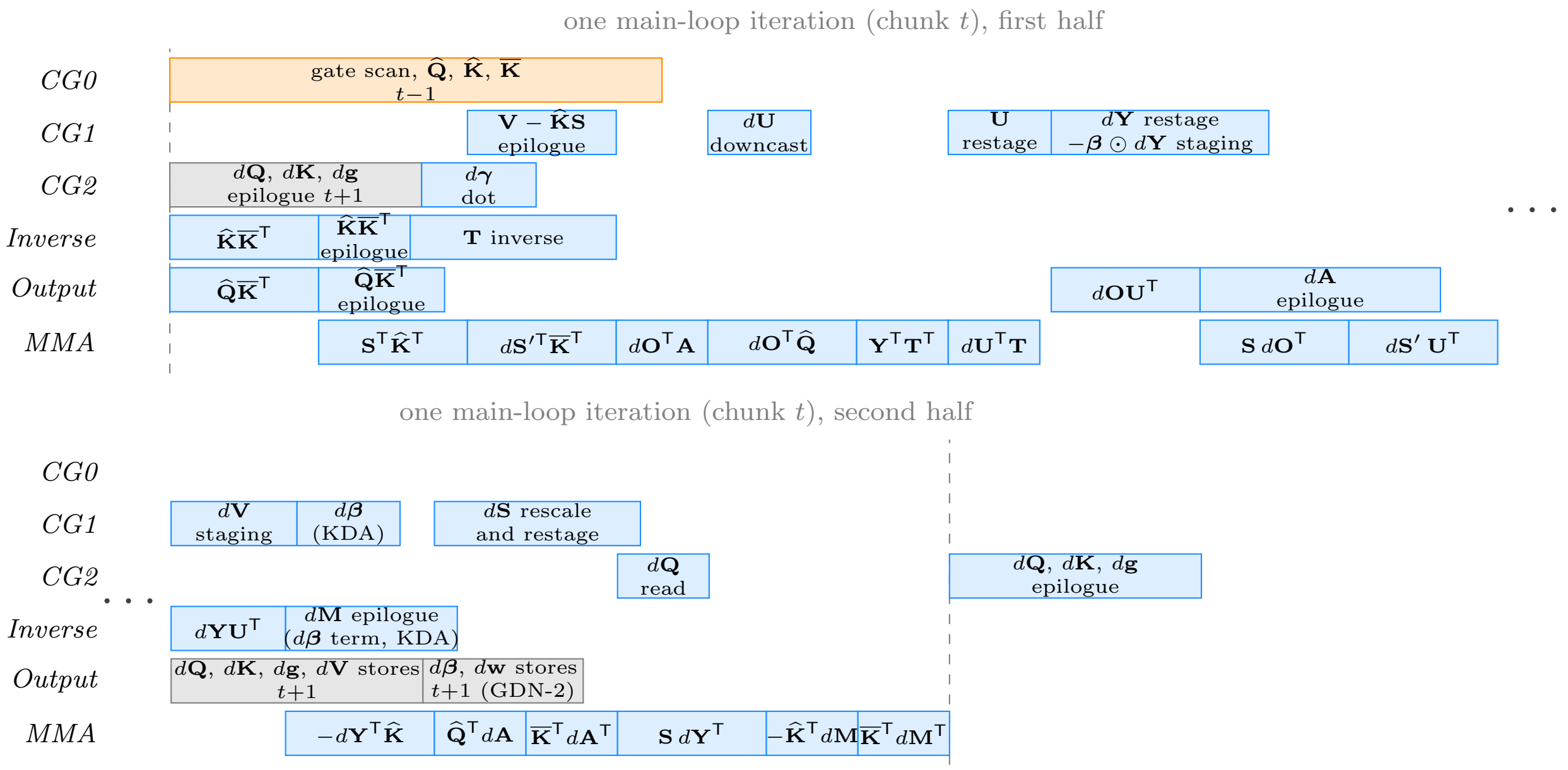

Figure~\ref{fig:pipeline-bwd-kda} shows one main-loop iteration. The KDA and GDN-2 kernels roughly share the same schedule with a few exceptions. GDN-2 adds the $d\mathbf{w}$ staging to compute group 1, the per-channel $d\boldsymbol{\beta}$ to compute group 2, and two additional stores to the epilogue warp. KDA instead computes its per-token $d\boldsymbol{\beta}$ in compute group 1 from the $\mathbf{Y}$ terms and the $d\mathbf{M}$ term that the register-MMA warp reduces, and writes it with plain global memory stores.

Similar to prefill, compute group 0 runs one chunk ahead. It runs chunk $t{-}1$'s gate prefix scan and scales $\mathbf{Q}$ and $\mathbf{K}$ into the decay-folded shared memory operands $\widehat{\mathbf{Q}}$, $\widehat{\mathbf{K}}$, and $\overline{\mathbf{K}}$. Note that for the backward kernels we only allocate one warpgroup to this because we have enough MMAs to fully hide this latency. Compute group 1 rescales $d\mathbf{S}$ in TMEM, stages $\mathbf{Y}$, downcasts $d\mathbf{U}$, restages $\mathbf{U}$ and $d\mathbf{Y}$, stages $-\boldsymbol{\beta} \odot d\mathbf{Y}$ and $d\mathbf{V}$, and owns the $\mathbf{Y}$ term of $d\boldsymbol{\beta}$. Compute group 2 reduces the $d\boldsymbol{\gamma}$ dot products and the state share of $d\mathbf{g}$, reads the $d\mathbf{Q}$ accumulator, and after the last $d\mathbf{K}$ MMA folds the $d\mathbf{Q}$ and $d\mathbf{K}$ parts with the $d\mathbf{A}$ diagonal, applies the L2-norm backward, finishes $d\mathbf{g}$ with the reverse cumulative sum, and stages all three for the epilogue warp. The inverse row is the register-MMA warp, which computes $\widehat{\mathbf{K}}\overline{\mathbf{K}}^{\trans}$, masks it to its strictly lower triangle, and computes the inverse before the $\mathbf{U}$ MMA, then $d\mathbf{M}_{[t]}$ from $d\mathbf{Y}$ and $\mathbf{U}$. The output row is the epilogue warp, which computes $\mathbf{A}_{[t]}$ and $d\mathbf{A}_{[t]}$ on register MMAs and stores $d\mathbf{Q}$, $d\mathbf{K}$, $d\mathbf{V}$, and $d\mathbf{g}$, and $d\boldsymbol{\beta}$ and $d\mathbf{w}$ for GDN-2.
\end{document}

%% file: figures/results_fwd_lowbh.tex
{\footnotesize \textcolor{flacolor}{\rule{0.75em}{0.75em}}\,FLA (Triton) \quad \textcolor{mssmcolor}{\rule{0.75em}{0.75em}}\,mamba\_ssm (Triton) \quad \textcolor{qlacolor}{\rule{0.75em}{0.75em}}\,FlashQLA (TileLang) \\
\textcolor{fkdacolor}{\rule{0.75em}{0.75em}}\,FlashKDA (CUTLASS) \quad \textcolor{ourscolor}{\rule{0.75em}{0.75em}}\,Ours \quad \textcolor{oursoncolor}{\rule{0.75em}{0.75em}}\,Ours (state on)\par}
\smallskip
\begin{tikzpicture}
\begin{groupplot}[group style={group size=3 by 1, horizontal sep=1.4cm}, perfaxis]
\nextgroupplot[title={Mamba-2, $H_{qk} = 8$, $H_v = 16$}, ymax=500, ylabel={TFLOPS}]
    \addplot[fill=flacolor] coordinates {(2048,34) (8192,42) (32768,43)};
    \addplot[fill=mssmcolor] coordinates {(2048,69) (8192,115) (32768,114)};
    \addplot[fill=ourscolor] coordinates {(2048,91) (8192,236) (32768,377)};
    \addplot[fill=oursoncolor] coordinates {(2048,81) (8192,206) (32768,340)};
    \node[speedup] at (axis cs:2048,91) {$2.6\times$};
    \node[speedup] at (axis cs:8192,236) {$5.5\times$};
    \node[speedup] at (axis cs:32768,377) {$8.7\times$};
\nextgroupplot[title={GDN, $H_{qk} = 4$, $H_v = 16$}, ymax=400]
    \addplot[fill=flacolor] coordinates {(2048,47) (8192,56) (32768,59)};
    \addplot[fill=qlacolor] coordinates {(2048,50) (8192,94) (32768,129)};
    \addplot[fill=ourscolor] coordinates {(2048,71) (8192,170) (32768,287)};
    \addplot[fill=oursoncolor] coordinates {(2048,64) (8192,150) (32768,243)};
    \node[speedup] at (axis cs:2048,71) {$1.5\times$};
    \node[speedup] at (axis cs:8192,170) {$3.0\times$};
    \node[speedup] at (axis cs:32768,287) {$4.9\times$};
\nextgroupplot[title={GDP ($n_h = 3$), $H = 16$}, ymax=300]
    \addplot[fill=flacolor] coordinates {(2048,38) (8192,42) (32768,43)};
    \addplot[fill=ourscolor] coordinates {(2048,105) (8192,175) (32768,224)};
    \addplot[fill=oursoncolor] coordinates {(2048,100) (8192,167) (32768,209)};
    \node[speedup] at (axis cs:2048,105) {$2.8\times$};
    \node[speedup] at (axis cs:8192,175) {$4.2\times$};
    \node[speedup] at (axis cs:32768,224) {$5.2\times$};
\end{groupplot}
\end{tikzpicture}
\par\vspace{2pt}
\begin{tikzpicture}
\begin{groupplot}[group style={group size=2 by 1, horizontal sep=1.4cm}, perfaxis]
\nextgroupplot[title={KDA, $H = 16$}, ymax=150, ylabel={TFLOPS}]
    \addplot[fill=flacolor] coordinates {(2048,17) (8192,17) (32768,18)};
    \addplot[fill=fkdacolor] coordinates {(2048,14) (8192,15) (32768,15)};
    \addplot[fill=ourscolor] coordinates {(2048,36) (8192,69) (32768,90)};
    \addplot[fill=oursoncolor] coordinates {(2048,31) (8192,58) (32768,75)};
    \node[speedup] at (axis cs:2048,36) {$2.1\times$};
    \node[speedup] at (axis cs:8192,69) {$4.0\times$};
    \node[speedup] at (axis cs:32768,90) {$4.9\times$};
\nextgroupplot[title={GDN-2, $H = 16$}, ymax=150]
    \addplot[fill=flacolor] coordinates {(2048,16) (8192,17) (32768,18)};
    \addplot[fill=ourscolor] coordinates {(2048,34) (8192,68) (32768,90)};
    \addplot[fill=oursoncolor] coordinates {(2048,30) (8192,57) (32768,73)};
    \node[speedup] at (axis cs:2048,34) {$2.1\times$};
    \node[speedup] at (axis cs:8192,68) {$4.0\times$};
    \node[speedup] at (axis cs:32768,90) {$5.0\times$};
\end{groupplot}
\end{tikzpicture}

%% file: figures/results_bwd_lowbh.tex
{\footnotesize \textcolor{flacolor}{\rule{0.75em}{0.75em}}\,FLA (Triton) \quad \textcolor{mssmcolor}{\rule{0.75em}{0.75em}}\,mamba\_ssm (Triton) \quad \textcolor{qlacolor}{\rule{0.75em}{0.75em}}\,FlashQLA (TileLang) \quad \textcolor{ourscolor}{\rule{0.75em}{0.75em}}\,Ours \quad \textcolor{oursoncolor}{\rule{0.75em}{0.75em}}\,Ours (state on)\par}
\smallskip
\begin{tikzpicture}
\begin{groupplot}[group style={group size=3 by 1, horizontal sep=1.4cm}, perfaxis]
\nextgroupplot[title={Mamba-2, $H_{qk} = 8$, $H_v = 16$}, ymax=500, ylabel={TFLOPS}]
    \addplot[fill=flacolor] coordinates {(2048,47) (8192,55) (32768,59)};
    \addplot[fill=mssmcolor] coordinates {(2048,39) (8192,59) (32768,66)};
    \addplot[fill=ourscolor] coordinates {(2048,87) (8192,199) (32768,286)};
    \addplot[fill=oursoncolor] coordinates {(2048,112) (8192,251) (32768,361)};
    \node[speedup] at (axis cs:2048,112) {$2.4\times$};
    \node[speedup] at (axis cs:8192,251) {$4.6\times$};
    \node[speedup] at (axis cs:32768,361) {$6.2\times$};
\nextgroupplot[title={GDN, $H_{qk} = 4$, $H_v = 16$}, ymax=550]
    \addplot[fill=flacolor] coordinates {(2048,47) (8192,56) (32768,58)};
    \addplot[fill=qlacolor] coordinates {(2048,93) (8192,168) (32768,221)};
    \addplot[fill=ourscolor] coordinates {(2048,105) (8192,210) (32768,299)};
    \addplot[fill=oursoncolor] coordinates {(2048,154) (8192,296) (32768,404)};
    \node[speedup] at (axis cs:2048,154) {$3.3\times$};
    \node[speedup] at (axis cs:8192,296) {$5.3\times$};
    \node[speedup] at (axis cs:32768,404) {$7.0\times$};
\nextgroupplot[title={GDP ($n_h = 3$), $H = 16$}, ymax=450]
    \addplot[fill=flacolor] coordinates {(2048,38) (8192,41) (32768,42)};
    \addplot[fill=ourscolor] coordinates {(2048,148) (8192,219) (32768,255)};
    \addplot[fill=oursoncolor] coordinates {(2048,204) (8192,295) (32768,339)};
    \node[speedup] at (axis cs:2048,204) {$5.4\times$};
    \node[speedup] at (axis cs:8192,295) {$7.1\times$};
    \node[speedup] at (axis cs:32768,339) {$8.1\times$};
\end{groupplot}
\end{tikzpicture}
\par\vspace{2pt}
\begin{tikzpicture}
\begin{groupplot}[group style={group size=2 by 1, horizontal sep=1.4cm}, perfaxis]
\nextgroupplot[title={KDA, $H = 16$}, ymax=150, ylabel={TFLOPS}]
    \addplot[fill=flacolor] coordinates {(2048,19) (8192,21) (32768,22)};
    \addplot[fill=ourscolor] coordinates {(2048,36) (8192,56) (32768,65)};
    \addplot[fill=oursoncolor] coordinates {(2048,51) (8192,77) (32768,89)};
    \node[speedup] at (axis cs:2048,51) {$2.6\times$};
    \node[speedup] at (axis cs:8192,77) {$3.6\times$};
    \node[speedup] at (axis cs:32768,89) {$4.0\times$};
\nextgroupplot[title={GDN-2, $H = 16$}, ymax=150]
    \addplot[fill=flacolor] coordinates {(2048,19) (8192,21) (32768,22)};
    \addplot[fill=ourscolor] coordinates {(2048,33) (8192,52) (32768,60)};
    \addplot[fill=oursoncolor] coordinates {(2048,47) (8192,72) (32768,81)};
    \node[speedup] at (axis cs:2048,47) {$2.5\times$};
    \node[speedup] at (axis cs:8192,72) {$3.5\times$};
    \node[speedup] at (axis cs:32768,81) {$3.7\times$};
\end{groupplot}
\end{tikzpicture}

%% file: figures/results_fwd_b4.tex
{\footnotesize \textcolor{flacolor}{\rule{0.75em}{0.75em}}\,FLA (Triton) \quad \textcolor{mssmcolor}{\rule{0.75em}{0.75em}}\,mamba\_ssm (Triton) \quad \textcolor{qlacolor}{\rule{0.75em}{0.75em}}\,FlashQLA (TileLang) \\
\textcolor{fkdacolor}{\rule{0.75em}{0.75em}}\,FlashKDA (CUTLASS) \quad \textcolor{ourscolor}{\rule{0.75em}{0.75em}}\,Ours \quad \textcolor{oursoncolor}{\rule{0.75em}{0.75em}}\,Ours (state on)\par}
\smallskip
\begin{tikzpicture}
\begin{groupplot}[group style={group size=3 by 1, horizontal sep=1.4cm}, perfaxis]
\nextgroupplot[title={Mamba-2, $H_{qk} = 8$, $H_v = 64$}, ymax=1000, ylabel={TFLOPS}]
    \addplot[fill=flacolor] coordinates {(2048,51) (8192,53) (32768,53)};
    \addplot[fill=mssmcolor] coordinates {(2048,184) (8192,201) (32768,198)};
    \addplot[fill=ourscolor] coordinates {(2048,571) (8192,703) (32768,747)};
    \addplot[fill=oursoncolor] coordinates {(2048,505) (8192,620) (32768,658)};
    \node[speedup] at (axis cs:2048,571) {$11.3\times$};
    \node[speedup] at (axis cs:8192,703) {$13.4\times$};
    \node[speedup] at (axis cs:32768,747) {$14.2\times$};
\nextgroupplot[title={GDN, $H_{qk} = 16$, $H_v = 64$}, ymax=850]
    \addplot[fill=flacolor] coordinates {(2048,102) (8192,106) (32768,108)};
    \addplot[fill=qlacolor] coordinates {(2048,187) (8192,195) (32768,197)};
    \addplot[fill=ourscolor] coordinates {(2048,498) (8192,587) (32768,619)};
    \addplot[fill=oursoncolor] coordinates {(2048,393) (8192,444) (32768,456)};
    \node[speedup] at (axis cs:2048,498) {$4.9\times$};
    \node[speedup] at (axis cs:8192,587) {$5.5\times$};
    \node[speedup] at (axis cs:32768,619) {$5.7\times$};
\nextgroupplot[title={GDP ($n_h = 3$), $H = 40$}, ymax=450]
    \addplot[fill=flacolor] coordinates {(2048,79) (8192,81) (32768,80)};
    \addplot[fill=ourscolor] coordinates {(2048,263) (8192,318) (32768,344)};
    \addplot[fill=oursoncolor] coordinates {(2048,243) (8192,293) (32768,317)};
    \node[speedup] at (axis cs:2048,263) {$3.3\times$};
    \node[speedup] at (axis cs:8192,318) {$3.9\times$};
    \node[speedup] at (axis cs:32768,344) {$4.3\times$};
\end{groupplot}
\end{tikzpicture}
\par\vspace{2pt}
\begin{tikzpicture}
\begin{groupplot}[group style={group size=2 by 1, horizontal sep=1.4cm}, perfaxis]
\nextgroupplot[title={KDA, $H = 96$}, ymax=300, ylabel={TFLOPS}]
    \addplot[fill=flacolor] coordinates {(2048,34) (8192,35) (32768,35)};
    \addplot[fill=fkdacolor] coordinates {(2048,96) (8192,98) (32768,98)};
    \addplot[fill=ourscolor] coordinates {(2048,190) (8192,199) (32768,201)};
    \addplot[fill=oursoncolor] coordinates {(2048,127) (8192,130) (32768,131)};
    \node[speedup] at (axis cs:2048,190) {$5.5\times$};
    \node[speedup] at (axis cs:8192,199) {$5.6\times$};
    \node[speedup] at (axis cs:32768,201) {$5.7\times$};
\nextgroupplot[title={GDN-2, $H = 64$}, ymax=250]
    \addplot[fill=flacolor] coordinates {(2048,30) (8192,31) (32768,31)};
    \addplot[fill=ourscolor] coordinates {(2048,178) (8192,187) (32768,189)};
    \addplot[fill=oursoncolor] coordinates {(2048,121) (8192,125) (32768,125)};
    \node[speedup] at (axis cs:2048,178) {$5.9\times$};
    \node[speedup] at (axis cs:8192,187) {$6.0\times$};
    \node[speedup] at (axis cs:32768,189) {$6.1\times$};
\end{groupplot}
\end{tikzpicture}

%% file: figures/results_bwd_b4.tex
{\footnotesize \textcolor{flacolor}{\rule{0.75em}{0.75em}}\,FLA (Triton) \quad \textcolor{mssmcolor}{\rule{0.75em}{0.75em}}\,mamba\_ssm (Triton) \quad \textcolor{qlacolor}{\rule{0.75em}{0.75em}}\,FlashQLA (TileLang) \quad \textcolor{ourscolor}{\rule{0.75em}{0.75em}}\,Ours \quad \textcolor{oursoncolor}{\rule{0.75em}{0.75em}}\,Ours (state on)\par}
\smallskip
\begin{tikzpicture}
\begin{groupplot}[group style={group size=3 by 1, horizontal sep=1.4cm}, perfaxis]
\nextgroupplot[title={Mamba-2, $H_{qk} = 8$, $H_v = 64$}, ymax=650, ylabel={TFLOPS}]
    \addplot[fill=flacolor] coordinates {(2048,76) (8192,78) (32768,78)};
    \addplot[fill=mssmcolor] coordinates {(2048,125) (8192,141) (32768,140)};
    \addplot[fill=ourscolor] coordinates {(2048,379) (8192,427) (32768,424)};
    \addplot[fill=oursoncolor] coordinates {(2048,439) (8192,489) (32768,495)};
    \node[speedup] at (axis cs:2048,439) {$5.8\times$};
    \node[speedup] at (axis cs:8192,489) {$6.3\times$};
    \node[speedup] at (axis cs:32768,495) {$6.4\times$};
\nextgroupplot[title={GDN, $H_{qk} = 16$, $H_v = 64$}, ymax=700]
    \addplot[fill=flacolor] coordinates {(2048,93) (8192,95) (32768,93)};
    \addplot[fill=qlacolor] coordinates {(2048,244) (8192,258) (32768,263)};
    \addplot[fill=ourscolor] coordinates {(2048,371) (8192,400) (32768,403)};
    \addplot[fill=oursoncolor] coordinates {(2048,476) (8192,509) (32768,506)};
    \node[speedup] at (axis cs:2048,476) {$5.1\times$};
    \node[speedup] at (axis cs:8192,509) {$5.4\times$};
    \node[speedup] at (axis cs:32768,506) {$5.5\times$};
\nextgroupplot[title={GDP ($n_h = 3$), $H = 40$}, ymax=550]
    \addplot[fill=flacolor] coordinates {(2048,73) (8192,73) (32768,72)};
    \addplot[fill=ourscolor] coordinates {(2048,281) (8192,326) (32768,342)};
    \addplot[fill=oursoncolor] coordinates {(2048,339) (8192,388) (32768,401)};
    \node[speedup] at (axis cs:2048,339) {$4.6\times$};
    \node[speedup] at (axis cs:8192,388) {$5.4\times$};
    \node[speedup] at (axis cs:32768,401) {$5.6\times$};
\end{groupplot}
\end{tikzpicture}
\par\vspace{2pt}
\begin{tikzpicture}
\begin{groupplot}[group style={group size=2 by 1, horizontal sep=1.4cm}, perfaxis]
\nextgroupplot[title={KDA, $H = 96$}, ymax=200, ylabel={TFLOPS}]
    \addplot[fill=flacolor] coordinates {(2048,32) (8192,32) (32768,31)};
    \addplot[fill=ourscolor] coordinates {(2048,94) (8192,96) (32768,97)};
    \addplot[fill=oursoncolor] coordinates {(2048,121) (8192,124) (32768,125)};
    \node[speedup] at (axis cs:2048,121) {$3.8\times$};
    \node[speedup] at (axis cs:8192,124) {$3.9\times$};
    \node[speedup] at (axis cs:32768,125) {$4.0\times$};
\nextgroupplot[title={GDN-2, $H = 64$}, ymax=150]
    \addplot[fill=flacolor] coordinates {(2048,30) (8192,31) (32768,31)};
    \addplot[fill=ourscolor] coordinates {(2048,78) (8192,80) (32768,81)};
    \addplot[fill=oursoncolor] coordinates {(2048,95) (8192,97) (32768,98)};
    \node[speedup] at (axis cs:2048,95) {$3.2\times$};
    \node[speedup] at (axis cs:8192,97) {$3.2\times$};
    \node[speedup] at (axis cs:32768,98) {$3.2\times$};
\end{groupplot}
\end{tikzpicture}